\documentclass[a4paper,fleqn]{cas-sc}

\usepackage[numbers]{natbib}
\usepackage{times,latexsym}
\usepackage{url}
\usepackage[T1]{fontenc}
\usepackage{enumitem}
\usepackage{amsmath}
\usepackage{booktabs}
\usepackage{subcaption}  
\usepackage{listings}
\usepackage{tcolorbox}
\usepackage{xcolor}
\usepackage{tabularx}
\usepackage{siunitx}
\usepackage{float}
\newtheorem{defi}{Definition:}

\tcbuselibrary{listings,breakable}
\def\tsc#1{\csdef{#1}{\textsc{\lowercase{#1}}\xspace}}
\tsc{WGM}
\tsc{QE}

\begin{document}
\let\WriteBookmarks\relax
\def\floatpagepagefraction{1}
\def\textpagefraction{.001}

\shorttitle{Trustworthiness of LLM-summarizers}    


\title [mode = title]{Reexamining zero-shot summarization: Empirical investigation of trustworthiness of LLM-summarizers}  



%
\author[1]{Vasudha Bhatnagar}[
                         orcid=0000-0002-9706-9340 
                          ]

\ead{vbhatnagar@cs.du.ac.in}



\affiliation[1]{organization={Department of Computer Science, University of Delhi},
            city={Delhi},
            postcode={110007}, 
            state={Delhi},
            country={India}}

\author[1]{Purnima Bindal}[
                          orcid=0000-0001-5108-0117]

\cormark[1]
\ead{pbindal@cs.du.ac.in}



%

\author[1]{Vikas Kumar}[
                          orcid=0000-0001-9882-7310
                          ]

\ead{vikas@cs.du.ac.in}


            
\author[2]{Raj Kumari Bahl}[
                          ]

\ead{rajkumaribahl@ramjas.du.ac.in}


\affiliation[2]{organization={Department of Statistics, Ramjas College, University of Delhi},
            city={Delhi},
            postcode={110007}, 
            state={Delhi},
            country={India}}
            
\cortext[cor1]{Corresponding author}



\begin{abstract}
Zero-shot summarization using Large Language Models (LLMs) has significantly advanced the abstractive summarization task by producing coherent and fluent summaries.  However,  underlying  stochasticity of the large language models  raises concerns about the \textit{stability} and \textit{trustworthiness} of the LLM-generated summaries. This issue has become increasingly important due to proliferation of LLM-generated summaries in educational settings, where students and researchers summarize complex academic materials in zero-shot manner. 

We propose a novel two-level diagnostic protocol for benchmarking LLM-summarizers based on the \textit{stability} of the generated summaries. At the lower level, document-level stability analysis is performed over multiple LLM-summaries generated under controlled environment, and the \textit{stability coefficient} is computed. Each generated summary is scored for semantic and factual alignment with the original document, enabling estimation of stability along more than one dimensions.  At the next level, observations from a stratified  sample of documents drawn from the corpus are consolidated to estimate the \textit{stability index} of the LLM-summarizer, which is the proxy for its \textit{trustworthiness}. 

Our  empirical investigation of  three LLM-summarizers across three genres of documents reveals statistically significant differences in  the generation-level variability among LLMs across summary evaluation metrics. This  study  advances the LLM-summarization research by evidential recognition of the stability problem in LLM-summaries and motivates further research towards development of robust, reliable and  trustworthy LLM-summarizers. 
\end{abstract}
 



\maketitle

\section{Introduction}
\label{sec:introduction}
The ability of Large Language Models (LLMs)  to generate fluent, human-like summaries has led to their widespread adoption for quick assimilation of information from text documents of varying lengths and genres. Their applications span multiple domains, including news and sports articles consumed by general public \citep{2024-sports-kang,2025-news-premnath,2023-news-tam,2025-news-you,2024-news-zhang}, scientific literature  distilled by students, scholars and academics \citep{2023-literature-antu,2024-scientific-keya,2024-scientific-procko,2025-student-xie, 2025-education-zhong}, legal documents summarized by law students and professionals \cite{2024-legal-deroy,2024-legal-hakim,2024-legal-liu,2025-legal-mentzingen, 2025-legal-song},  summarizing conversations and meetings \cite{2025-conversation-gupta, 2024-dialogue-ramprasad}, etc. In the present era of generative AI (GenAI), LLMs have become the de-facto engines behind modern text summarization systems \cite{survey-akter2025comprehensive, survey-dhaini2024explainability, survey-liu2024sum, survey-2024-zhang}.  Variations due to rephrasing of prompts often yield large performance differences in the LLM output \cite{2023-related-sclar-quantifying}. 

\begin{figure}
    \centering
    \includegraphics{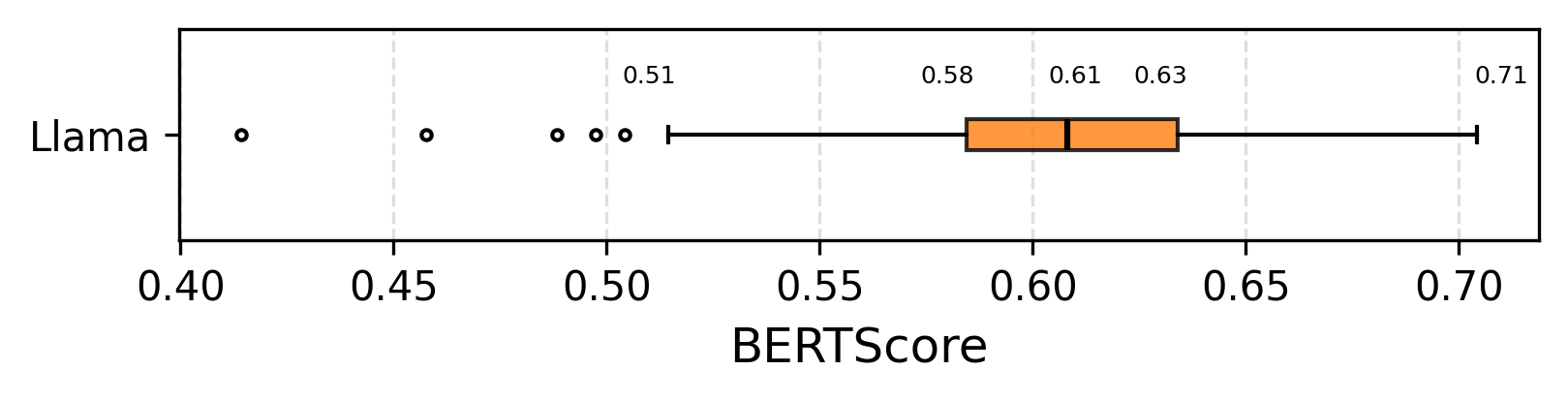}
    \caption{Distribution of BERTScore values  of 100 summaries generated by Llama-summarizer for a document from CNN/DM dataset.}
    \label{fig:bertscore-3506}
\end{figure}

Benchmarking, which is an importance task in  LLM-summarization research \cite{2025-zhang-survey},  essentially entails extensive multi-dimensional evaluation of LLM-generated summaries over multiple corpora. However, the conventional summary evaluation framework  relies on point estimate based  on assessment of a single summary per document for the corpus, while ignoring the  stochastic variations due to probabilistic decoding strategies (e.g.  sampling, temperature scaling, etc.) employed in LLMs. To gain insight into the stochastic variations in LLM-generated summaries,  we select a random document  from CNN/DM dataset\footnote{See Sec. \ref{sec:datasets} for details of the CNN/DM dataset} and generate hundred  summaries  using  Llama summarizer\footnote{See details in Sec. \ref{subsec:llms}} with fixed input prompt and parameters. Next, we compute BERTScore for each generated summary against the corresponding gold standard reference summary of the document and visualize the score distribution using a box-plot (Fig.~\ref{fig:bertscore-3506}). Substantial and significant semantic variations in the generated summaries are obvious, with the BERTScore values ranging from [0.41, 0.71]. These variations are  too significant to be  ignored, and raise concerns about repeatability and reproducibility of LLM-generated summaries. Furthermore, this observation  motivates the conjecture that the  assessment of an LLM-summarizer based on  single generated summary per document  leads to \textit{untrustworthy} outcome. 

\subsection{Significance of the Problem}
Robustness, one of the dimensions of trustworthy AI, is defined by \citet{2023-AI-risk-tabassi} as a \textit{system’s ability to maintain its performance level under  various circumstances}. Variations along semantic and factuality dimensions in LLM-generated summaries, even under controlled (static) environment may lead to differences in emphasized facts and  omitted information, conflicting interpretations. Such variability may undermine the robustness and \textit{trustworthiness} of LLM-summarizers. Figure \ref{fig:bertscore-3506}, which deprecates the stability of  LLM-generated summaries, is a grim reminder to revisit zero-shot summarization and audit it for \textit{robustness}.  Students, researchers, medical and law practitioners, who often summarize long documents to enhance productivity, are likely to undermine the LLM-generated summaries because of the lack of robustness and reliability.  For example, members in a team of law professionals using an LLM-summarizer for  case preparation may arrive at  different conclusions if the the summaries are different. Significant variations in the summaries of news articles may escalate spread of misinformation.  If  summarization is a step in an NLP pipeline, then the variability in the LLM-generated summary may cascade  to the performance of the downstream application. 

Another casualty arising from the lack of robustness is the conventional summarization evaluation framework, which relies on a single generated summary for quality assessment. Traditional summary evaluation methods (pre-LLM era) are inadequate for capturing high-level semantic qualities of summaries \cite{2024-llm-generated-field}. Summary evaluation frameworks  have  evolved to assess more summary attributes including  truthfulness and robustness of LLM-summarizers.  However, variability in LLM-generated summaries  has not been attended. Ignoring this caveat may result in potentially incomplete and unreliable evaluation of summary quality, which may lead to unreliable benchmarking of the LLM-summarizers.  Ergo, \textit{variability} (conversely, stability)  in LLM-generated summaries merits a careful and detailed study to strengthen research for  developing \textit{robust and trustworthy} LLM-summarizers. 
\subsection{Related Work}
Recently, reliability of  LLMs and LLM-NLP tasks has  come under close  scrutiny by the research community \citep{2026-efficient-inference-chen,2024-trustllm-huang,2025-using-huidrom,2025-llms-li,2025-not-abstain-liu,2025-investigating-lupart,2025-tools-lymperopoulos,2025-evaluating-nikitha,2024-can-trust-schroeder,2026-coin-flip-schwinn,2026-trustworthy-zhang}. \citet{2024-trustllm-huang} establish a benchmark of trustworthiness for mainstream LLMs based along six dimensions including truthfulness, safety, fairness, robustness, privacy, and machine ethics. The authors observe that robustness of LLMs shows significant variability, especially in open-ended tasks and out-of-distribution tasks. Recent studies regarding robustness and trustworthiness of LLM-as-judge report similar serious concerns \citep{2026-efficient-inference-chen,2025-using-huidrom,2025-llms-li,2024-can-trust-schroeder,2026-coin-flip-schwinn}. The sensitivity of large language models (LLMs) to prompt templates is investigated  by \citet{2024-mizrahi-state}, who introduce a multi-prompt evaluation framework that aggregates performance across paraphrased instruction templates.
Efforts to estimate performance distributions over multiple prompt variants have been proposed to improve robustness in LLM evaluation~\cite{2024-related-polo-efficient}. \citet{23-GPT-consistency} investigate \textit{trustworthiness} of two GPT models for consistency in logical predictions based on negation symmetry, transitivity, and semantic equivalence. The study indicates  that the two models fail to generate logically consistent predictions, motivating deeper investigation of consistency in each property.

\citet{2025-investigating-lupart} study LLM output variability and underscore the importance of rigorous experimental design when evaluating LLM-personalized conversational IR pipelines. \citet{2025-evaluating-nikitha} highlights the limitations of current evaluation practices and presents a critical examination of  the issue of \textit{variance} in performance caused by  stochastic model outputs, training seed sensitivity, and hyperparameter configurations. Uncertainty in LLM-QA systems has been studied by \citet{2025-tools-lymperopoulos}. \citet{2024-wang-prompt} employ multi-prompt formulations to assess the reliability of LLMs in medical question answering, showing that reliability varies substantially across prompt styles through repeated querying. 

Existing benchmarking studies for LLM-summarizers focus on conventional summary attributes like relevance, coherence, factual consistency, summary length, etc., and report results based on the single summary per document \cite{2025-zhang-survey,survey-2024-zhang}. The outcomes reported in these  studies are completely oblivious of the possible variations in LLM-summaries at the generation level \citep{gilhuly2025consistency,2024-llm-generated-field, zhang2026assessing}. \citet{peturbation-oved2021pass} introduce variations in the input to generate diverse summary candidates, and select the most coherent and informative summary by improving overall summary quality and consistency.  Furthermore, recent studies suggest that large language models are sensitive to textual perturbations such as typographical errors, character-level, and word-level modifications, which often lead to noticeable changes in generated responses~\cite{peturbation-alahmari2025large,peturbation-romero2024resilient, peturbation-singh2024robustness}.


\textit{Systematic  investigation and quantification of the stability of LLM-summarizers under controlled (static) input parameters remains unexplored. We address this research gap by designing an empirical study of generational variability  in the LLM-generated summaries under static input parameters and controlled environment.  Based on the results, we advocate for variation-aware  evaluation framework  to foster trustworthy benchmarking of LLM-summarizers.}

\subsection{Contributions}
The current study addresses the problem of variability in LLM-generated summaries and admits \textit{stability}  as a proxy for its \textit{robustness} and \textit{reliability} of an  LLM-summarizer. An extensive empirical study  expounds compelling reasons to broaden the scope of evaluation of LLM-summarizers  by  including multi-run evaluation, and  reporting the \textit{stability} of the LLM-summarizers. To the best of our knowledge, this is the first work to systematically investigate and quantify the \textit{stability} of LLM-summarizers. Our findings provide timely and impactful insights for developing trustworthy LLM-summarizers. The following  are the specific contributions.
\begin{enumerate}[label=(\roman*), nosep]
    \item We propose a simple, efficient, and robust diagnostic protocol to analyze variability in  LLM-generated summaries under static input environment.
    \item We propose a measure to estimate the  \textit{stability} of an LLM-summarizer, and  benchmark three  LLM-summarizers for three document genres after extensive experimentation based on more than 28K LLM-summaries.
    \item We demonstrate notable differences in the behavior of evaluation metrics due to variability in  LLM-generated summaries. 
    \item We establish strong empirical evidence to support argument for broadening  the scope of evaluation of LLM-summarizers and include \textit{stability} as additional dimension.
\end{enumerate}
\section{Problem Statement and Approach}
\label{sec:formal-problem-statement}
Trustworthiness of an LLM-summarizer primarily hinges on several summary attributes including truthfulness, safety,  reliability and robustness.  However, repeatability and reproducibility of LLM-summaries, which are the hallmark of the robustness of an LLM-summarizer have been overlooked. Considering that the \textit{stability} of LLM-summaries connotes robustness of an LLM-summarizer, we address the problem of quantifying its \textit{stability}. Stability measures the extent of similarity between  multiple summaries generated  for a document in a  static environment, and informs the end-user about the trust they can place in the LLM-generated summary. 

Broadly, the problem  addressed in this work is to quantify the stability of an LLM-summarizer. It is well known fact that the quality of LLM-summary is influence by several factors such as document length, type, linguistic style, structural complexity and domain-specific terminology. Furthermore, the summary quality is assessed along different dimensions (e.g. lexical, semantic, faithfulness, relevance, fluency etc.), and is quantified by different metrics having  their own idiosyncrasies.  Overall,  the problem is not straight forward and its different aspects need to be dis-entangled  as follows. \\ \textit{Given an LLM-summarizer $\mathcal L$, a document $D$ belonging to a genre $\mathcal G$, and a summary evaluation metric $m$, the problem is to quantify the trust a user can place in the quality of generated summary as measured by $m$.} \\
An LLM-summarizer that produces semantically similar summaries of a document across repeated generations inspires greater confidence than one whose summaries vary substantially. Since, quantifying the \textit{stability} of the LLM-summary for a single document is of little practical utility, it is imperative to generalize the notion of stability of the LLM-summarizer for documents belonging to a specific genre. Accordingly, we first estimate the \textit{stability} of the LLM-summarizer for a document, and subsequently generalize it to the corpus-level, as described below.
%
\subsection{Document-level stability} 
    Document-level stability measures the extent to which the repeatedly generated LLM-summaries for a given document remains consistent. Formally the problem is stated as follows. 
    
    Let $D$ be a  document and  $\mathbf{S}$ be the space of possible summaries of $D$, generated by an LLM-summarizer $\mathcal L$ ($\mathcal L:  D \rightarrow \mathbf S$). Let $\mathcal S \subseteq \mathbf{S}$ be a sample set of $k$ generated summaries, $\mathcal S = \{s_1, s_2 \ldots, s_k\}$ of $D$, each evaluated using metric $m$. Let, the corresponding  quality scores set be denoted by  $\mathcal{X} = \{x_1^m,x_2^m, \ldots, x_k^m\}, 0\le x_i^m \le 1$. The problem is to quantify the \textit{stability} of $\mathcal L$ by analyzing the  summaries in set $\mathcal S$.
    
     Our approach is based on the assumption that the \textit{difference} in the metric scores of two  text documents reflects \textit{variation} in the content. Thus, lesser variation among the summaries in $\mathcal S$ indicates higher \textit{stability} of the summarizer  for metric $m$. We exploit the differences in scores of $k$ summaries to  estimate \textit{stability} of  $\mathcal L$ with respect to the summary attribute measured by metric $m$, for $D$.  We define the the \textit{stability coefficient} below to recognize  the variation in scores as the characteristic of the \textit{stability}  of $\mathcal L$ for $D$ and metric $m$.

    \begin{defi}
    \label{def:doc-lvl-stab}
   The \textit{stability coefficient} $\epsilon_\mathcal L^m$ of an LLM-summarizer $\mathcal L$ for a document $D$ and metric $m$, is defined as $\epsilon_\mathcal L^m   = 1 - \Delta_\mathcal L^m $, where $ \Delta_\mathcal L^m  = \max\{ \delta_{ij}^m\}$, and $\delta_{ij}^m = |x_i^m - x_j^m|,  \forall \; 1 \le i < j \le k $.
\end{defi}
Here $\delta_{ij}^m$ denotes the absolute difference between the scores of the  summaries $s_i$ and $s_j$, and  $\Delta_\mathcal L^m$ denotes the maximum observed pairwise difference among the summary scores for $D$. Higher value of $\Delta_\mathcal L^m$ indicates higher \textit{variability} among the $k$ generated summaries, leading to lower \textit{stability} of the summarizer $\mathcal L$, for the document $D$ and metric $m$. It is noteworthy that $\epsilon_\mathcal L^m$ is a pessimistic estimate and hints  of the maximum observed variation among the summaries  in $\mathcal S$. The optimistic estimate, which advises about the least observed variation among the summaries, can be determined from the same set of scores. Naturally, the quality of estimate is influenced by the cardinality of the set. 

 We also compute pairwise similarity  scores among  $\binom{k}{2}$ summary pairs as an additional measure, to assess  semantic similarity among $k$ summaries. Higher average score indicates higher semantic similarity among the $k$ summaries. 
\subsection{Corpus-level stability} 
Assessment of the \textit{stability} of an LLM-summarizer for a document is of little information to the end-user and researchers, about its trustworthiness. Instead, knowledge about the stability of generated summaries for documents belonging to a specific genre may determine its utility for a specific use case. Our argument is based on the accepted fact that the behavior of an LLM-summarizer  varies with variations in  document type, length and writing style. 

Let $\mathcal C = \{ D_1, D_2, \ldots, D_N\} $ be a corpus of $N$ documents belonging to a specific genre $\mathcal{G}$. For each document $D_i$, a set of $k$ summaries $\mathcal S_i = \{s_{i1},s_{i2} , \ldots, s_{ik},\}$  generated by the LLM-summarizer $\mathcal L$.  The  score of summary $s_{ij}$, evaluated using metric $m$, is denoted by $x_{ij} = m(s_{ij})$, and the  scores corresponding to the $k$ summaries for document $D_i$ is denoted by  $\mathcal{X}_i = \{x_{i1},x_{i2} , \ldots, x_{ik},\}$, $0 \leq x_{ij} \leq 1$, and $1 \leq j \leq k$. The problem is to generalize  the stability of $\mathcal L$  for genre $\mathcal{G}$  by inspecting the variability in $N \times k$ scores scores in  $\{\mathcal{X}_1, \ldots, \mathcal{X}_N\}$.

Our approach is based on the premise that the \textit{stability} of an LLM-summarizer $\mathcal L$ for a corpus belonging to a specific genre $\mathcal{G}$ can be estimated by generalizing  the document-level stability over documents in $\mathcal C$. A summarizer that exhibits consistently high document-level stability across documents is expected to be more reliable than one whose stability varies substantially. Therefore, we aggregate the document-level stability coefficients in a principled manner to obtain a robust stability estimate for the LLM-summarizer $\mathcal L$.  Given $N$ documents in the corpus $\mathcal C$, we define the  stability coefficient of an LLM-summarizer below.
 \begin{defi}
\label{def:llm-lvl-stab}
   Given the stability coefficients $\{\epsilon^1_m, \epsilon^2_m, \ldots, \epsilon^N_m\}$ of an LLM-summarizer $\mathcal L$ for $N$ documents of a genre $\mathcal G$, and a metric $m$, the \textit{stability coefficient} of $\mathcal L$ for $m$ is defined as $\mathcal{E}_\mathcal L^m = min\{\epsilon^1, \epsilon^2, \ldots, \epsilon^N\}$.
\end{defi}
 Building upon $\epsilon_\mathcal L^m$, stability coefficient $\mathcal{E}_\mathcal L^m$ in Def.~\ref{def:llm-lvl-stab} is also a pessimistic estimate of the \textit{stability} of $\mathcal L$ for the metric $m$ over $N$ documents in the corpus $\mathcal{C}$. It is the barometer of variations that can be expected in $k$ summaries for a document of genre $\mathcal{G}$. For large $N$, the cost of estimating the coefficient may be computationally infeasible. Examining a representative sample of documents from a corpus, and ensuring statistical validity of the estimate is reasonable alternative.  Statistically large sample of documents ($N$) and summaries ($k$), may divulges the worst possible variations in summaries generated by $\mathcal{L}$.  We drop $m$ and $\mathcal L$ from the notation for brevity, as it will be clear by the context in the rest of the paper.

Additionally, we compute the confidence interval to quantify the variability or uncertainty associated with the summarizer performance. The confidence interval provides an indication of the degree of trust a user can place in the reported summary quality. A narrower confidence interval indicates greater confidence in the measured stability, whereas a wider interval suggests higher variability in the summarizer behavior across documents. 

The proposed approach is encapsulated in a diagnostic protocol proposed in Sec.~\ref{sec:stability-analysis-protocol}, which enables a statistically grounded assessment of the variability in summaries generated by an LLM-summarizer. The resulting stability coefficient and the corresponding interval estimate of the quality metrics provide a measure of the reliability and robustness of the summarizer, and serve as a quantitative evidence regarding the trustworthiness of the generated summaries.

\section{Methodology}
\label{sec:methodology}
The overarching objective of this research is to estimate the \textit{stability} of an LLM-summarizer for a document belonging to a specific genre. We define stability in terms of the consistency of summary quality across repeated generations for a set of documents. Specifically, we argue that variation in evaluation scores of the summaries over multiple dimensions is a significant pathology of generational variability inherent in LLM-summarizers. To investigate this, we employ a zero-shot setting to generate multiple summaries for each document, and a set of evaluation metrics to characterize variability in the generated summaries at the document-level. Subsequently, we analyze a variety of visual and descriptive measures to estimate \textit{stability coefficient} for each document. Finally, we aggregate these coefficients over the documents in the corpus or its representative sample, and obtain a pessimistic estimate of the overall stability of an LLM-summarizer along with the confidence interval. Figure~\ref{fig:methodology} illustrates the  methodology, and the details are presented in the following subsections.

\subsection{Summary Generation}
\begin{figure}
     \begin{tcolorbox}[width=0.95\linewidth, colback=gray!10,colframe=black,title=Summarization Prompt]
        \small
        You are an expert text summarizer.

        Your task is to read the provided document and produce a summary.
        
        Instructions:\\
        - Focus only on the main ideas and key information from the document. \\
        - Use clear, simple, and concise language. \\
        - Do NOT include any new information, assumptions, or personal opinions. \\
        - Ensure the meaning and intent of the original text are fully preserved. \\
        Limit the summary to <summary-length> words. 
        Output only the summary text, with no additional commentary.
        
        Document:<Document-Text>
        
        Summary:
    \end{tcolorbox}  
    \caption{Prompt for LLM-summarization.}
    \label{fig:summarization-prompt}
\end{figure}
We repeatedly generate summaries in a \textit{zero-shot} setting using an LLM-summarizer under static input environment. 
For a given document $D$,  $k$ summaries $\{s_1, \ldots, s_k\}$ are generated using a \textit{fixed} and \textit{structured prompt} (Fig.~\ref{fig:summarization-prompt}) and default model parameters, which ensures identical input conditions across $k$ runs. This controlled setting assures that observed variations arise solely from the inherent stochasticity of the generation process.

\subsection{Evaluation Metrics}
Repeatedly generated summaries by an LLM-summarizer for the same document ensconce variability across multiple facets. Though summary quality can be evaluated along several dimensions \citep{2025-zhang-survey},  we examine variability in two critical attributes of summary quality, viz., semantic alignment and factuality.
\begin{figure}
    \centering
    \includegraphics[width=0.95\linewidth]{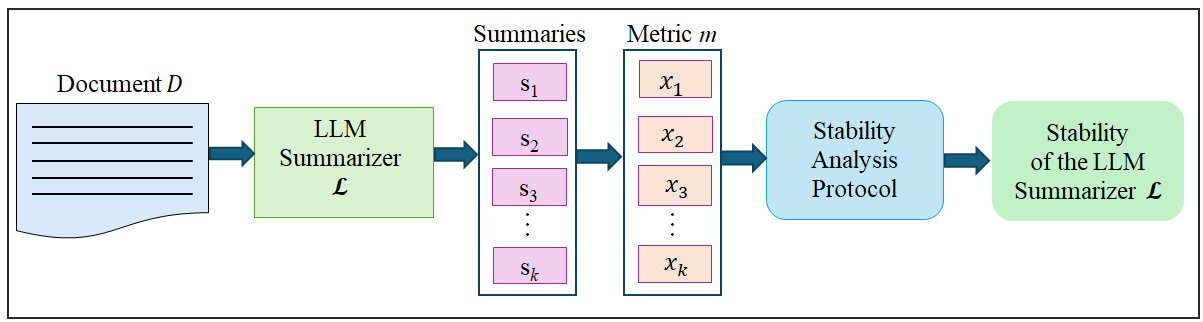}
    \caption{Pipeline for estimating stability coefficient of $\mathcal{L}$ for a document $D$ and a quality metric $m$.}
    \label{fig:methodology}
\end{figure}

We consider two dimensions for veracity. First, \textit{semantic alignment} to capture contextual similarity between each generated summary and the gold standard reference summary as assessed using BERTScore~\cite{2019-zhang-bertscore}. Difference in the BERTScore of the two summaries informs semantic variation between them. The other dimension is factual faithfulness, which measures the factual alignment between the summary and the source document using AlignScore \cite{2023-zha-alignscore}. Difference in the AlignScore of two summaries captures the difference in the extent of misalignment of two summaries. The study of variability can be trivially expanded for other summary attributes using appropriate metrics \cite{2025-zhang-survey}. Thus, the stability of an LLM-summarizer can be estimated for the chosen evaluation metrics based on the corresponding scores of $k$ summaries as shown in the pipeline (Fig.~\ref{fig:methodology}).

Next, we assess the semantic consistency among the generated summaries, which indicates the extent of agreement among them. We compute the Semantic Consistency (SC) score for document $D$ by measuring the semantic similarity among all $\binom{k}{2}$ pairs of generated summaries using cosine similarity, as defined below.
    \begin{equation}
    \label{eq-sc-score}
        SC_{D} = \frac{2}{k(k-1)} \sum_{i<j} \cos(s_i, s_j), \; \forall  \; 1 \le i < j \le k,
    \end{equation}
where, $\mathbf{s}_i$ and $\mathbf{s}_j$ are summaries, and $k$ is the number of generated summaries.  A higher SC score reveals greater similarity (i.e., higher semantic consistency), whereas a lower SC score indicates greater variability among $k$ summaries. 

\subsection{Stability Analysis}
Analyses are performed at two levels of granularity. First,  document-level analysis is performed by analyzing the underlying distributional behavior of the metric  for the $k$  summaries. At this level, we examine the generational variability of the LLM-summarizer ($\mathcal{L}$ ) through a set of complementary and mutually reinforcing statistical measures and tests.  Subsequently, we estimate the \textit{stability coefficient} for the document. At the corpus-level, we build on the document-level analysis and draw inferences about the \textit{stability} of $\mathcal{L}$ for genre $\mathcal{G}$, using another set of interrelated statistical measures and tests. This two stage analysis leads to an elegant and efficient \textit{Stability Analysis Protocol}, which is consistent and reproducible. The proposed protocol is  described in the next section, and demonstrated in Sec.~\ref{subsec:result-benchmark-llm-summarizer} for benchmarking LLM-summarizers. 


%
\section{Stability Analysis Protocol}
\label{sec:stability-analysis-protocol}
The proposed \textit{Stability Analysis Protocol} (SAP) is a structured framework for diagnosing and  quantifying  the \textit{stability} of an LLM-summarizer $\mathcal{L}$ for genre $\mathcal{G}$.  The protocol design is  based on the hypothesis that the \textit{variations} in the scores of summary  attributes (e.g. relevance, fluency, factuality, etc.) reflect the \textit{variability} in the summary texts in an unbiased fashion.  High variability among the scores of a metric indicates lower \textit{stability} of the model. 
\begin{figure}
    \centering
    \scriptsize
    \includegraphics[width=0.95\linewidth]{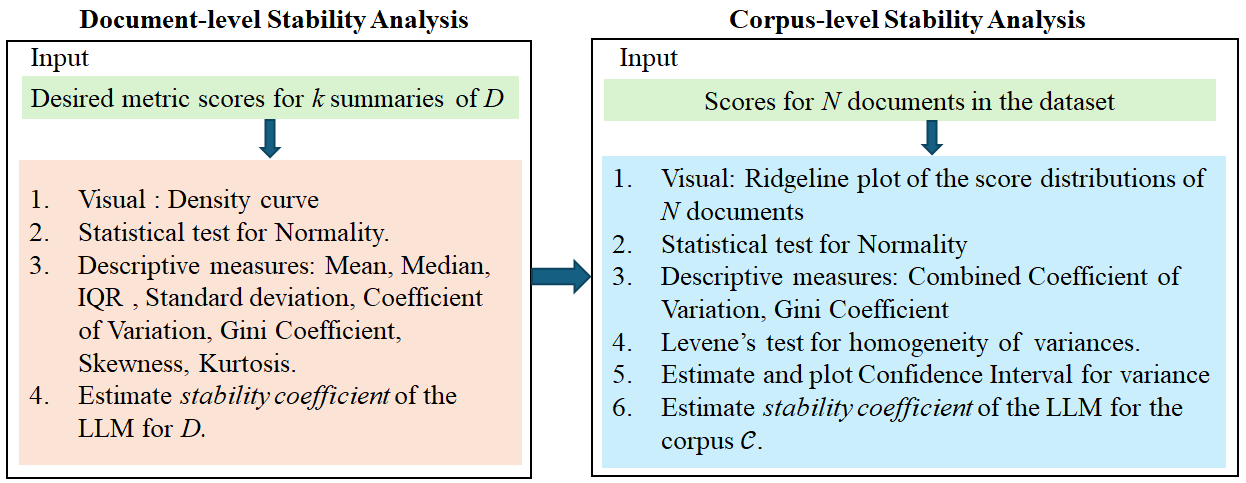}
    \caption{Two stages of the proposed Stability Analysis Protocol.}
    \label{fig:stabilty-analysis-protocol}
\end{figure}

We investigate the generational variability of the LLM-summarizer at two levels using a suite of complementary and interrelated statistical measures and hypothesis tests, thereby enabling a comprehensive assessment of its consistency. At  first-level, we analyse the stability of the document summaries by collating the variability among the scores of the $k$ generated summaries, and estimate its \textit{stability coefficient} $\epsilon$. At corpus-level, we analyze variability in summaries of $N$ documents in the corpus and investigate combined variability using appropriate exploratory and descriptive measures. We further estimate the stability coefficient $\mathcal{E}$ for summarizer $\mathcal{L}$ and compute the confidence interval for the variability observed in its generated summaries.
At both levels, we use a judicious mix of visual techniques for preliminary exploration, descriptive statistics and statistical tests as confirmatory tools. The proposed protocol is illustrated in Fig.~\ref{fig:stabilty-analysis-protocol}, and its two stages are described below. 

\subsection{Document-level Stability Analysis}
At document-level, we explore the distributional characteristics of summary scores corresponding to the chosen evaluation metric. The generational variability revealed by plotting the score distribution offers a glimpse of variations in summaries for document $D$. Subsequently, we quantify the observed variability by computing \textit{ measures of central tendency} (mean and median), \textit{dispersion} (standard deviation, inter-quartile range and coefficient of variation), and \textit{higher-order distributional properties} (skewness, kurtosis, and Gini coefficient). Though the measures of central tendency indicate the quality of the generated summaries, we maintain focus on the dispersion  measures to capture the variability in the scores. Skewness and kurtosis quantify the asymmetry and peakedness of a distribution, respectively. The Gini coefficient (GC) measures the degree of inequality in the score distribution $\{x_1, x_2, \dots, x_k\}$ of $k$ summaries as follows.
\begin{equation*}
    \text{GC} = \frac{1}{2k^2 \bar{x}} \sum_{i=1}^{k} \sum_{j=1}^{k} |x_i - x_j|, \mbox{ where } \bar{x} = \frac{1}{k}\sum_{i=1}^{k} x_i
\end{equation*}
Zero value of Gini coefficient indicates uniformity of scores and reflects perfect document-level \textit{stability} in summaries, while a higher value indicates greater variability in the score distribution. The GC, along with the other descriptive statistics, provides a comprehensive characterization of document-level stability of the LLM. 

Additionally, we estimate the stability coefficient $\epsilon$  as per Def.~\ref{def:doc-lvl-stab}, by finding the maximum pairwise absolute difference in metric scores of $k$ summaries for the document $D$. Note that higher value of $\epsilon$ indicates higher stability among the repeatedly generated summaries for the chosen evaluation metric.
\subsection{Corpus level Stability Analysis}
Leveraging document-level analysis, we extend our evaluation to the corpus-level for investigating the variability in summaries generated by $\mathcal L$.  First, we employ \textit{Ridgeline plots} of the score distributions for documents in the corpus to visually reveal candid differences in their  distributional behavior. We also examine \textit{normality} of the distributions using the Shapiro{-}Wilk test \cite{2011-yap-stats-test} to  guide us to choose suitable statistical tests for subsequent analysis. To quantitatively assess score dispersion at the corpus-level, we compute the combined coefficient of variation ($CV_c$) over the summary scores of all documents.

Let the corpus consist of $N$ documents \{$ D_1, D_2, \ldots, D_N $\} of genre $\mathcal{G}$. Correspondingly, there are $N$ sets of $k$ LLM-generated summaries, which are evaluated using metric $m$ .  Now, considering set $\mathcal{X}_i = \{x_{i1},x_{i2}, \ldots, x_{ik}\}$  of summary scores corresponding to summary set $S_i$ as a group, we compute combined coefficient of variation ($CV_c$) for $kN$ summaries. This measure reveals overall heterogeneity of the system, while incorporating both stochastic dispersion \textit{within the group} (i.e., variation between $k$ summaries of individual document) and structural heterogeneity \textit{among $N$ groups} (i.e., variations due to documents).   
\begin{equation*}
\label{equation:-cvc}
    CV_c = \frac{\sigma_c}{\mu_c}, \; where  \; \mu_c = \frac{1}{N}\sum_{i=1}^{N}\bar{x}_i,  \; and \; 
    \sigma_c^2
        =
        \frac{1}{N}
        \sum_{i=1}^{N}
        \left[
        \sigma_i^2 + (\bar{x}_i-\mu_c)^2
        \right]
\end{equation*}
Here, \(\bar{x}_i\) and \(\sigma_i\) denote mean and standard deviation of $k$ summary scores for the $i^{th}$ document, respectively. $\mu_c$ denotes the pooled mean of $kN$ scores and $\sigma_c^2$ denotes the pooled variance\footnote{For derivation see~\ref{appendix-cvc-derivation}}. Combined coefficient of variation is an important and useful measure for benchmarking the stability of LLM-summarizers.

Next, we investigate the second-order variation in the generated summaries. Specifically, we check  \textit{whether the extent of variations in the summaries for different documents is significant}. As a preliminary indicator of the document-level differences in the variability of generated summaries, we visually compare the Gini coefficients of $N$ documents in the corpus.  Subsequently, we apply Levene's test~\citep{1974-levenes-test-brown} at significance level $\alpha$ for homogeneity of variances among the $N$ summary sets $S_i$'s to assess if the variability of summary scores in $\mathcal{X}_i$ differs significantly across the documents in the corpus. If $\sigma_1^2, \sigma_2^2, \ldots, \sigma_N^2$ denote the variances of the respective groups, the test investigates the null hypothesis,\\ \textit{$H_0$: $\sigma_1^2 = \sigma_2^2 = \ldots = \sigma_N^2$}, against the alternative hypothesis,\\ \textit{$H_1$: $\sigma_i^2  \neq \sigma_j^2 $, for at least one pair (i, j)}. 

If the null hypothesis of Levene's test is accepted at significance level $\alpha$, it implies that the variability in summaries is statistically non-significant across the documents of genre $\mathcal G$.  Rejection of $H_0$ indicates that the variability of at least one pair of summary sets differs significantly. In this case, the question that needs to be answered is, \textit{``What is the  range of variability that a user can expect while generating summaries using $\mathcal L$?''}. To answer this question,  we estimate bootstrap confidence interval for variance of the summary scores for the summarizer $\mathcal{L}$ and genre $\mathcal G$. The confidence interval provides an indication to the users that how much variability they can expect in the generated summaries. A narrow confidence interval invokes higher trust in the generated summaries. 

The  next step in the protocol is to estimate the \textit{stability} coefficients $\{\epsilon^1, \epsilon^2, \ldots, \epsilon^N\}$ for all documents and metric $m$. Each  $\epsilon^i$ represents the pessimistic estimate (lower bound) of the stability among $k$ repeatedly generated summaries of document $D_i$. Finally, we estimate \textit{stability coefficient} $\mathcal E$  of the summarizer $\mathcal L$  as per Def.~\ref{def:llm-lvl-stab}. Lower value of $\mathcal E$ indicates that the \textit{LLM-summarizer}  produces summaries that are not \textit{stable}, and hence is not trustworthy. 
\subsection{Benchmarking LLM Summarizers}
\label{sec:benchmarking-LLM}
Given the set of  scores for a summary evaluation metric $m$, the proposed protocol delivers clear, consistent and reproducible results for assessing stability of an LLM-summarizer. Therefore, it can be used for benchmarking two or more LLM-summarizers by comparing their \textit{stability coefficients} $\mathcal{E}$'s for the corpus of genre $\mathcal{G}$. 

Document-wise ranked heatmap of the metric scores for the compared LLM-summarizers provides the first glimpse of their relative performances. Considering the highest rank to be a \textit{win}, the \textit{win/loss} count reflects how often a summarizer outperforms or underperforms the other summarizers, indicating its competitive performance. LLM-summarizers are then ranked according to \textit{win/loss} count and their \textit{average} ranks are computed as an additional indicator of their comparative performance. Further, we perform non-parametric \textit{Friedman} test ~\citep{2006-demvsar-statistical} to statistically validate the differences in the average ranks, and to draw robust inference about the relative stability of the LLM-summarizers under consideration. The null hypothesis, \\
\textit{$H_0$: The competing LLM-summarizers perform equally}, is tested against the alternative, \\ 
\textit{$H_1$: The performance of summarizers differs significantly}. \\ 
In case, $H_0$ is rejected at significance level $\alpha$, we perform pairwise comparison of models using \textit{Nemenyi} post-hoc test to identify which summarization models differ significantly from one another. Two models are significantly different if the difference in their average ranks exceeds the critical difference (CD) at significance level $\alpha$. Finally, we compare the \textit{stability coefficient}  ($\mathcal{E}$) to benchmark the LLM-summarizers. 
 
\section{Experimental Design}
\label{sec:expt-design}
The objective of the experimental design is to demonstrate the use of \textit{Stability Analysis Protocol} (SAP) for  benchmarking  LLM-summarizers. The pipeline and evaluation strategy described in Sec.~\ref{sec:methodology} are implemented in Python 3.12 using PyTorch 1.13.1+cu117, integrating multiple LLMs. All experiments are carried out on a workstation configured with an Intel(R) Xeon(R) Gold 6226R CPU (2.90GHz), an NVIDIA RTX A6000 GPU, and 256 GB RAM, deployed on Ubuntu 22.04.2 LTS.  We describe below the datasets and LLMs used in this study. 

\subsection{Datasets}
\label{sec:datasets}
To estimate stability of an LLM-summarizer, we assess summary variability across diverse content types, we experiment with documents spanning three genres: news, medical, and legal domain. These genres differ substantially in writing style, document length and structure, vocabulary, and informational density.

For the news domain, we use the CNN/DM dataset, a widely adopted benchmark featuring multi-sentence articles with reference summaries \cite{CNN-Dailymail-2016}. The PubMed dataset consists of long-form scientific articles paired with structured abstracts, represents medical domain \cite{pubmed-2018}. For the legal domain, we use IN-Abs dataset, comprising Indian Supreme Court case documents paired with abstractive headnotes that serve as expert-authored summaries \cite{IN-Abs-2022}. The documents in PubMed and IN-Abs datasets pose challenges related to input length and domain-specific terminology.
\begin{table}[width=.56\linewidth,cols=4,pos=h]
\caption{Average document and summary length (in words) for the corpus and the sample sets. Standard deviations are shown alongside.}
\label{tab:datasets-stats}
\begin{tabular*}{\tblwidth}{l c c c}
    \toprule
    \textbf{Datasets} & \textbf{Docs} & \textbf{Document} & \textbf{Summary} \\ \midrule
     \multicolumn{4}{c}{\textbf{Corpus}}\\ \midrule
     \textbf{CNN/DM}  & 2,87,113 & 691.87 $\pm$ 336.50& 51.57 $\pm$ 21.26 \\
        \textbf{PubMed }  & 1,19,224 & 3044.24 $\pm$ 2455.29 & 202.24 $\pm$ 78.23 \\
        \textbf{IN-Abs}  & 7128 & 4375.17 $\pm$ 5118.56& 841.28 $\pm$ 904.86 \\ \midrule
        \multicolumn{4}{c}{ \textbf{Sample Set (32 documents)}}\\ \midrule
        \textbf{CNN/DM } & & 661.06 $\pm$ 286.80 & 57.66 $\pm$ 24.24 \\
        \textbf{PubMed } & & 3183.13 $\pm$ 2399.45& 204.13 $\pm$ 81.62 \\
        \textbf{IN-Abs}  & & 4203.22 $\pm$ 3754.15& 879.97 $\pm$ 466.56 \\
    \bottomrule
\end{tabular*}
\end{table}
Generating multiple summaries for all documents in large datasets is highly resource intensive. Limited budget for computational resources necessitates sampling of documents for the study. Since input length can influence model behavior, especially under fixed context window that may require truncation or chunking, we adopt a \textit{length-aware} document selection strategy to draw a statistically large and representative sample of documents from the corpus. In order to ensure balanced coverage across the full length spectrum of the dataset, including short, medium, and long inputs, we stratify each dataset into four quartiles based on document length, and randomly sample eight documents from each quartile. Table~\ref{tab:datasets-stats} shows the statistics of the corpora (populations) and respective sample sets.
\begin{table}[width=.45\linewidth,cols=4,pos=h]
\caption{Z-test \textit{p-values} for difference in mean document length.}
\label{tab:difference-of-means}
    \begin{tabular*}{\tblwidth}{l r l c}\toprule    
        \textbf{Dataset}&\textbf{Z-score}& \textbf{p-value}& \textbf{Result}\\\midrule
        CNN/DM &  -0.5170& 0.6052 & Not Significant \\
        PubMed &  0.3190& 0.7497 & Not Significant \\
        IN-Abs &  -0.1900& 0.8493 & Not Significant \\ 
        \bottomrule
    \end{tabular*}
\end{table}

To check the \textit{representativeness} of the sample, we  test for difference in the means and variances of the document lengths in the samples and the respective populations (original datasets). We use Z-test for difference of means, under the null hypothesis, \\ 
$H_0$: \textit{There is no significant difference between the mean document lengths in the sample and population}, is tested against the alternative,\\
\textit{$H_1$: Sample and population means are different}. \\
The resulting \textit{p-values} indicate that we may accept $H_0$ at $5$\% level of significance for all three cases (Table~\ref{tab:difference-of-means}), implying that the mean length of the  documents in samples is representative of their respective populations. 
\begin{table}[width=.45\linewidth,cols=4,pos=t]
\caption{$\chi^2$-test \textit{p-values} for difference in variance.*: \textit{p-values} after removing outliers. }
\label{tab:diff-of-std}
 \begin{tabular*}{\tblwidth}{l c c c}
    \toprule    
    \textbf{Dataset} & $\chi^\textbf{2}$ & \textbf{p-value} & \textbf{Result} \\\midrule
    CNN/DM & 22.5199 & 0.2680 & Not Significant \\
    PubMed & 29.6063 & 0.9246 & Not Significant \\
    IN-Abs&  16.6783 & 0.0337 & Significant \\
    IN-Abs*& 27.6317  & 0.9247 & Not Significant \\
     \bottomrule
\end{tabular*}
\end{table}

To ensure that the sample variances  are not significantly different from  the respective  population variances, we perform $\chi^2$ test. The null hypothesis,\\ 
\textit{$H_0$: There is no significant difference between the variances of the document lengths in the sample and population}, is tested against the alternative, \\
\textit{$H_1$: Sample and population variances are different}.\\
The \textit{p-values} of the test indicate that we may accept $H_0$ at $5\%$ level of significance for CNN/DM and PubMed datasets (Table~\ref{tab:diff-of-std}). However, the difference is  statistically significant in the case of IN-Abs sample dataset at $5\%$ level of significance. Deeper examination of the length distribution of the sample and population reveals presence of \textit{outliers} in both sets (see Fig.~\ref{fig:app-doc-length-distribution-plot} in ~\ref{sec:app-dataset}). After excluding the outliers from the sample and population, the chi-square test yielded \textit{p-value} which favors acceptance of $H_0$, indicating no significant difference between their variances at $5\%$ level of significance. We re-checked for the difference in means, and found that no statistically significant difference exists in mean document lengths after removing outliers. 
\subsection{Language Models and Summary Generation}
\label{subsec:llms}
We use publicly available LLMs with fewer than 4B parameters as summarizers. Their reduced computational footprint makes them particularly suitable for systematic analysis of variability in text generation for repeated runs under limited computational resources. 
We examine Gemma3 (4B) Instruct\footnote{\scriptsize\url{https://huggingface.co/google/gemma-3-4b-it}}, Llama-3.2 (3B) Instruct\footnote{\footnotesize \url{https://huggingface.co/meta-llama/Llama-3.2-3B}}, and Qwen-2.5 (3B) Instruct\footnote{\footnotesize\url{https://huggingface.co/Qwen/Qwen2.5-3B-Instruct}} models. 


For each document in the sample, the model is prompted to generate $k=100$ summaries using the prompt shown in Fig.~\ref{fig:summarization-prompt}. These summaries represent statistically large random sample from the space ($\mathcal S$) of all possible summaries of the document, and form the basis of \textit{stability} analysis of the LLM-summarizers.  With $32$ documents sampled per genre, the resulting corpus comprises $3200$ summaries which are evaluated for three metrics. Thus, the three sets of $3200$ metric scores provide a statistically reliable basis for estimating \textit{stability coefficient} ($\mathcal{E}$) of the LLM-summarizer.
\section{Empirical Results}
In this section, we report the findings of the empirical study based on \textit{Stability Analysis Protocol}. All reported scores are aggregated over the set of hundred summaries for each document.
\begin{figure}
    \centering
        \begin{subfigure}{0.32\textwidth}
            \centering
            \includegraphics[width=\textwidth, height=3.1cm]{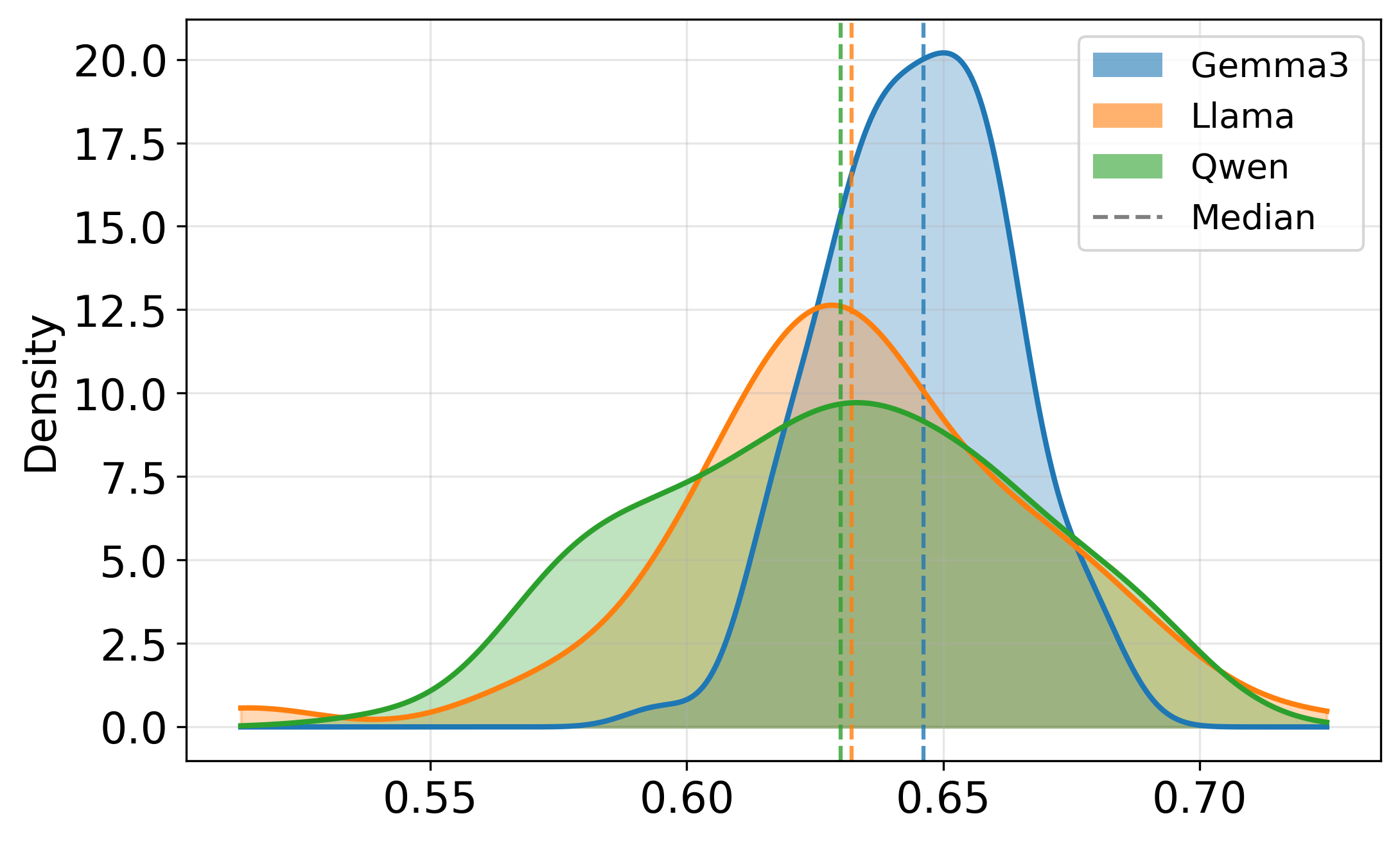}
            \caption{BERTScore}
        \end{subfigure}
        \hfill
        \begin{subfigure}{0.32\textwidth}
            \centering
            \includegraphics[width=\textwidth, height=3.1cm]{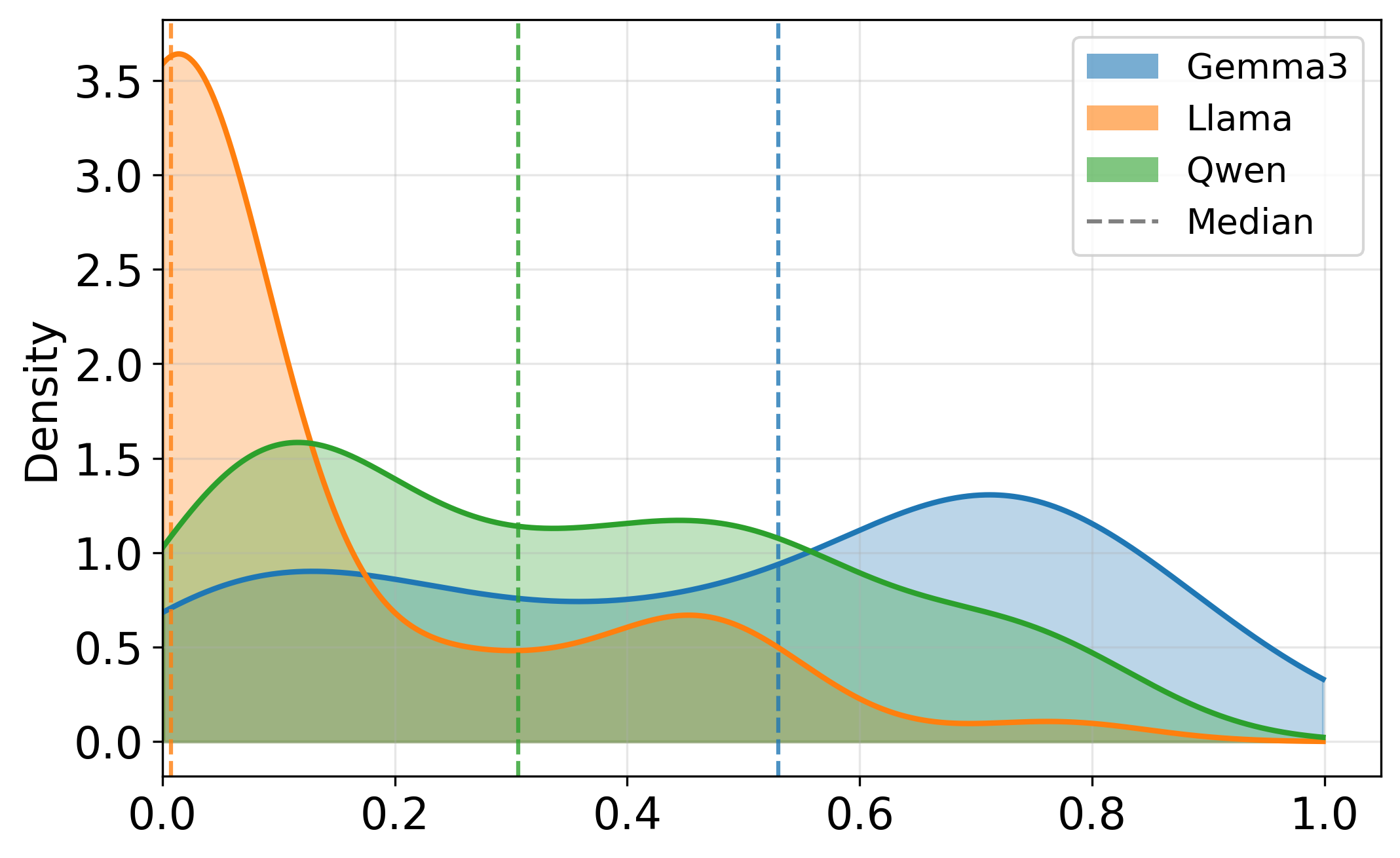}
            \caption{AlignScore}
            \label{fig:density-plot-alignScore}
        \end{subfigure}
        \hfill
        \begin{subfigure}{0.32\textwidth}
            \centering
            \includegraphics[width=\textwidth, height=3.1cm]{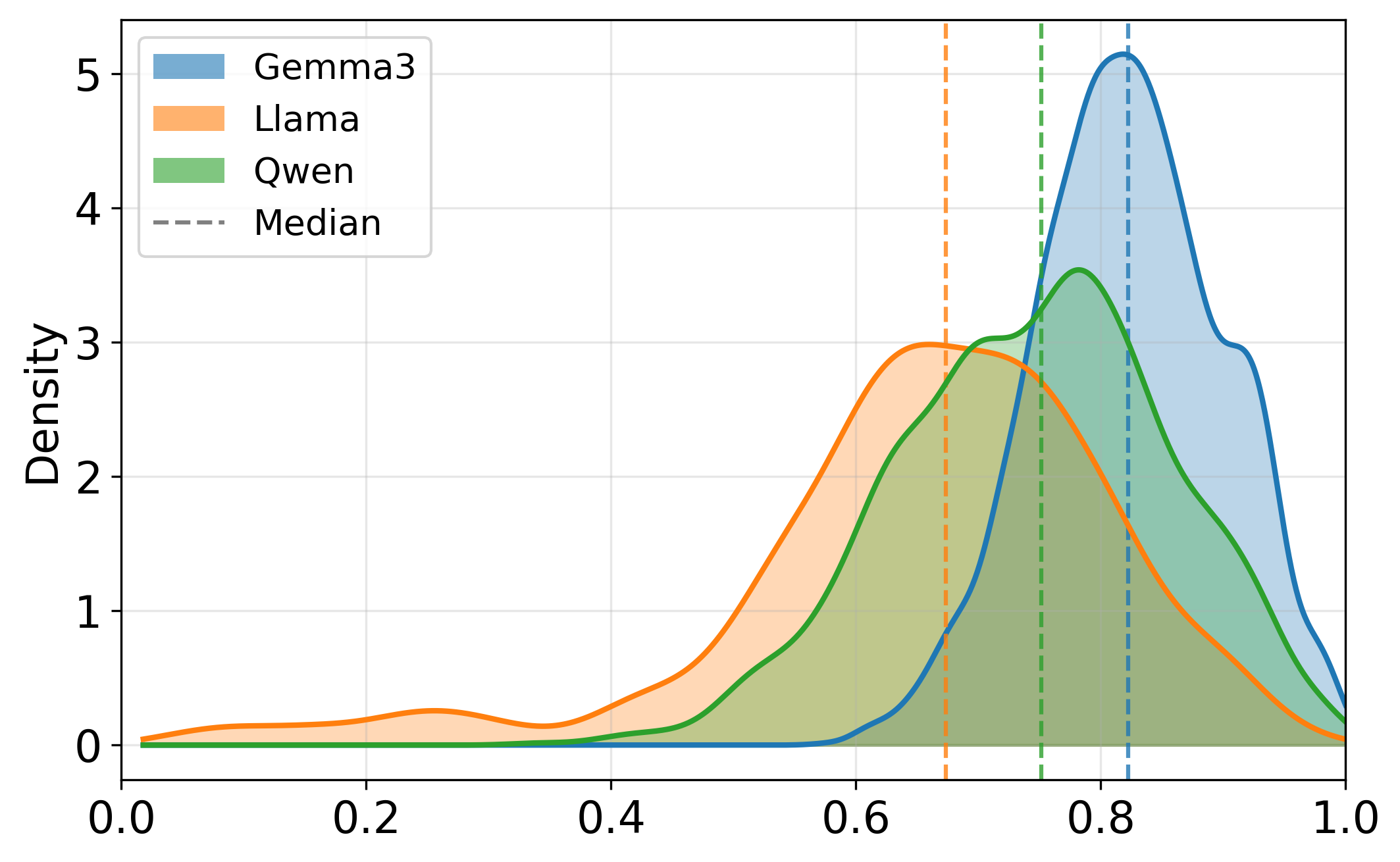}
            \caption{SC Score}
        \end{subfigure}
        \caption{Density plots of $100$ summaries scores generated by three LLMs for the selected CNN/DM article.}
        \label{fig:kde-plots}
\end{figure}   
\subsection{Results for Document-level Stability }
To exemplify the document-level analysis, we select a document\footnote{\footnotesize Document-id: doc\_424 was taken from the sample.} from the CNN/DM sample set and generate $100$ summaries using three LLMs under consideration. We compute BERTScore and AlignScore for each generated summary, and Semantic Consistency (SC) score for each summary set generated by the three summarizers. Following the proposed \textit{Stability Analysis Protocol} (SAP), we visualize the results and compute the relevant statistics to assess the stability of the summarizers for this document. 

\noindent
\textit{Visual Analysis}: Figure~\ref{fig:kde-plots} shows the density plots of the three scores of generated summaries for the chosen document. The dashed vertical lines indicate the median score for each distribution. The distributions are mostly asymmetric indicating  deviations from normality for  all metrics. Highest medians for BERTScore and AlignScore for Gemma3 summaries hint that the overall quality (semantic alignment and factual correctness) is better than  other two sets of LLM-summaries. The highest median SC score achieved by Gemma3 indicates greater  semantic consistency among the generated summaries than the other LLMs. Additionally, the sharply peaked distributions of its BERTScore values and SC scores suggest more stable and consistent performance than the other two models. Among the three metrics, the distributions for AlignScore values are discernible because of distinctly high variability for all LLM-summarizers.  Despite extreme swings in AlignScore values of the generated summaries by three models, Gemma3 maintains the highest central alignment, which is evident by the medians (dashed lines). This indicates Gemma3 summaries are best grounded relative to the source document. Contrastingly, the peak for Llama distribution lies towards extremely low values (close to zero), which reveals minimal factual alignment with the source documents, and reveals weakest \textit{stability}. 
 
\begin{table}[width=.65\linewidth,cols=7,pos=h]
    \centering
     \caption{Descriptive statistics for the  score distributions corresponding to the hundred LLM-summaries of the selected document  from CNN/DM sample set. Bold values represent the best score. $\uparrow$: Highest is best, $\downarrow$: Lowest is best, CV: Coefficient of Variation, GC: Gini Coefficient.}
    \label{Tab:stats-for-one-doc}
     \begin{tabular*}{\tblwidth}{l|c c |c c c c}
        \toprule            
                \textbf{LLM} & \textbf{Mean $\uparrow$} & \textbf{Median $\uparrow$} &  \textbf{IQR $\downarrow$}&\textbf{Std $\downarrow$} & \textbf{CV $\downarrow$} & \textbf{GC $\downarrow$} \\ \midrule
                \multicolumn{7}{c}{\textbf{BERTScore}}\\ \midrule
                \textbf{Gemma3} & \textbf{0.6450}&\textbf{ 0.6461}&  \textbf{0.0245}&\textbf{0.0172}& \textbf{0.0267}&\textbf{ 0.0151} \\
                \textbf{Llama} & 0.6331& 0.6321&  0.0395& 0.0348& 0.0550& 0.0299\\
                \textbf{Qwen} & 0.6290& 0.6300&  0.0529&0.0353& 0.0562& 0.0322\\ \midrule
                
                \multicolumn{7}{c}{\textbf{AlignScore}}\\ \midrule
                \textbf{Gemma3}&\textbf{ 0.4787}& \textbf{0.5256}&  0.5123&0.2917& \textbf{0.6095}& \textbf{0.3488} \\
                \textbf{Llama } & 0.1206& 0.0076&  \textbf{0.1512}& \textbf{0.1897}& 1.5729& 0.7303\\
                \textbf{Qwen} & 0.3332& 0.3063&  0.3836&0.2346& 0.7039& 0.4014\\  \midrule
                 \multicolumn{7}{c}{\textbf{SC Score}}\\ \midrule
                \textbf{Gemma3} &\textbf{ 0.8231}&\textbf{ 0.8227 } & \textbf{ 0.1041}&\textbf{0.0752}& \textbf{0.0913}& \textbf{0.0519}  \\
                \textbf{Llama } & 0.6551& 0.6735&  0.1702& 0.1593& 0.2432& 0.1288\\
                \textbf{Qwen} & 0.7443& 0.7513&  0.1541&0.1119& 0.1504& 0.0852\\ 
        \bottomrule
    \end{tabular*}
\end{table}
\noindent
\textit{Descriptive measures}: The descriptive measures corresponding to the score distributions for the same document are reported in Table~\ref{Tab:stats-for-one-doc}. Gemma3 exhibits highest mean and median scores for all three metrics, which indicate that the  generated summaries are qualitatively the best among the three summary sets. Extremely low mean and median for AlignScore values for Llama summaries  reflect poor factual alignment, which match the corresponding distribution plot. Dispersion measures, the focus of this study, are the lowest for Gemma3 summaries for both semantic  metrics (BERTScore and SC Score) and in consonance with the density plots. Lower values of both the coefficient of variation (CV) and Gini coefficient (GC)  confirm  better consistency of scores for Gemma3 summaries for all three metrics. The least IQR and standard deviation of AlignScore values for Llama summaries (Table~\ref{Tab:stats-for-one-doc}) are deceptive due to high concentration of data points towards lower end of scale. This is also evident by the corresponding distributional characteristics (Fig.~\ref{fig:kde-plots}) and high values of CV and GC (Table~\ref{Tab:stats-for-one-doc}). 
\begin{table}[width=.38\linewidth,cols=3,pos=h]
    \centering
    \caption{Metric-wise stability coefficient ($\epsilon$) of three LLMs for a document from CNN/DM.}
    \label{tab:cnn-dm-424-epsilon-stability}
    \begin{tabular*}{\tblwidth}{lcc}
    \toprule
         \textbf{LLM}&  \textbf{BERTScore}& \textbf{AlignScore}\\ \midrule
         \textbf{Gemma3}&  \textbf{0.9118}  & 0.0058 \\
         \textbf{Llama}&  0.7883 & \textbf{0.2185} \\
         \textbf{Qwen}&  0.8436& 0.1903 \\ 
         \bottomrule
   \end{tabular*}
\end{table}

\noindent
\textit{Stability Coefficient}: Table~\ref{tab:cnn-dm-424-epsilon-stability} shows the estimated \textit{stability coefficients} of the three LLM-summarizers for the selected document, and quality metrics. We humbly remind the reader that the numbers in the table are \textit{not} the indicators for summary quality. Instead, these values indicate the stability among hundred summaries of the document under scrutiny, with higher stability indicated by higher $\epsilon$ values. It is observed from the Table~\ref{tab:cnn-dm-424-epsilon-stability} that Gemma3 displays the most \textit{stable} behavior for BERTScore, Qwen shows moderate stability, and Llama exhibits least stability. Note that the \textit{stability coefficient} for AlignScore is abysmally low for all summarizers, with Gemma3 being the lowest. Though Llama exhibits better stability than the other two models, AlignScore values for the hundred summaries are very weak, as is evident from the low \textit{mean} and \textit{median} (Table \ref{Tab:stats-for-one-doc}). Thus summaries generated by Llama-summarizer for this document are not only qualitatively weak, but  also exhibit  greater variability along the semantic and factual dimensions. 

\textit{Based on the BERTScore stability coefficient, we conclude that Gemma3 generates most stable summaries, whereas Llama generates the least stable summaries for the chosen document. Furthermore, the results also indicate that irrespective of the summarizer, BERTScore is the more stable metric than AlignScore.}
\subsection{Results for Corpus-level Stability}
\label{subsec:dataset-level-stability-analysis}
\begin{figure}
    \centering
    \begin{subfigure}[b]{0.32\textwidth}
        \centering
        \includegraphics[width=\textwidth]{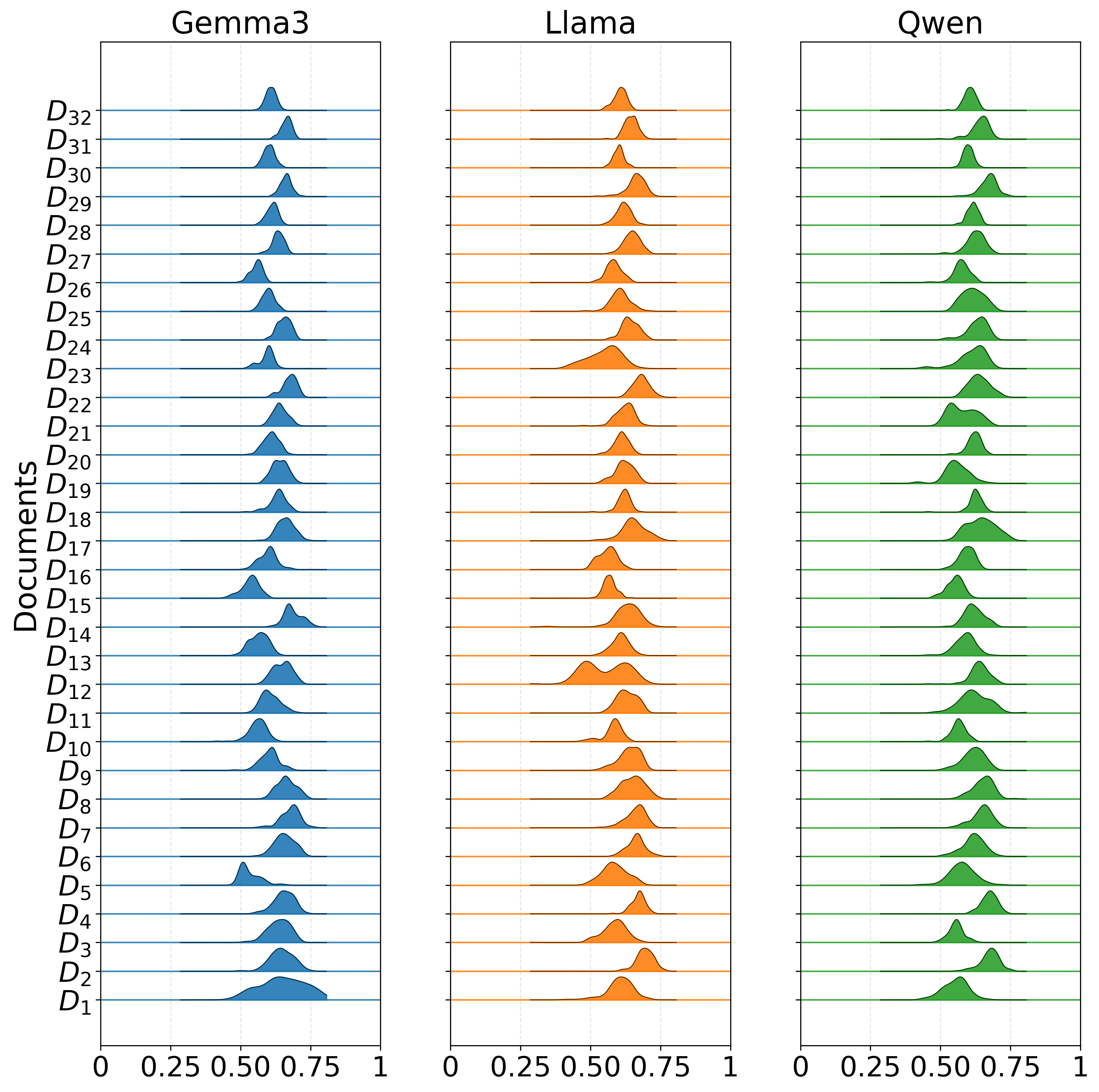}
        \caption{BERTScore}
        \label{fig:cnn-dm-bertscore-ridge}
    \end{subfigure}
    \hfill
    \begin{subfigure}[b]{0.32\textwidth}
        \centering
        \includegraphics[width=\textwidth]{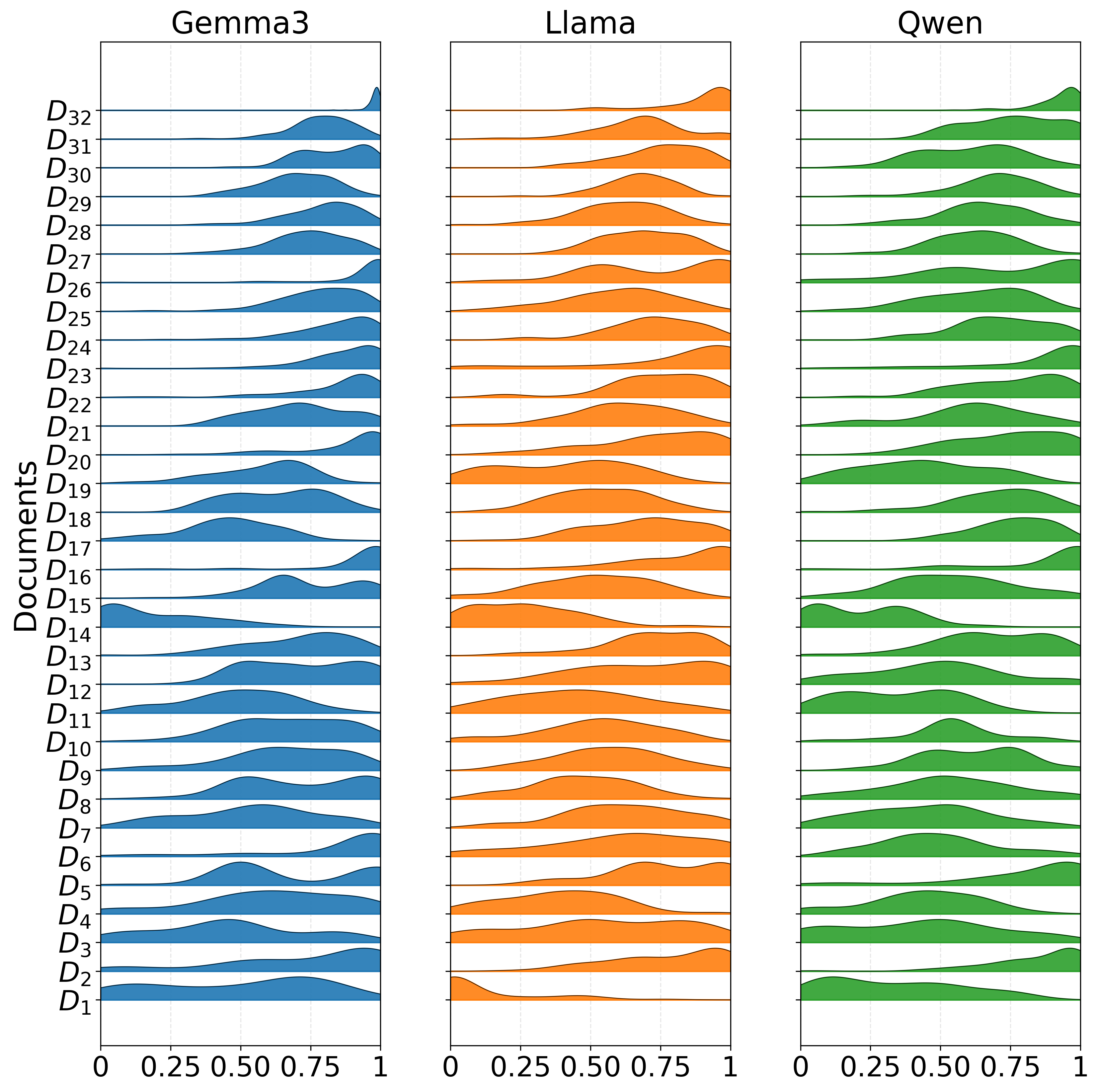}
        \caption{AlignScore}
        \label{fig:cnn-dm-alignscore-ridge}
    \end{subfigure}
    \hfill
    \begin{subfigure}[b]{0.32\textwidth}
        \centering
        \includegraphics[width=\textwidth]{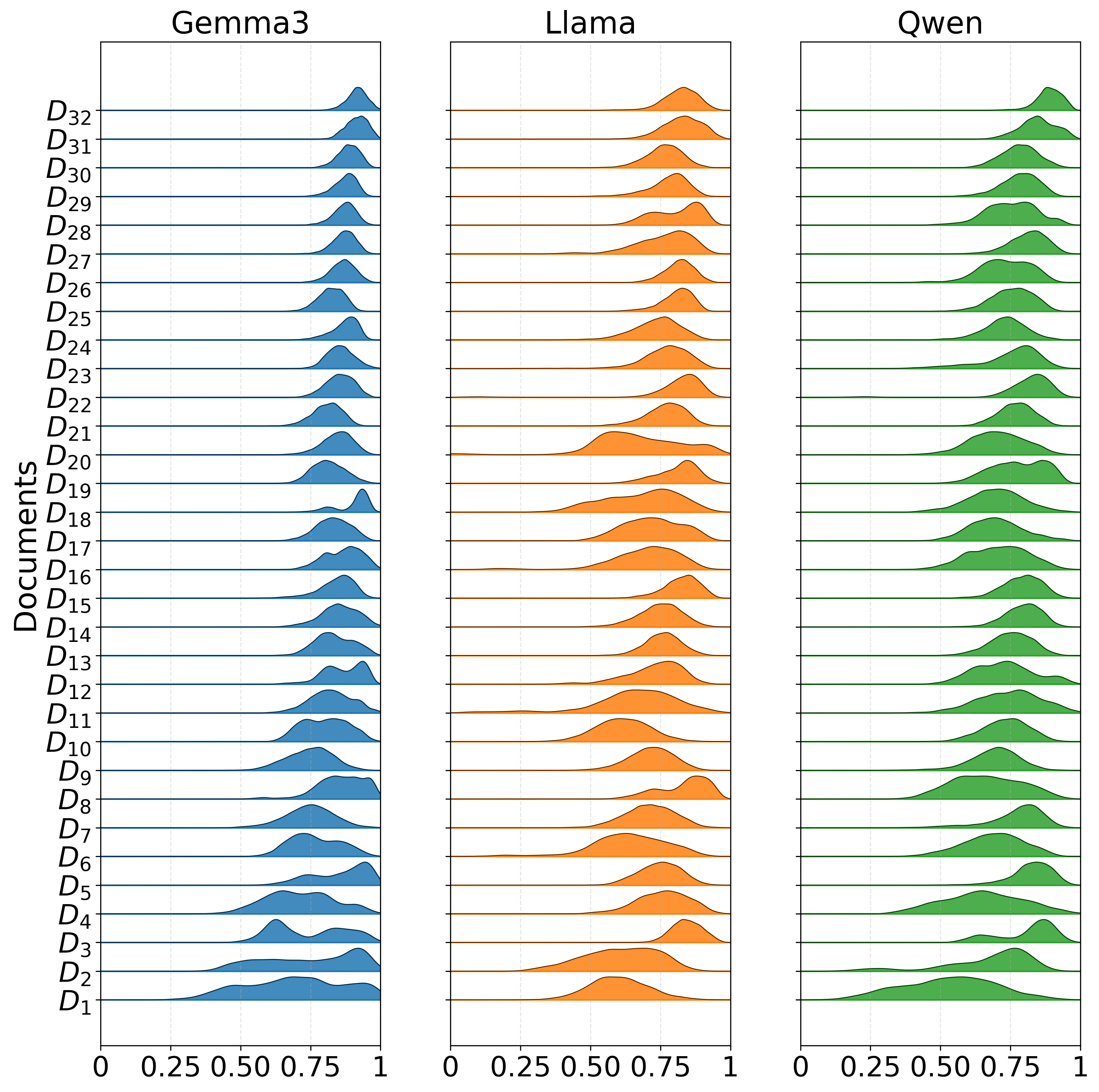}
        \caption{SC Score}
         \label{fig:cnn-dm-scscore-ridge}
    \end{subfigure}
    \caption{Ridgeline plots of the three score distributions of Gemma3, Llama, and Qwen generated summaries for the CNN/DM sample dataset of $32$ documents.}
    \label{fig:cnn-dm-ridge-plots}
\end{figure}
The corpus-level analysis, the second stage of SAP,  is elucidated using documents from three corpora that represent three different genres (Sec.~\ref{sec:datasets}). As per the protocol, visual plots depict the broad trends, while statistical measures and tests either confirm or refute  them.  We present the detailed results and analyses for CNN/DM sample dataset in this subsection. Corresponding results for PubMed and IN-Abs datasets are reported in~\ref{sec:app-dataset-level-results}. 

\noindent
\textit{Visual Analysis}: Figure~\ref{fig:cnn-dm-ridge-plots} presents ridgeline plots of the distributions of the  three metric scores  for $32$ documents in the sample. Each horizontal ridgeline corresponds to one document and represents the score distribution of $100$ generated summaries. The stacked ridgeline plots clearly evince the document-level variability for all three summarizers and metrics. Distributions for  BERTScore  display  least variability (Fig.~\ref{fig:cnn-dm-bertscore-ridge}), while AlignScore distributions are markedly broader (Fig.~\ref{fig:cnn-dm-alignscore-ridge}), flatter and more irregular. This  reflects substantial variation in alignment of individual summaries in the set with their respective source document for the  three LLM-summarizers.  Apparently,  BERTScore exhibits greatest \textit{stability} and AlignScore displays least \textit{stability} for all  summarizers. In Fig.~\ref{fig:cnn-dm-scscore-ridge}, SC score that reflects the variability among the generated summaries also exhibits moderate variation in its distribution. Visual inspection of the plots in Fig.~\ref{fig:cnn-dm-ridge-plots} hints lack of \textit{normality} in distributions, especially for SC scores and AlignScore values for the three LLM-summarizers.  Interestingly, ridgeline plots for PubMed and IN-Abs sample sets also exhibit similar trend (Fig.~\ref{fig:app-pubmed-in-abs-ridge-plots} in~\ref{sec:app-dataset-level-results}). 

\noindent
\textit{Statistical Test for Normality}:  The Shapiro{-}Wilk test confirms that the distributions deviate from \textit{normality}. The null hypotheses\footnote{$H_0$: The data distribution follows normality vs. $H_1$: The data distribution is not normal.}  is rejected  for large number of  documents at $5\%$ level of significance for all the score distributions\footnote{Distribution of hundred scores for each of the 32 (documents) x 3 (genres) x 3 (metric) x 3 (summarizers)} and genres (Table~\ref{tab:shapiro-wilk-test-5} in~\ref{sec:app-dataset-level-results}). Guided by these findings, we  use non-parametric statistical tests for further investigations.

\begin{figure}
    \centering
    \begin{subfigure}[t]{0.48\linewidth}
        \centering
        \includegraphics[width=\linewidth, height=5cm]{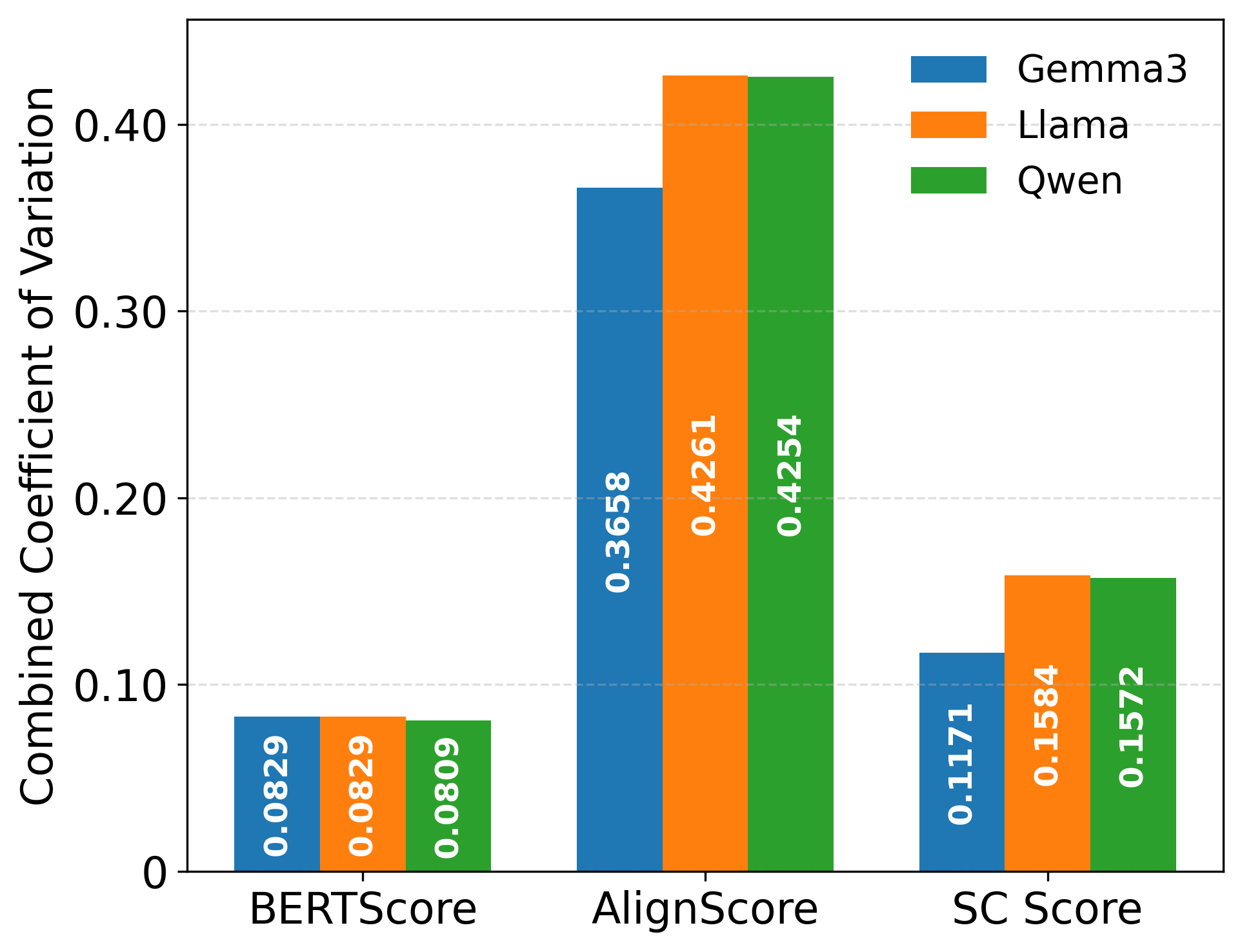}
        \caption{Combined Coefficient of Variation}
        \label{fig:combined-cv-cnn-dm}
    \end{subfigure}
    \hfill
    \begin{subfigure}[t]{0.48\linewidth}
        \centering
        \includegraphics[width=\linewidth, height=5cm]{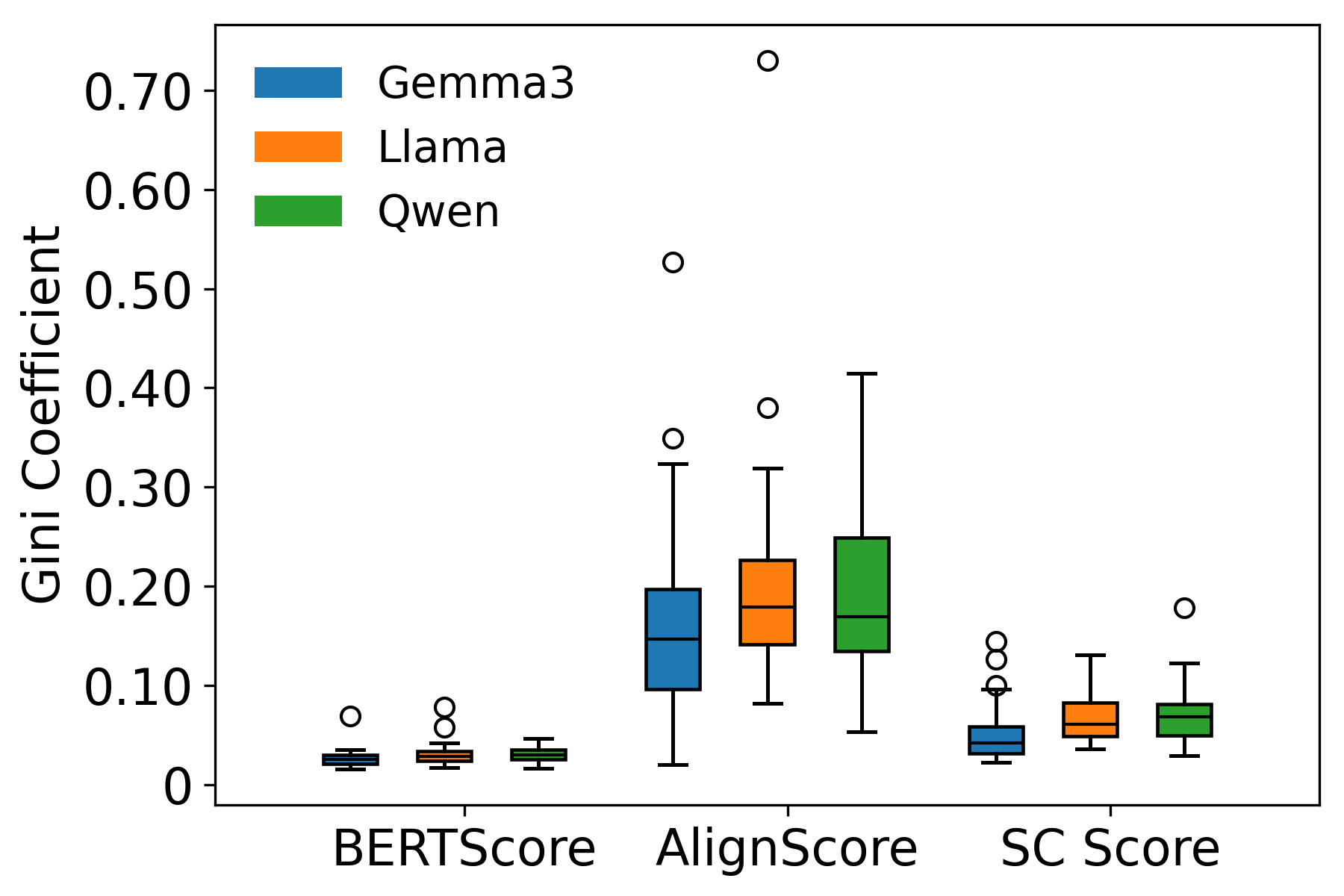}
        \caption{ Gini Coefficient}
        \label{fig:cnn-dm-gini-box-plot}
    \end{subfigure}
    \caption{(a) $CV_c$ values for the three metrics and LLM-summarizers, (b) boxplots of document-level GC values for the CNN/DM sample dataset.}
    \label{fig:combined-cvc-gc}
\end{figure}

\noindent
\textit{Descriptive measures}: The combined coefficients of variation, shown in Fig.~\ref{fig:combined-cv-cnn-dm}, indicate  the relative total variability of the three metric scores for the $3200$ summaries produced by all the three LLMs for CNN/DM articles.  It is clear that the variability in  BERTScore values is the lowest among all metrics, and comparable for all LLM-summarizers. Gemma3 exhibits lower variability than the other two summarizers for both SC and Align scores, while coefficient of variation for Llama and Qwen are very close. The trends for PubMed and IN-Abs documents are alike, with BERTScore values indicating least variability (highest \textit{stability}) and AlignScore being most variable (Fig.~\ref{fig:app-combined-cv} in~\ref{sec:app-dataset-level-results}). AlignScore $CV_c$ values for Llama-summarizer are notably higher than the other two summarizers. We believe that Gemma3-summaries align well with the source documents as compared to the Llama-summaries, despite both summarizers having comparable context windows of $128K$ tokens. Qwen-summarizer, which has much smaller context length of $32K$ tokens exhibits align score comparable to those of Llama. This suggests that context window size \textit{alone} does not explain the observed differences in the align score. \textit{Interestingly,  the earlier observation from document-level analysis that the \textit{stability} of an LLM-summarizer for different quality metrics varies is endorsed.}  

To complement the combined coefficients of variation, we compute the Gini coefficient to measure the extent of inequality in the $100$ summary scores. The box plots shown in Fig.~\ref{fig:cnn-dm-gini-box-plot} indicate that the  Gini coefficient (GC) values corresponding to the  BERTScore values for $32$ documents, are closely bunched towards the lower end of the scale. This implies that the BERTScore values of the $100$ summaries are not only close to each other, but the second order variations too are least among the three metrics for all LLM-summarizers. The interquartile range in the SC score boxplots reflects higher variability among the summaries of same set of documents as compared to BERTScore. The GC values for AlignScore show the highest fluctuation, which suggests that the degree of factual misalignment varies more widely among the generated summaries than for the other evaluation metrics. Corresponding plots for other two datasets reveal similar trend in variability of GC scores (Fig.~\ref{fig:app-gini-box-plot} in~\ref{sec:app-dataset-level-results}). 

To affirm the patterns observed in the GC scores, we scrutinize the variance of the summary scores using  Levene's test (for equality of variances) for all metrics and summarizers\label{levene-test}. Interestingly, the null hypotheses are \textit{rejected} at $5\%$ level of significance for all documents in the three datasets, all metrics and  summarizers. This asserts that the degree of variability i.e., second order variations in the summaries generated by the three summarizers, differs for the documents along both semantic and factual dimensions.

\noindent
\textit{Confidence Interval}: To estimate the  range of variability that a user can expect while generating summaries using $\mathcal L$, we compute $95\%$  non-parametric bootstrap confidence interval for the document-level variance of the summary quality for each summarizer $\mathcal{L}$ and genre $\mathcal G$. Given document-level variances $\Sigma = \{\sigma_1^2, \sigma_2^2, \ldots, \sigma_{32}^2\}$, we draw $10{,}000$ bootstrap samples of size $32$ from $\Sigma$, and compute the average variance for each bootstrap sample. Then we use resulting empirical distribution of $10{,}000$ average variances to obtain $95\%$ bootstrap confidence interval using percentile method. The confidence interval for the variance provides a range of plausible values of the metric scores across documents for a given dataset-LLM pair.
\begin{figure}
    \centering
    \begin{subfigure}{0.48\linewidth}
        \centering
        \includegraphics[width=0.9\linewidth, height=5cm]{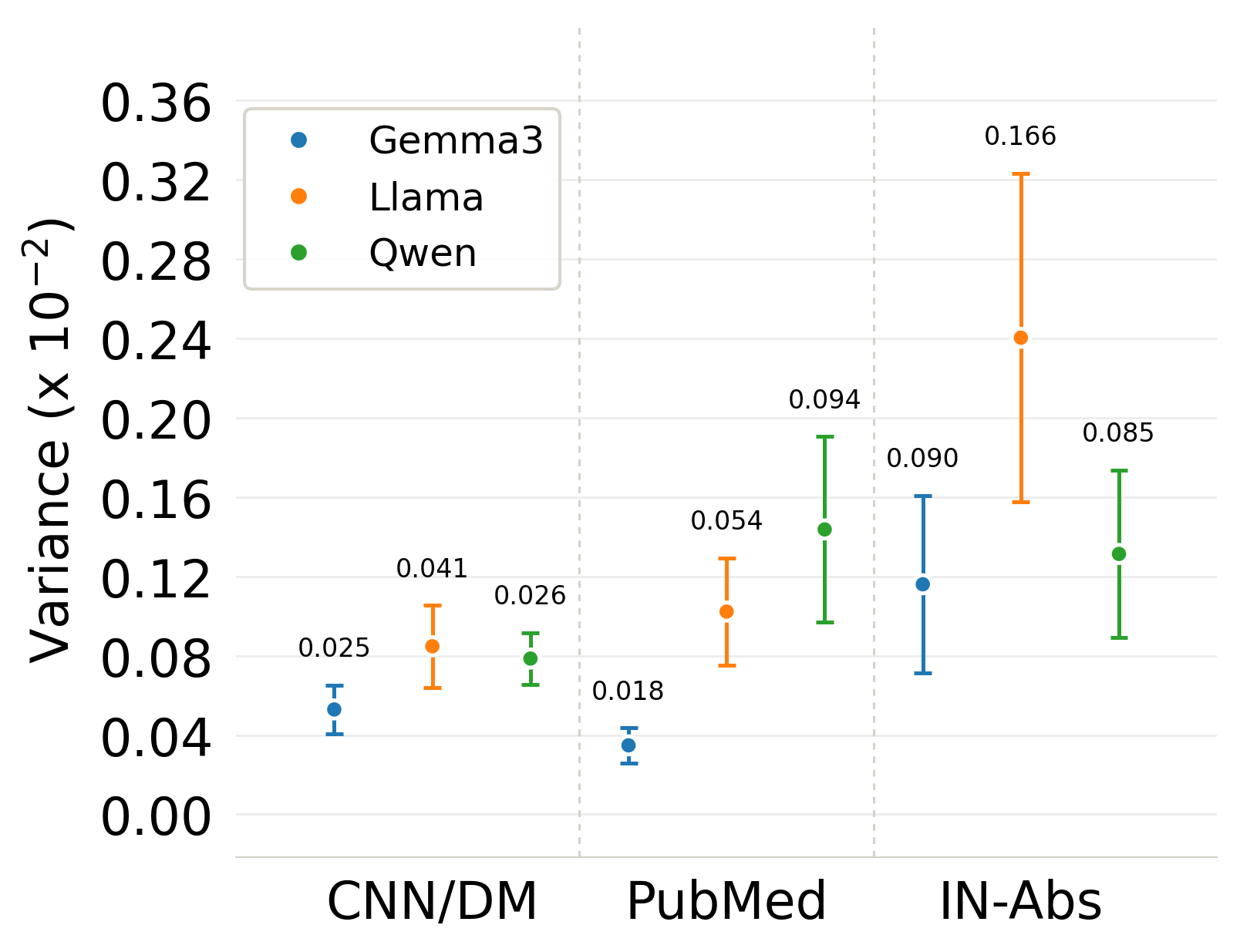}
        \caption{BERTScore}
        \label{fig:ci-bertscore-plot}
    \end{subfigure}
    \hfill
    \begin{subfigure}{0.48\linewidth}
        \centering
        \includegraphics[width=0.9\linewidth, height=5cm]{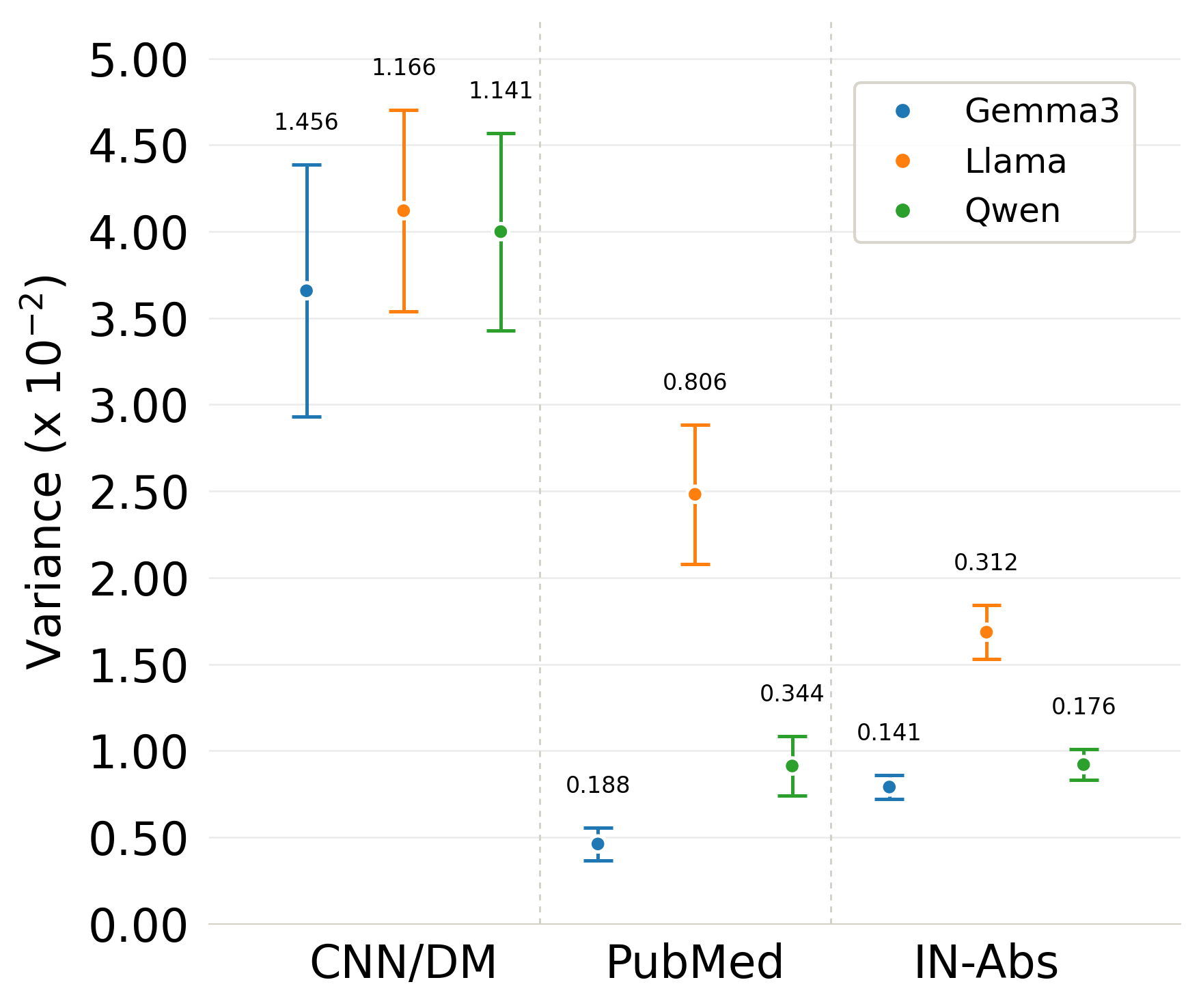}
        \caption{AlignScore}
        \label{fig:ci-alignscore-plot}
    \end{subfigure}
    \caption{Bootstrap confidence intervals for the mean document-level variance of BERTScore (× $10^{-2}$) and AlignScore (× $10^{-2}$) across the CNN/DM, PubMed, and IN-Abs datasets, grouped by model (Gemma3, Llama, and Qwen). Dots represent the bootstrap point estimate of the average document-level variance, and error bars denote the estimated 95\% confidence interval.}
    \label{fig:confidence-interval-plot}
\end{figure}

Figure~\ref{fig:confidence-interval-plot} shows the $95\%$ confidence intervals for variance in BERTScore and AlignScore for all LLM-summarizers and genres. Differences in the widths of the confidence intervals  across metrics, genres and summarizers indicates differences in summary variability over three facets. The observation is consistent with results of the Levene's test for equality of variance. For BERTScore, the estimated confidence interval (CI) is tight for CNN/DM and PubMed, but wide for IN-Abs. AlignScore shows a different trend, with CNN/DM dataset exhibiting highest variances and widest intervals among the three corpora and summarizers. 
Interestingly, Gemma3 consistently shows the lowest variance among the three summarizers and corpora, except on CNN/DM for AlignScore.  
The ridgeline plots in Figs.~\ref{fig:cnn-dm-ridge-plots} and~\ref{fig:app-pubmed-in-abs-ridge-plots} visually corroborate these observations by exhibiting analogous patterns of variability for all LLM-summarizers and datasets.

\textit{Overall, a narrow confidence interval indicates more consistent and predictable performance of the summarizer, which is likely to foster greater user trust. Conversely, wide interval reflects high uncertainty in summary quality, reducing users' confidence in relying on the summarizer across diverse genres.}
\begin{table}[width=.38\linewidth,cols=3,pos=h]
\centering
    \caption{Metric-wise stability coefficients ($\mathcal{E}$) of three LLMs for the CNN/DM sample dataset.}
    \label{tab:cnn-dm-epsilon-stability}
    \begin{tabular*}{\tblwidth}{l|cc}
    \toprule
         \textbf{LLM}&  \textbf{BERTScore}& \textbf{AlignScore}\\ \midrule
         \textbf{Gemma3}& \textbf{ 0.7957} & 0.0031  \\
         \textbf{Llama}&  0.6461& 0.0021   \\
         \textbf{Qwen}&  0.7857& \textbf{0.0100}\\ 
         \bottomrule
    \end{tabular*}    
\end{table}

\noindent
\textit{Stability Coefficient}:  Table~\ref{tab:cnn-dm-epsilon-stability} shows the estimated \textit{stability coefficients} of the three LLM-summarizers,  computed using Def. \ref{def:llm-lvl-stab}. Recall that the estimates are drawn from the analysis of hundred summaries for a representative sample of 32 documents for each corpus (genre) and LLM-summarizer. It is clear that among the two summary evaluation metrics, BERTScore is more \textit{stable} than AlignScore. The relative  $\mathcal{E}$ values endorse the conclusions based on $CV_c$ values (Fig. \ref{fig:combined-cv-cnn-dm}), $GC$ boxplots  of the summaries (Fig. \ref{fig:cnn-dm-gini-box-plot}) and the confidence intervals (Fig.~\ref{fig:confidence-interval-plot}). Extremely low $\mathcal{E}$'s for AlignScore metric signal conspicuously high variability in  summaries generated by all summarizers. Among the genres, \textit{stability coefficients} for PubMed and IN-Abs (long documents) are higher compared to CNN/DM dataset (Table~\ref{tab:pubmed-in-abs-epsilon-stability} in ~\ref{appendix:llm-level-stability-analysis}). \textit{ The analysis affirms that  Gemma3-summarizer generates  semantically more stable summaries for all datasets.}
\subsection{Benchmarking Trustworthiness of LLM-Summarizers}
\label{subsec:result-benchmark-llm-summarizer}
Benchmarking trustworthiness of LLM-summarizers is important not only from end-users' perspective. As the research community strives to develop reliable and trustworthy Generative AI tools and agents, stability of LLM-summarizers and reliability of summary evaluation metrics assume critical significance for further advancements in the field.  We compare reliability and trustworthiness of the three LLM-summarizers considered in this study to reveal the finer aspects of their relative performances for the three document genres. We consolidate the analyses done so far, and use visualization tools and statistical measures to justify our conclusions. 
\begin{figure}
    \centering
    \begin{subfigure}[b]{0.32\textwidth}
        \centering
        \includegraphics[width=\textwidth]{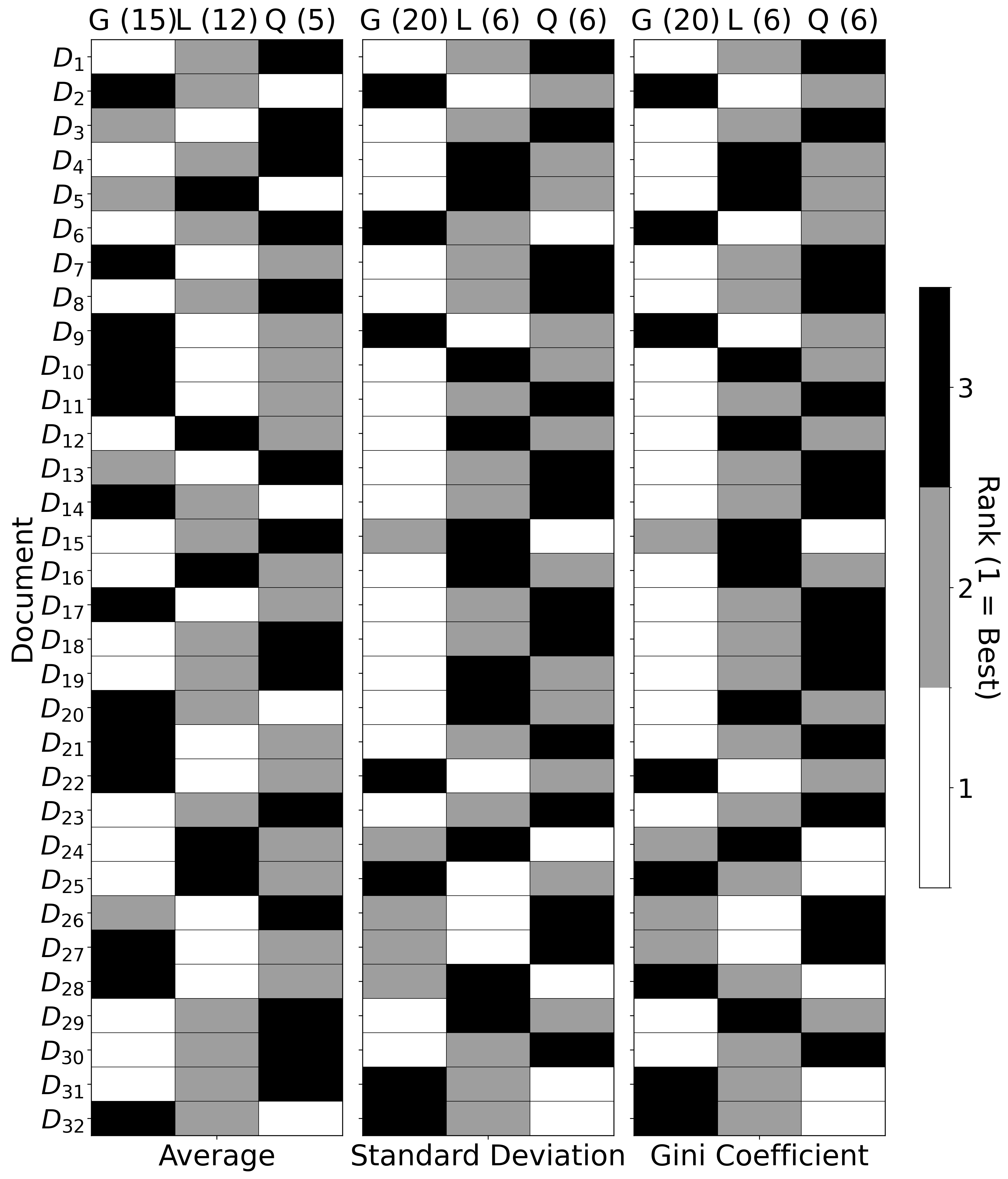}
        \caption{ BERTScore}
    \end{subfigure}
    \hfill
    \begin{subfigure}[b]{0.32\textwidth}
        \centering
        \includegraphics[width=\textwidth]{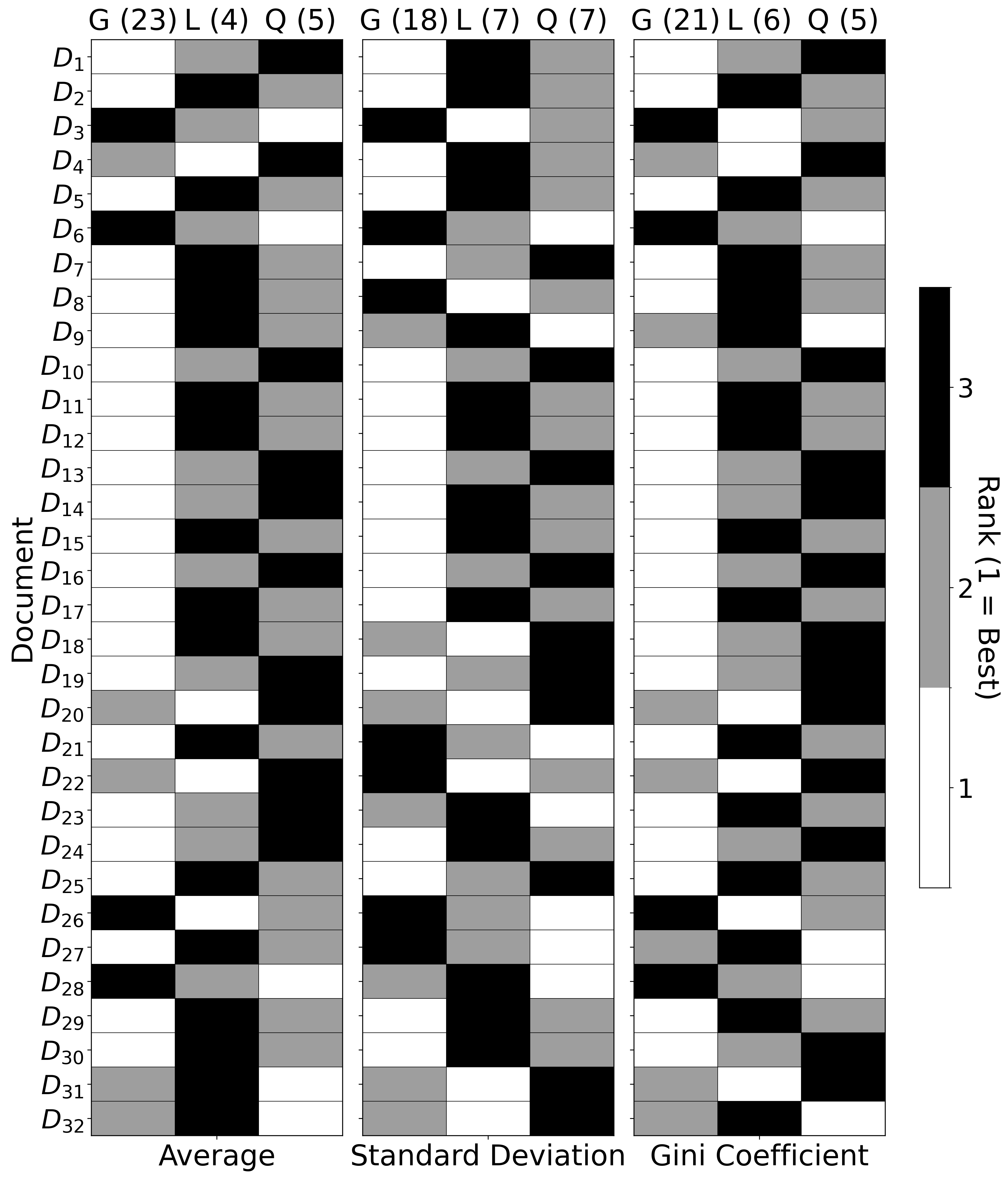}
        \caption{AlignScore}
        \label{fig:cnn-rank-heatmap-alignscore}
    \end{subfigure}  
    \hfill
    \begin{subfigure}[b]{0.32\textwidth}
        \centering
        \includegraphics[width=\textwidth]{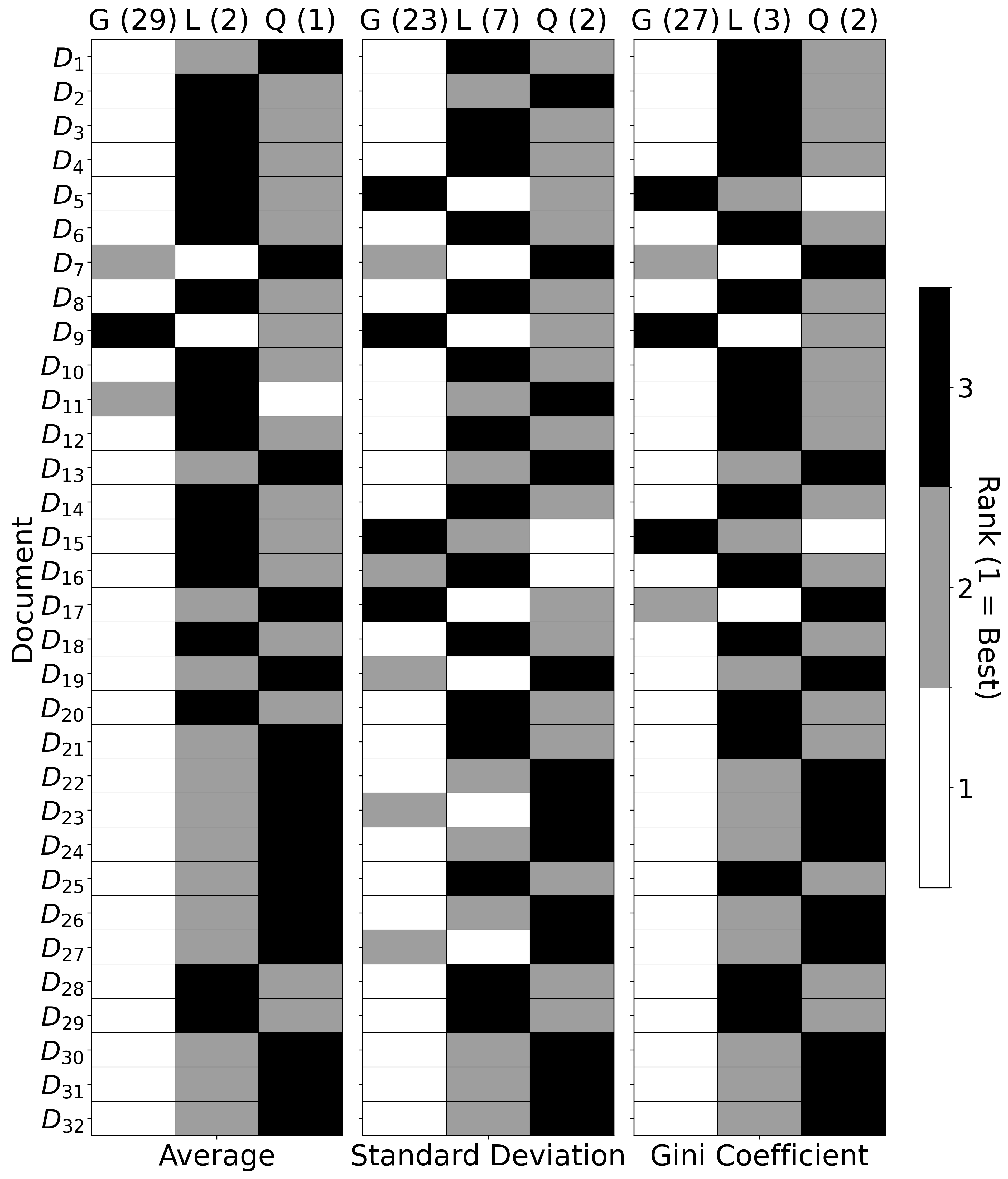}
        \caption{SC Score}
        \label{fig:cnn-rank-heatmap-scscore}
    \end{subfigure}
    \caption{Rank heatmaps for the three LLMs (\textbf{G}emma, \textbf{L}lama, \textbf{Q}wen), and three metrics for CNN/DM documents ($D_1$–$D_{32}$). In each plot, the three blocks represent average, standard deviation, and Gini coefficient, respectively. Ranks 1, 2, and 3 are shown in white,  gray and black respectively. The numbers at the top denote \textit{wins}. }
    \label{fig:cnn-dailymail-ranked-heatmap-plots}
\end{figure}

Figure~\ref{fig:cnn-dailymail-ranked-heatmap-plots} shows the rank heatmaps of the \textit{average, standard deviation,} and \textit{Gini coefficient} (GC) of the three metrics for each of the 32 documents ($D_1$–$D_{32}$) in the sample documents for the three corpora.  The \textit{average}\footnote{For the BERTScore and AlignScore metrics, the ranks are based on  the average computed over 100 generated summaries for each document.  SC score, which itself is an averaged metric, is used as computed in Eq.~\ref{eq-sc-score}.} for each metric (first block) reflects the quality of the summary, with the highest value ranked as 1. In contrast, the \textit{standard deviation} (second block) and \textit{GC} (third block), which quantify dispersion, receive rank 1 for the lowest value. The LLM-summarizer (model, henceforth) with rank 1 is designated as \textit{winner} for the respective document. 
 
The \textit{win} counts on top of the  heatmaps acknowledge Gemma3 as the best  performing summarizer for all three statistics and metrics.  Highest \textit{win} count for \textit{average} for all metrics indicates that Gemma3 summaries align best with the gold standard reference and the source document, and are also semantically most consistent among themselves. Maximum \textit{wins} for \textit{standard deviation} and GC for all the three metrics  beget higher \textit{stability} to Gemma3, compared  to the other two summarizers. Figure~\ref{fig:app-pubmed-in-abs-ranked-heatmap-plots} (\ref{appendix:llm-level-stability-analysis}) presents similar plots for PubMed and IN-Abs datasets, both of which exhibit remarkably similar trends. 

\begin{table}[width=.48\linewidth,cols=4,pos=h]
\centering
    \caption{Metric-wise  average ranks of three LLMs  based on the coefficients of variation (CV) of three metric scores for the CNN/DM sample dataset.}
    \label{tab:cnn-dm-avg-ranks}
    \begin{tabular*}{\tblwidth}{l|ccc}
    \toprule
         \textbf{LLM}& \textbf{BERTScore}&\textbf{AlignScore} & \textbf{SC Score}\\ \midrule
         \textbf{Gemma3}  & \textbf{1.59}  &\textbf{1.44} & \textbf{1.25} \\
         \textbf{Llama}&   2.13 &2.31 & 2.34 \\
         \textbf{Qwen}&   2.28 &2.25  & 2.41\\ 
         \bottomrule
    \end{tabular*}    
\end{table}
For further verification, we tabulate the \textit{average ranks} of LLM-summarizers based on a complementary  statistical measure, coefficients of variation (CV - lower value gets higher rank),  of document-level summary scores for the three datasets and evaluation metrics (Table~\ref{tab:cnn-dm-avg-ranks}). Since CV is a unit-free statistic, it enables scale-independent comparison of stability among models. The higher average rank of Gemma3 substantiates its \textit{stability} and the consistency among summaries for short news articles from CNN/DM corpus. Table~\ref{tab:pubmed-in-abs-average-rank-llm} (\ref{appendix:llm-level-stability-analysis}) presents similar result for long technical articles in PubMed (medical) and IN-Abs (legal) corpora.
\begin{figure}
    \centering
    \begin{subfigure}{0.32\linewidth}
            \centering
            \includegraphics[width=\linewidth]{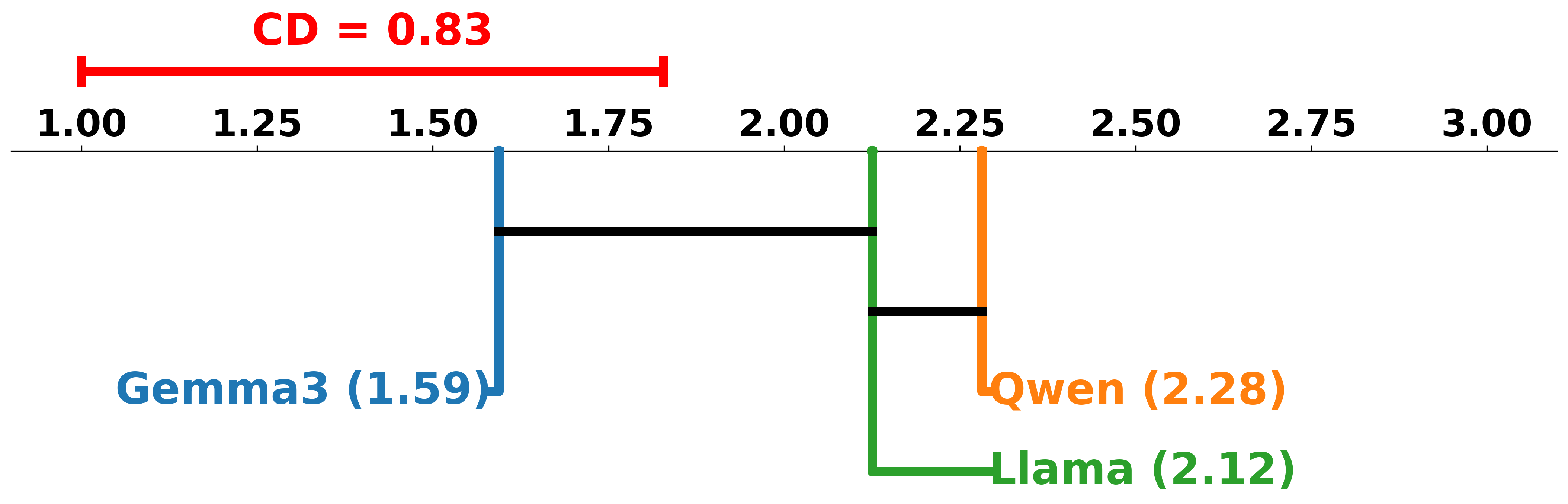}
            \caption{BERTScore}
            \label{fig:cnn-dm-cd-plot-bertScore}
        \end{subfigure}
         \hfill
        \begin{subfigure}{0.32\linewidth}
            \centering
            \includegraphics[width=\linewidth]{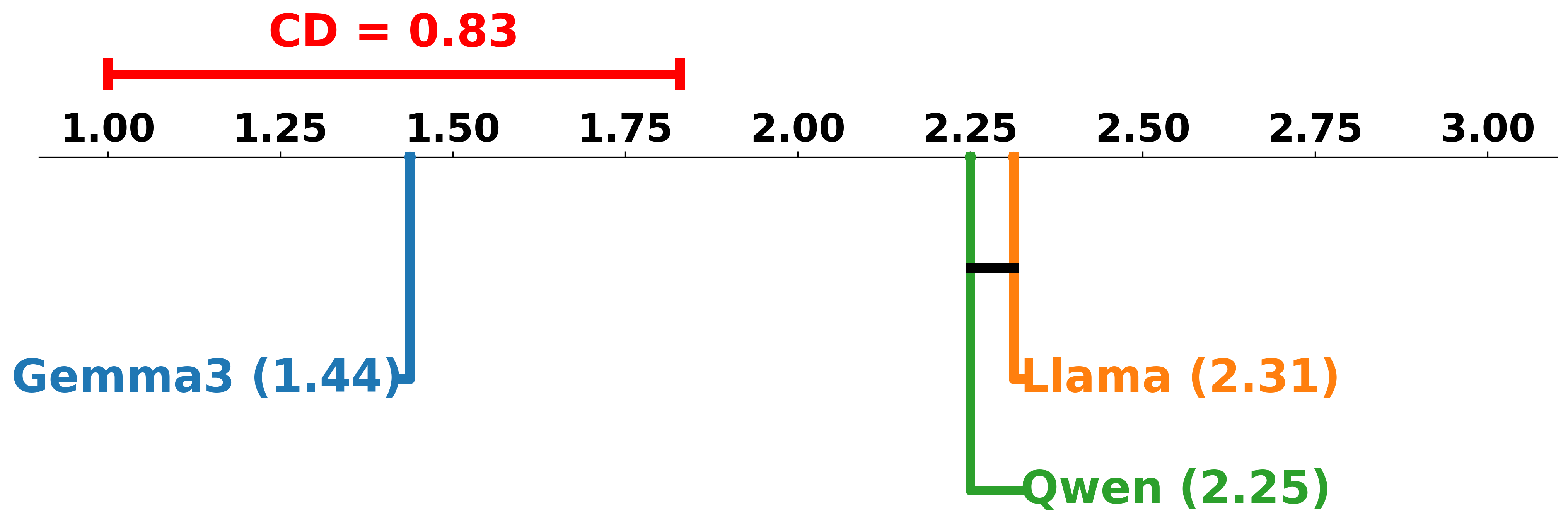}
            \caption{AlignScore}
            \label{fig:cnn-dm-cd-plot-alignScore}
        \end{subfigure}
        \hfill
        \begin{subfigure}{0.32\linewidth}
            \centering
            \includegraphics[width=\linewidth]{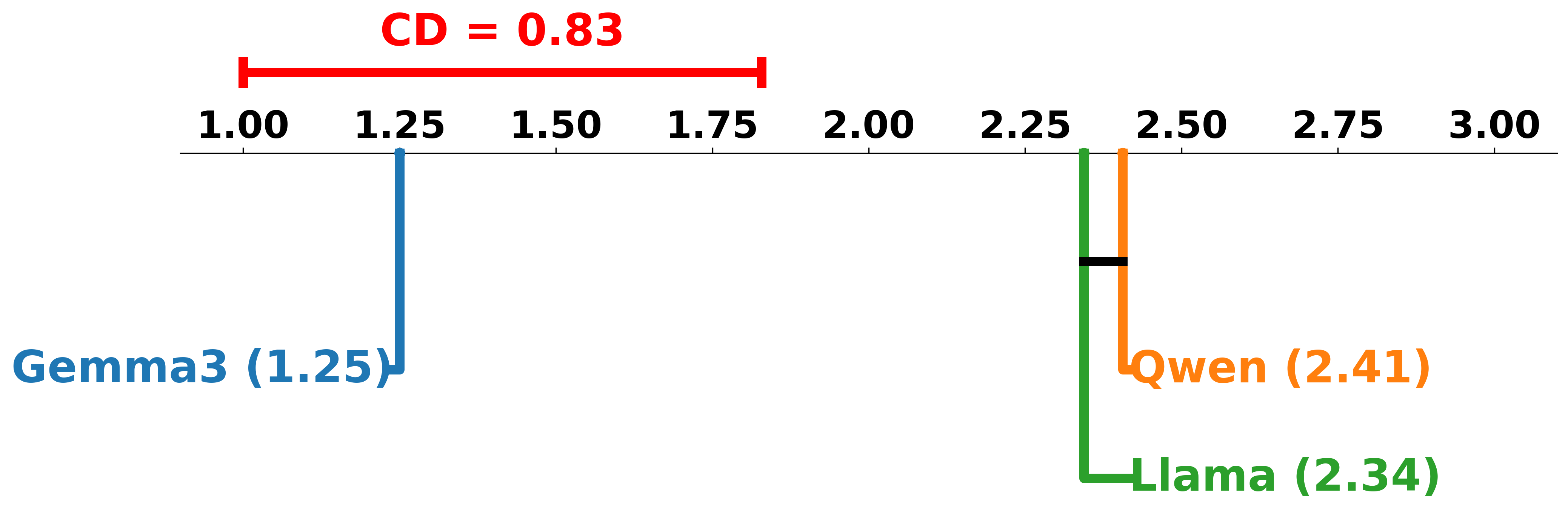}
            \caption{SC Score}
            \label{fig:cnn-dm-cd-plot-scScore}
        \end{subfigure}
       \caption{Critical difference (CD) diagrams showing the average ranks of LLM-summarizers based on BERTScore, AlignScore, and Semantic Consistency (SC) Score for the CNN/DM (news articles) sample set. The horizontal line denotes the critical difference (CD). Models connected by this line do not differ significantly according to the Nemenyi post-hoc test following the Friedman test.}
    \label{fig:cd-plots-cnn-dm}
\end{figure}
We  validate the differences in observed ranks of LLM-summarizers using the \textit{Friedman} test (Sec.~\ref{sec:benchmarking-LLM}).  The test \textit{rejects} $H_0$ for all documents in the three datasets at $5\%$ level of significance indicating that there is  significant difference in the ranks of \textit{at least} one pair of summarizers. The post-hoc \textit{Nemenyi} test identifies the  pairs with  significantly different ranks. The CD diagram shown in Fig.~\ref{fig:cd-plots-cnn-dm} shows that there is no difference in the BERTScore based ranks for  all pairs of summarizers (contradicts "all" in $H_0$ sentence). However, the pairs <Gemma3, Llama> and <Gemma3, Qwen> have statistically significant difference in ranks based on  SC score and AlignScore values for short news articles  (CNN/DM corpora). Llama and Qwen attain comparable ranks, and  no statistically significant difference exists between their average ranks for CNN/DM documents. The metric-wise average ranks shown in Table~\ref{tab:cnn-dm-avg-ranks} corroborate that Gemma3 achieves the best average rank for all metrics, indicating highest \textit{stability} in the generated summaries. 
 
For the PubMed and IN-Abs sample sets, the results of \textit{Friedman-Nemenyi} test are not completely leveled with CNN/DM documents (Figure~\ref{fig:app-cd-plots-pubmed-inabs} in~\ref{appendix:llm-level-stability-analysis}). However, the superiority of Gemma3 ranks is striking  for all metrics and both datasets (Table \ref{tab:pubmed-in-abs-average-rank-llm}). The detailed analysis of the test results for these two datasets is presented in~\ref{appendix:llm-level-stability-analysis}.
\begin{table}[width=.92\linewidth,cols=8,pos=h]
\caption{Stability coefficients ($\mathcal{E}$) and Coefficients of variation ($CV_c$) for the considered metrics and the three LLM-summarizers, with ranks in parenthesis.}
\label{tab:meta-stability-cvc}
\begin{tabular*}{\textwidth}{@{\extracolsep{\fill}}llll|lll}
    \toprule
     & & \multicolumn{2}{c}{\textbf{Stability Coefficient}}& \multicolumn{3}{c}{\textbf{Combined Coefficient of Variation}}\\ \midrule
          \textbf{Dataset} & \textbf{LLM} &  \textbf{BERTScore}& \textbf{AlignScore} & \textbf{BERTScore}& \textbf{SC Score}&\textbf{AlignScore} \\ \midrule
                   &   \textbf{Gemma3}& 0.7957 (i)&   0.0031 (ii) & 0.0829 (ii)& 0.1171 (i)&0.3658 (i) \\
    \textbf{CNN/DM}&  \textbf{Llama}&  0.6461 (iii)&  0.0021 (iii) & 0.0829 (ii)& 0.1584 (iii)&0.4261 (iii) \\
                   &  \textbf{Qwen}&  0.7857 (ii)&   0.0100 (i) & 0.0809 (i)& 0.1572 (ii)&0.4254 (ii) \\ \midrule
                   &\textbf{Gemma3}&  0.8036 (i)&   0.4345 (i) & 0.0765  (i)& 0.0702 (i)&0.1493 (i) \\
    \textbf{PubMed}&\textbf{Llama}&  0.6834 (ii)&   0.0893 (iii) & 0.0779  (ii)& 0.1582 (ii)&0.3074 (iii) \\
                    &\textbf{Qwen} &  0.3925 (iii)& 0.1624 (ii) & 0.0876 (iii)& 0.1620 (iii)&0.1821 (ii) \\  \midrule
                    &\textbf{Gemma3}&  0.6668 (i)  & 0.2725 (i) & 0.0788 (ii)& 0.0876 (i)&0.1574 (i) \\
    \textbf{IN-Abs} &\textbf{Llama}&  0.5660 (iii) & 0.1618 (iii) & 0.0851 (iii)& 0.1569 (iii)&0.2657 (iii) \\
                    &\textbf{Qwen}&  0.6477 (ii)  & 0.2352 (ii) & 0.0754 (i)& 0.1164 (ii)&0.1815 (ii)\\ 
    \bottomrule
\end{tabular*}
\end{table}
\section{Meta-Analysis and Discussion}
We present the condensed view of the  results discussed in the preceding sections, structured to elucidate the principal findings and key insights derived from this study. Specifically, we relate the performance of the LLM-summarizers with the three evaluation metrics (BERTScore, AlignScore, SC score) and genres (news article, medical articles, legal judgments). Recall from Sec.~\ref{sec:datasets} that the three chosen genres represent  diversity\footnote{%
\begin{minipage}[t]{0.95\linewidth}
CNN/DM documents are news articles, short in length and written in simple language for consumption of the public at large. \\
PubMed articles are lengthy scientific documents containing medical and technical terms and are typically consumed by medical professionals.\\
IN-Abs documents are generally long judgments containing complex legal language, comprising long complex sentences with legal terms and possibly ambiguous interpretations that are consequential for law practitioners.
\end{minipage}} in language complexity, writing style, document length, etc. 

\subsection{Stability of Metrics and  Summarizers}
Table \ref{tab:meta-stability-cvc} shows two complementary indicators of the stability in the LLM-generated summaries along with the corresponding summarizer ranks in parenthesis. The left partition of the table shows the estimated \textit{stability coefficients} ($\mathcal E$), and the right partition shows the combined coefficients of variations ($CV_c$). The former metric betokens the pessimistic  estimate of the summary \textit{stability}, while the latter expresses the observed \textit{variability} in the summaries. Since SC score itself is  a point estimate, estimating its \textit{stability coefficient} is not meaningful. 
 
\noindent
\textit{Metric-wise consolidation}: Based on the corpus-level analysis and LLM-benchmarking study, we infer the following from Table~\ref{tab:meta-stability-cvc}.  \\
(i)  It is evident  that BERTScore   is a more \textit{stable} evaluation metric compared to AlignScore (left partition of Table~\ref{tab:meta-stability-cvc}). Higher $\mathcal E$ values of  BERTScore indicate tighter semantic alignment among the generated summaries with respect to the gold standard reference summary for the three summarizers. Lower $\mathcal E$ values of AlignScore hint that  the summaries generated  at  different occasions may exhibit remarkably different factual alignments with respect to the original document, irrespective of the genre of the document. The observation is validated by highest coefficient of variation ($CV_c$) values among the three metrics (right partition of Table~\ref{tab:meta-stability-cvc}). Though the estimated \textit{stability coefficient} portrays the worst case scenario (Sec.~\ref{sec:formal-problem-statement}), yet its low value for AlignScore stands as a grim reminder that once-in-a-while an LLM-summarizer may generate summaries that are detached from the facts stated in the  original document and may confuse the consumer, thereby critically diminishing the user's trust.
   
AlignScore also exhibits wider confidence intervals (CI) for variance on short documents, but tighter confidence intervals  for longer and more information-dense documents in PubMed and IN-Abs datasets.  Lower variance and tighter CI in AlignScore for long documents is a deceptive indicator of the metric stability, and must be interpreted in conjunction with higher values of combined coefficient of variation (Fig. \ref{fig:combined-cv-cnn-dm}, \ref{fig:app-combined-cv-pubmed}, \ref{fig:app-combined-cv-in-abs}) and Gini Coefficient (Fig.~\ref{fig:cnn-dm-gini-box-plot}). These mixed signals for AlignScore impel further investigation of the behavior and characteristics of the metric. Moreover, consistently lower average  variance (indicated by the interval centroids) of BERTScore, relative  to  AlignScore,   demonstrates its  greater  reliability as a summary evaluation metric for all summarizers and genres.  \\
\noindent
 (ii) SC score, which measures the semantic consistency among the generated summaries, shows moderate combined coefficient of variation ($CV_c$) for all summarizers and  document genres (right partition of Table~\ref{tab:meta-stability-cvc}). The reported measure is computed from the $\binom{100}{2}$ pairwise cosine similarity scores for 32 sample documents for each genre. The moderate combined coefficient of variation ($CV_c$) indicates that substantial variability exists both among the summaries generated for a document, and among different documents. The latter is supported by the Levene's test (Sec.~\ref{subsec:dataset-level-stability-analysis}, page~\pageref{levene-test}). Consequently, the evaluation outcomes based on SC score should be interpreted with appropriate caution, particularly in domains where consistent and trustworthy evaluations are essential.
 
\begin{table}[width=.95\linewidth,cols=8,pos=h]
    \centering
    \caption{Metric-wise \textit{win-counts} based on average and standard deviation of 100 summaries for each document in the representative sample. \textit{Highest} average score is counted as a \textit{win}, while \textit{least} standard deviation is counted as a \textit{win}. }
    \label{tab:meta-win-count}
    \begin{tabular*}{\textwidth}{@{\extracolsep{\fill}}ll c c c|c c c}
    \toprule
 & & \multicolumn{3}{c}{\textbf{Average}}& \multicolumn{3}{c}{\textbf{Standard Deviation}}\\ \midrule
          \textbf{Dataset} & \textbf{LLM} &  \textbf{BERTScore}& \textbf{AlignScore} &  \textbf{SC Score}& \textbf{BERTScore}&\textbf{AlignScore} & \textbf{SC Score}\\ \midrule
                   &   \textbf{Gemma3}& \textbf{15}& \textbf{23}&  \textbf{29} & \textbf{20}& \textbf{18}&\textbf{23}\\
    \textbf{CNN/DM}&  \textbf{Llama}&  12& 4 &  2& 6& 7&7\\
                   &  \textbf{Qwen}&  5& 5 &  1 & 6& 7 & 2 \\ 
            \midrule
                   &\textbf{Gemma3}&  \textbf{20}& \textbf{26} &  \textbf{32} & \textbf{25}& \textbf{29}&\textbf{28}\\ 
    \textbf{PubMed}&\textbf{Llama}&  7&  1& 0 & 3& 0 & 1 \\
                    &\textbf{Qwen} &  5&  5& 0 & 4& 3&3\\  \midrule
                    &\textbf{Gemma3}&  \textbf{16}&  \textbf{26}& \textbf{32} & \textbf{20}& \textbf{23}&\textbf{26}\\
    \textbf{IN-Abs} &\textbf{Llama}&  11&  0& 0 & 3& 0  & 1 \\
                    &\textbf{Qwen}&  5&  6 & 0 & 9&  9 & 5\\ 
         \bottomrule
    \end{tabular*}
\end{table}

\noindent
\textit{Summarizer-wise consolidation:} The following summarizer-wise findings can be deduced directly from the stability coefficients (Table~\ref{tab:meta-stability-cvc}), and a detailed document-wise \textit{win-count} analysis (Table \ref{tab:meta-win-count}). The \textit{win-count} analysis  takes into account the representative sample of 32 documents for each genre, and counts the number of documents for which the summarizer exhibits the highest average score (left partition of Table \ref{tab:meta-win-count}) and the lowest standard-deviation (right partition of Table \ref{tab:meta-win-count}).\\
(i) For BERTScore, Gemma3 consistently achieves the highest \textit{stability coefficient} for  all genres, while Llama and Qwen   demonstrate mixed  performance (Table~\ref{tab:meta-stability-cvc}). Thus, Gemma3 summaries are most \textit{trustworthy}, but \textit{trustworthiness} of Llama and Qwen summaries is genre-dependent. All three summarizers falter at the \textit{stability} of AlignScore metric (factual alignment) for short news articles (CNN/DM)  summaries.  For other two genres, Gemma3 summaries display highest \textit{stability} for AlignScore, and Llamma the least.  Qwen typically ranks in the middle for AlignScore. \\
\noindent
(ii) The \textit{qualitative} performance of the summarizers based on the \textit{win-count} analysis of the \textit{average} scores of the summary sets  shows that  Gemma3, with highest average BERTScore \textit{wins}, consistently generates best quality summaries across the three genres (Table~\ref{tab:meta-win-count}).  On the other hand, Llama and Qwen show mixed performance.  The highest number of \textit{wins} for average AlignScore by Gemma3 summaries also indicates the best alignment  for most  documents in the sample set. Highest \textit{win-count} attained by  Gemma3 for SC score emphasizes its superior semantic consistency among the generated summaries for all  genres. This observation, in conjunction with its highest \textit{win-count} for standard deviation and its least coefficient of variation,  corroborates its highest  \textit{stability coefficient} for Gemma3 BERTScore (Table~\ref{tab:meta-stability-cvc}). 

Standard deviation, a statistical measure of  dispersion, is   another reliable  signal of the \textit{stability}  of the 100 summaries generated for a document.  Ergo, the \textit{win-count} analysis based on the  \textit{standard deviation} of the scores is a credible indicator of the \textit{stability} of the summarizer. The observation from Table~\ref{tab:meta-win-count}  affirms that Gemma3, the consistent winner with  a significant margin  for all three metric scores exhibits least dispersion among the   summary scores. Thus, Gemma3-summarizer  is observed as the most \textit{stable} summarizer.\\
(iii) The confidence intervals shown in Fig.~\ref{fig:confidence-interval-plot} mark that Gemma3 generally produces the narrowest intervals and lowest variance, whereas Llama exhibits the widest intervals and highest variance across datasets, with Qwen lying in between. This corroborates the \textit{win-count} analysis based on standard deviation. From the  \textit{trust} perspective, these results together ratify that Gemma3 provides the most reliable evaluation estimates, whereas Llama's evaluations should be interpreted with comparatively greater caution due to their higher uncertainty.
\subsection{Key takeaways as one-liners}
\begin{enumerate}
    \item BERTScore is the most \textit{stable} and \textit{trustworthy} summary evaluation metric, whereas AlignScore is the least \textit{stable} and needs to be investigated.
    \item Gemma3  is the most \textit{reliable} and \textit{trustworthy} summarizer among the three LLMs examined.
    \item LLM-generated summaries of short documents (CNN/DM) exhibit the lowest factual reliability among the evaluated datasets.
\end{enumerate}
\subsection{Limitations of the Study}
\noindent
This study has several limitations that provide directions for future research.\\
(i) The empirical evaluation is limited to summarization of documents from three genres, and the findings may not generalize to other document types like dialogues, conversations, meetings, etc. Summary stability of these genres also needs to be examined, as important practical decisions may be based on the generated summaries. \\
(ii)  Our investigation is limited to  \textit{free} and \textit{small} LLM-summarizers, the limitation that arises due to financial and computational resources. The study can be extended to \textit{proprietary} and \textit{large} LLM-summarizers. Moreover, with rapid advances in LLM technology, these results are likely to change with upgraded versions of Gemma3, Llama and Qwen summarizers.\\
(iii) Though proposed statistical framework and embedding-based automatic evaluation metrics provide a rigorous and reproducible assessment of the LLM-summarizer’s stability, incorporating expert or user evaluations would complement the proposed protocol by validating the measured stability.\\
(iv) While repeated summarization enables statistically robust estimation of stability and confidence intervals, it incurs additional computational cost compared with single-run evaluation. Consequently, the number of repetitions required to obtain reliable estimates may need to be balanced against available computational resources in practical benchmarking scenarios.
\section{Conclusion}
In this paper, we formally define the problem of estimating the \textit{stability} of an LLM-summarizer and report a systematic investigation of the  variations in the repeatedly generated summaries  by LLM-summarizers  under the condition of static input environment. We also propose a \textit{Stability Analysis Protocol} (SAP) for benchmarking LLM-summarizers. SAP comprises systematic and standardized set of statistical tests and procedures, enabling scientifically rigorous and reproducible conclusions.  Based on our extensive empirical investigation, we argue for broadening the scope of performance evaluation of an LLM-summarizer beyond the conventional quality metrics by incorporating  \textit{stability} as an additional evaluation dimension. We further advocate the adoption of multi-run or distribution-based evaluation strategies, together with confidence interval estimation, to quantify performance uncertainty and provide statistically grounded measures of reliability and robustness. Such evaluation practices enable more reliable and comprehensive assessment of the performance of LLM-summarizers. To the best of our knowledge, this is the first work to systematically investigate \textit{stability} phenomenon for LLM-summarizers. We believe that our findings provide timely and impactful insights for the research community, especially in light of the growing reliance on LLMs in real-world applications. 

\section*{Acknowledgments}
The authors acknowledge financial support from the xxx, xxx, xxx, under grant No. xxx.




\clearpage
\appendix
\renewcommand{\thesection}{Appendix \Alph{section}}
\renewcommand{\thesubsection}{\Alph{section}.\arabic{subsection}}

%
\section{Derivation of the Combined Coefficient of Variation}
\label{appendix-cvc-derivation}
Let there be $N$ sets of $k$ LLM-generated summaries. Considering each set of $k$ summary scores as a group, we compute combined coefficient of variation ($CV_c$) for $N$ documents (groups) and $kN$ summaries.
For the \(i\)-th group, let \(\bar{x}_i\) and \(\sigma_i\) denote its mean and standard deviation, respectively. Let ${x}_{ij}$ denote the  score of the \(j\)-th summary of the \(i\)-th document.

The pooled mean of all \(N\) groups is given by:
\begin{equation}
\mu_c = \frac{1}{N}\sum_{i=1}^{N}\bar{x}_i, where \; \bar{x}_i = \frac{1}{k}\sum_{j=1}^{k}{x}_{ij}
\end{equation}

The pooled variance can be expressed as the sum of the within-group variance and the between-group variance:
\begin{align*}
\sigma_c^2 
&= \sigma^2_{\text{within}} + \sigma^2_{\text{between}} \\
&= \frac{1}{kN} \left[ \sum_{i=1}^{N} k \sigma_i^2 \right]
+ \frac{1}{kN} \left[ k \left\{ (\bar{x}_1 - \mu_c)^2 + (\bar{x}_2 - \mu_c)^2 + \cdots + (\bar{x}_N - \mu_c)^2 \right\} \right] \\
&= \frac{1}{kN} \left[ \sum_{i=1}^{N} k \sigma_i^2 
+ k \left\{ (\bar{x}_1 - \mu_c)^2 + (\bar{x}_2 - \mu_c)^2 + \cdots + (\bar{x}_N - \mu_c)^2 \right\} \right] \\
&= \frac{1}{kN} \left[ k \sum_{i=1}^{N} \sigma_i^2 
+ k \sum_{i=1}^{N} (\bar{x}_i - \mu_c)^2 \right] \\
&= \frac{1}{N} \left[ \sum_{i=1}^{N} \sigma_i^2 
+ \sum_{i=1}^{N} (\bar{x}_i - \mu_c)^2 \right] \\
\\
\Rightarrow \sigma_c^2 
&= \frac{1}{N} \left[ \sum_{i=1}^{N} \left\{ \sigma_i^2 + (\bar{x}_i - \mu_c)^2 \right\} \right] \\
\\
\Rightarrow \sigma_c 
&= \sqrt{ \frac{1}{N} \left[ \sum_{i=1}^{N} \left\{ \sigma_i^2 + (\bar{x}_i - \mu_c)^2 \right\} \right] }
\end{align*}



Therefore, the combined coefficient of variation is obtained as $CV_c = \frac{\sigma_c}{\mu_c}$.\\
Substituting \(\sigma_c\) into the above expression, we get $CV_c =
\frac{
\sqrt{
\frac{1}{N}
\sum_{i=1}^{N}
\left[
\sigma_i^2 + (\bar{x}_i-\mu_c)^2
\right]
}
}{
\mu_c
}$.

\section{Datasets}
\label{sec:app-dataset}
\begin{figure}[width=.95\linewidth,pos=t]
    \centering
    \includegraphics[width=\textwidth]{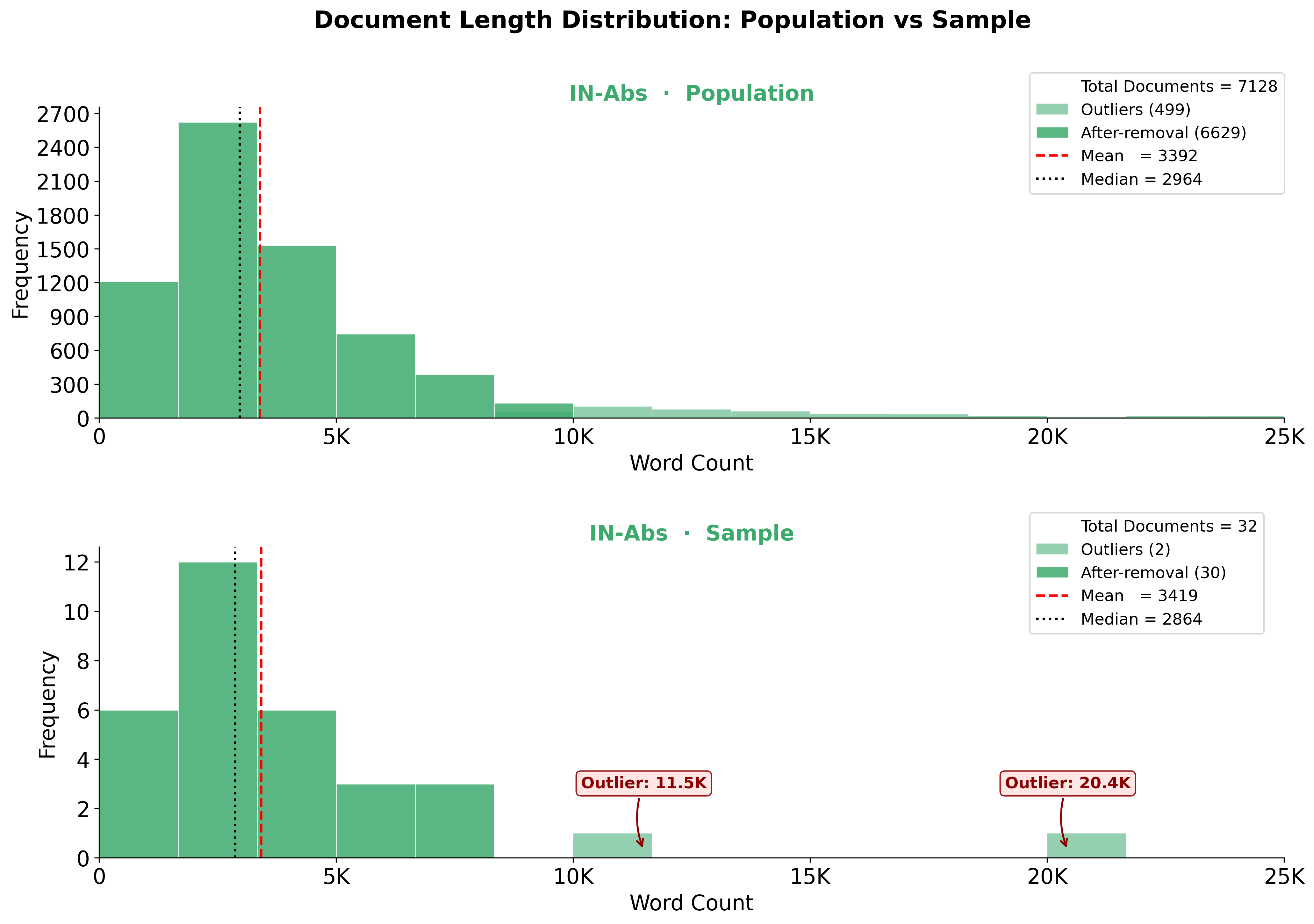}
    \caption{Distribution of document lengths (word counts) in the IN-Abs dataset, comparing the full dataset $(7,128)$ with the selected sample $(32)$ set. Darker bars denote documents within the IQR threshold $(1.5)$, while lighter bars indicate identified outliers. The red dashed and black dotted vertical lines represent the mean and median, respectively, computed from the outlier-filtered data. A chi-square test applied to the outlier-removed distributions, which indicates no statistically significant difference between the sample and population at $5\%$ level of significance. These results suggest that the selected sample closely reflects the population distribution of document length in the IN-Abs dataset.}
    \label{fig:app-doc-length-distribution-plot}
\end{figure}

Figure~\ref{fig:app-doc-length-distribution-plot} shows the distribution of document lengths in the IN-Abs dataset and corresponding sample of $32$ documents. After removing outliers, the chi-square test shows no significant difference between the sample and population at the $5\%$ level of significance, indicating the sample accurately represents the document length distribution in the IN-Abs dataset.

\section{Corpus-level Results} 
\label{sec:app-dataset-level-results}
The corpus-level results for PubMed and IN-Abs sample sets are shown in the following subsections.
\subsection{Ridgeline plots} 
\begin{figure}[pos=h]
    \centering
    \vspace{0.5in}
    \begin{subfigure}[b]{0.32\textwidth}
        \centering
        \includegraphics[width=\textwidth]{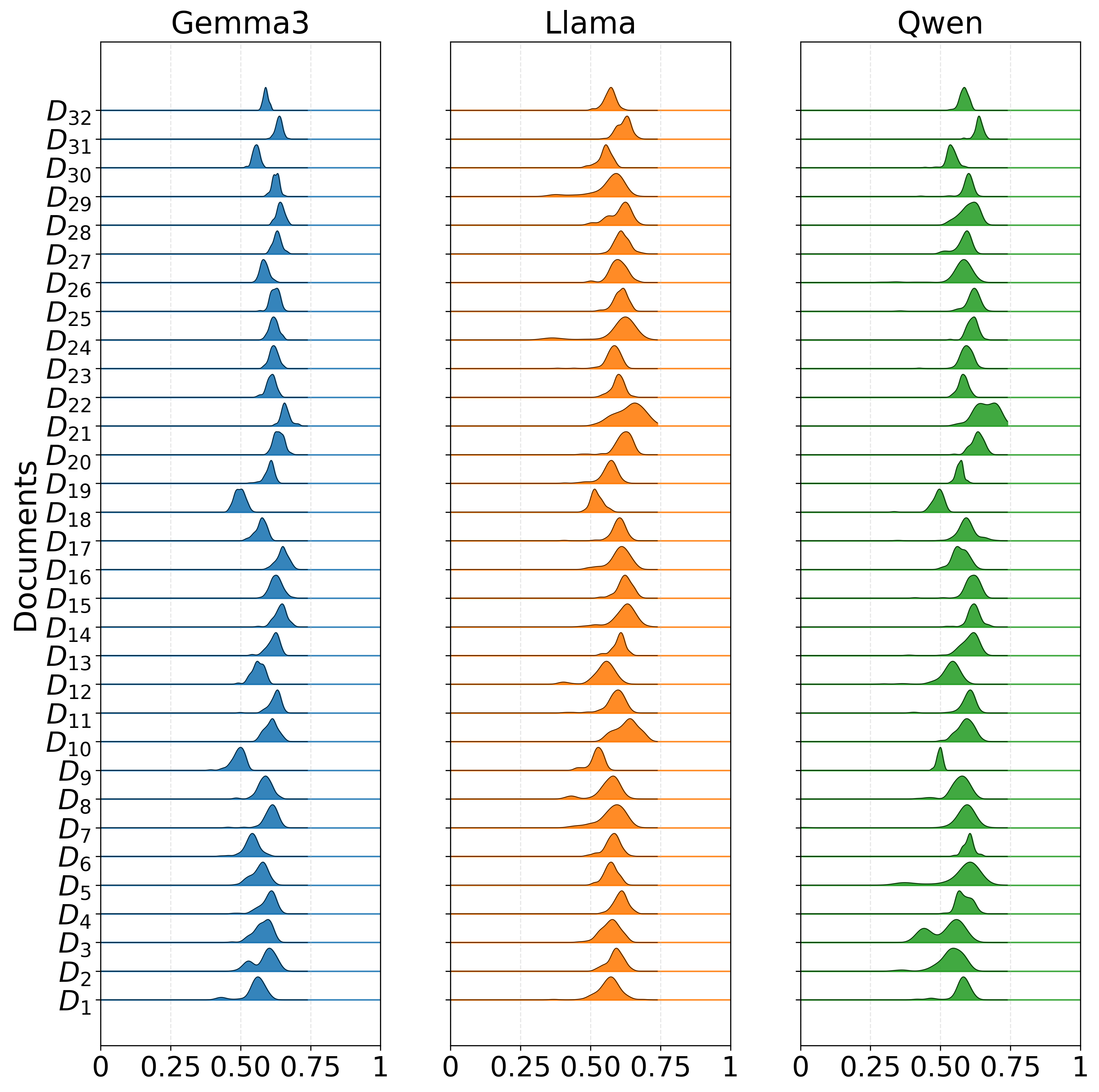}
        \caption{BERTScore}
    \end{subfigure}
    \hfill
    \begin{subfigure}[b]{0.32\textwidth}
        \centering
        \includegraphics[width=\textwidth]{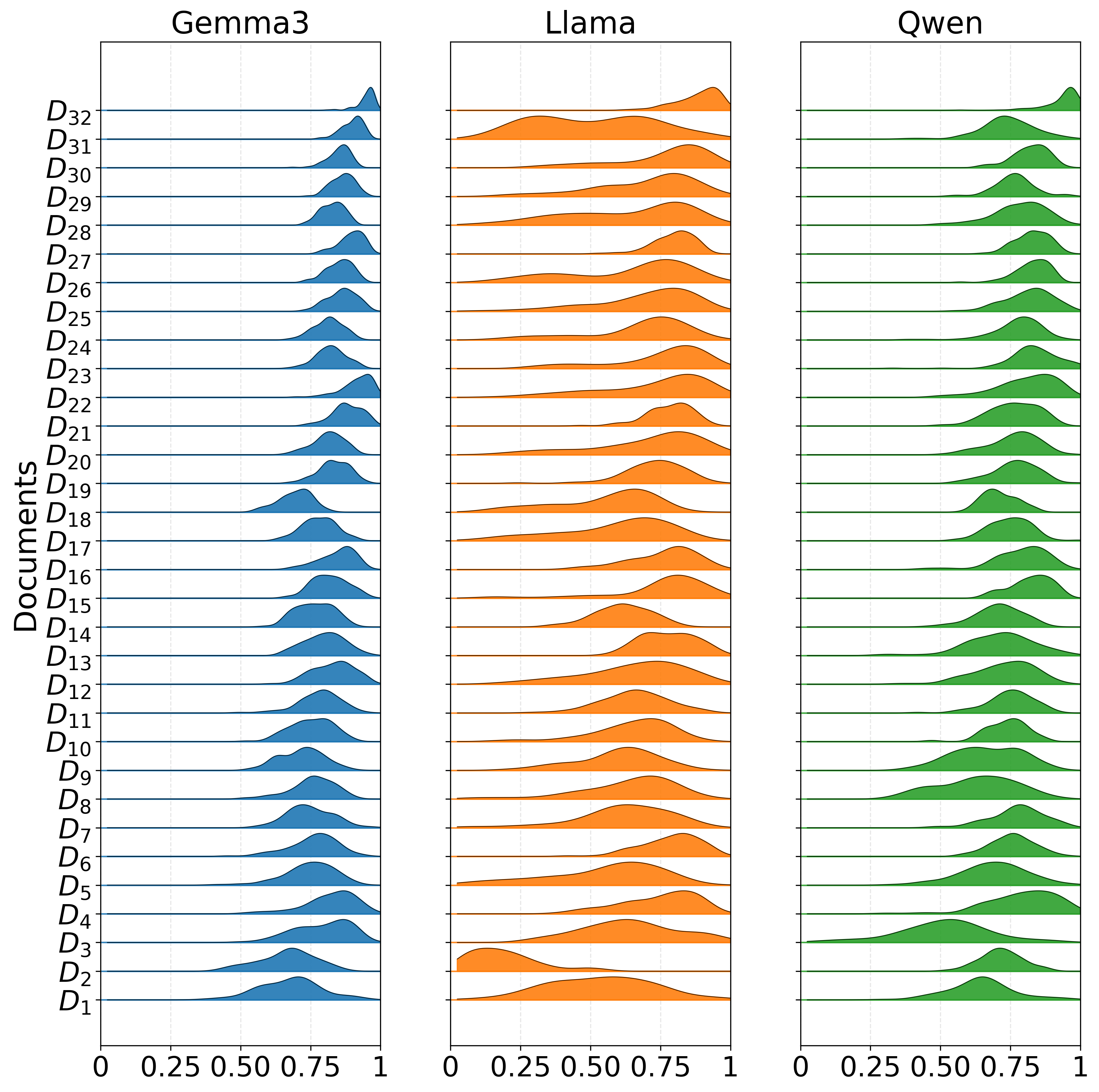}
        \caption{AlignScore}
        \label{fig:pubmed-alignscore-ridge}
    \end{subfigure}  
    \hfill
    \begin{subfigure}[b]{0.32\textwidth}
        \centering
        \includegraphics[width=\textwidth]{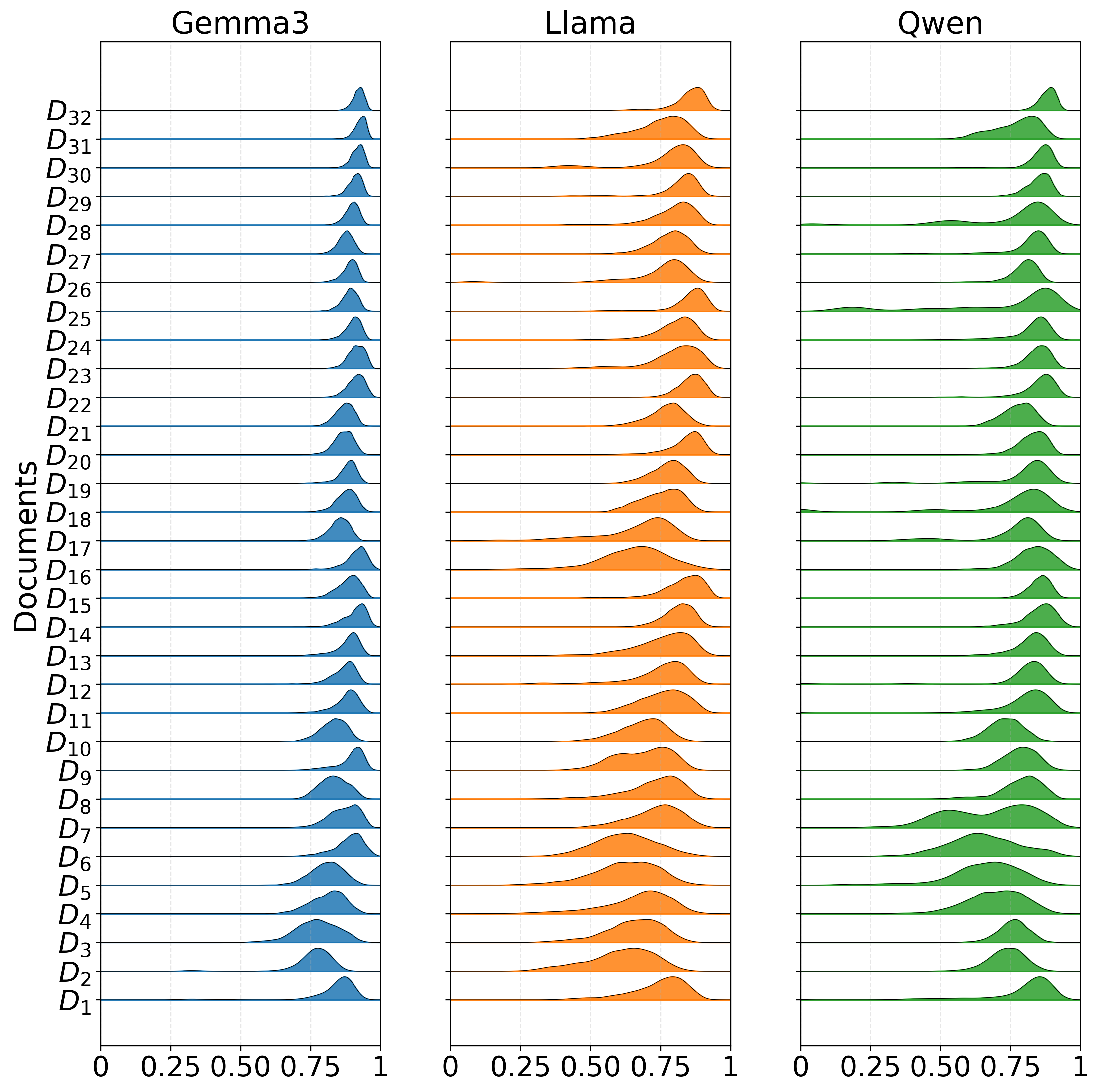}
        \caption{SC Score}
    \end{subfigure}
    
    \textbf{PubMed} \\[4mm]
    \vspace{4mm}
    \begin{subfigure}[b]{0.32\textwidth}
        \centering
        \includegraphics[width=\textwidth]{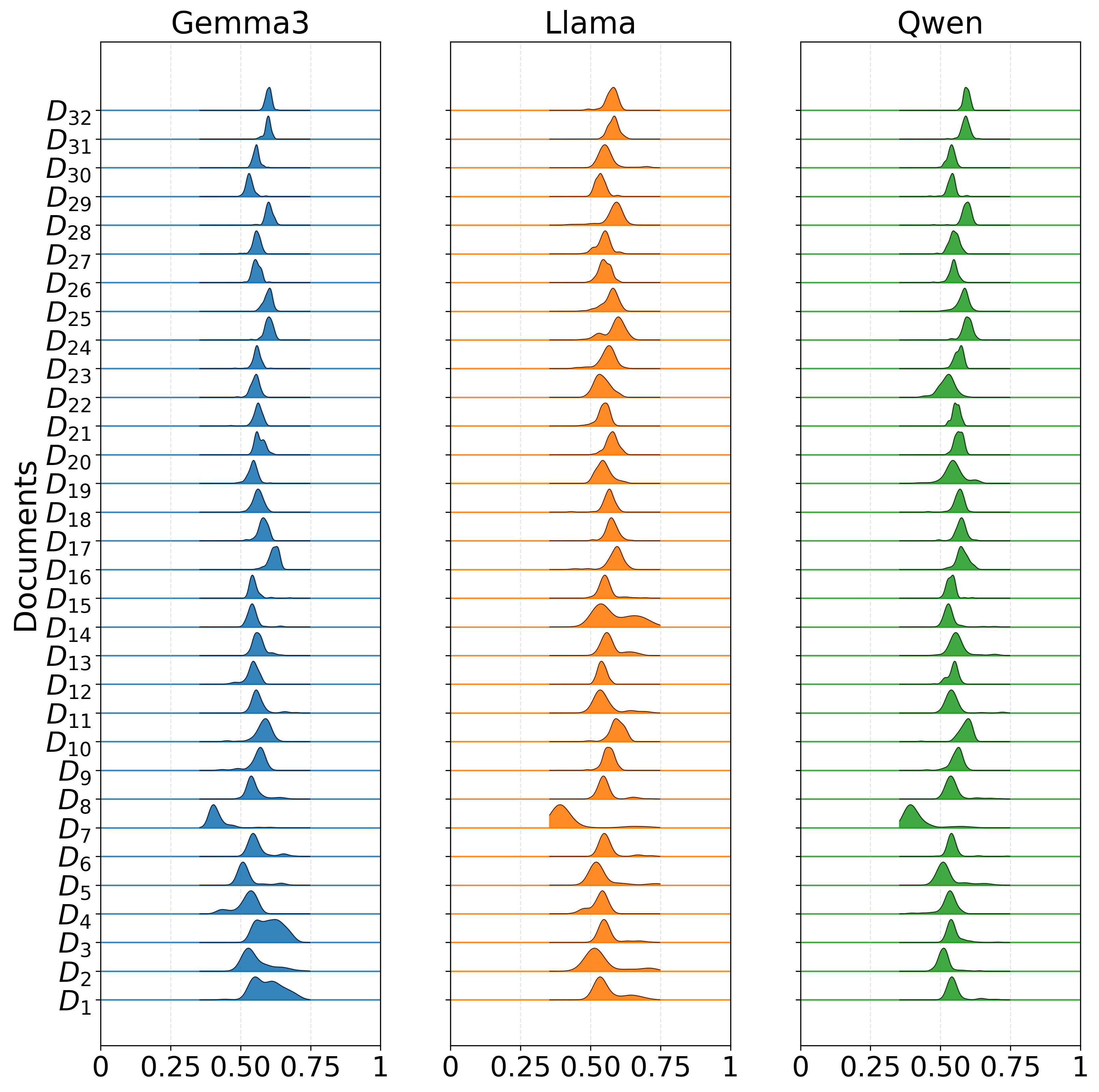}
        \caption{BERTScore}
    \end{subfigure}
    \hfill
    \begin{subfigure}[b]{0.32\textwidth}
        \centering
        \includegraphics[width=\textwidth]{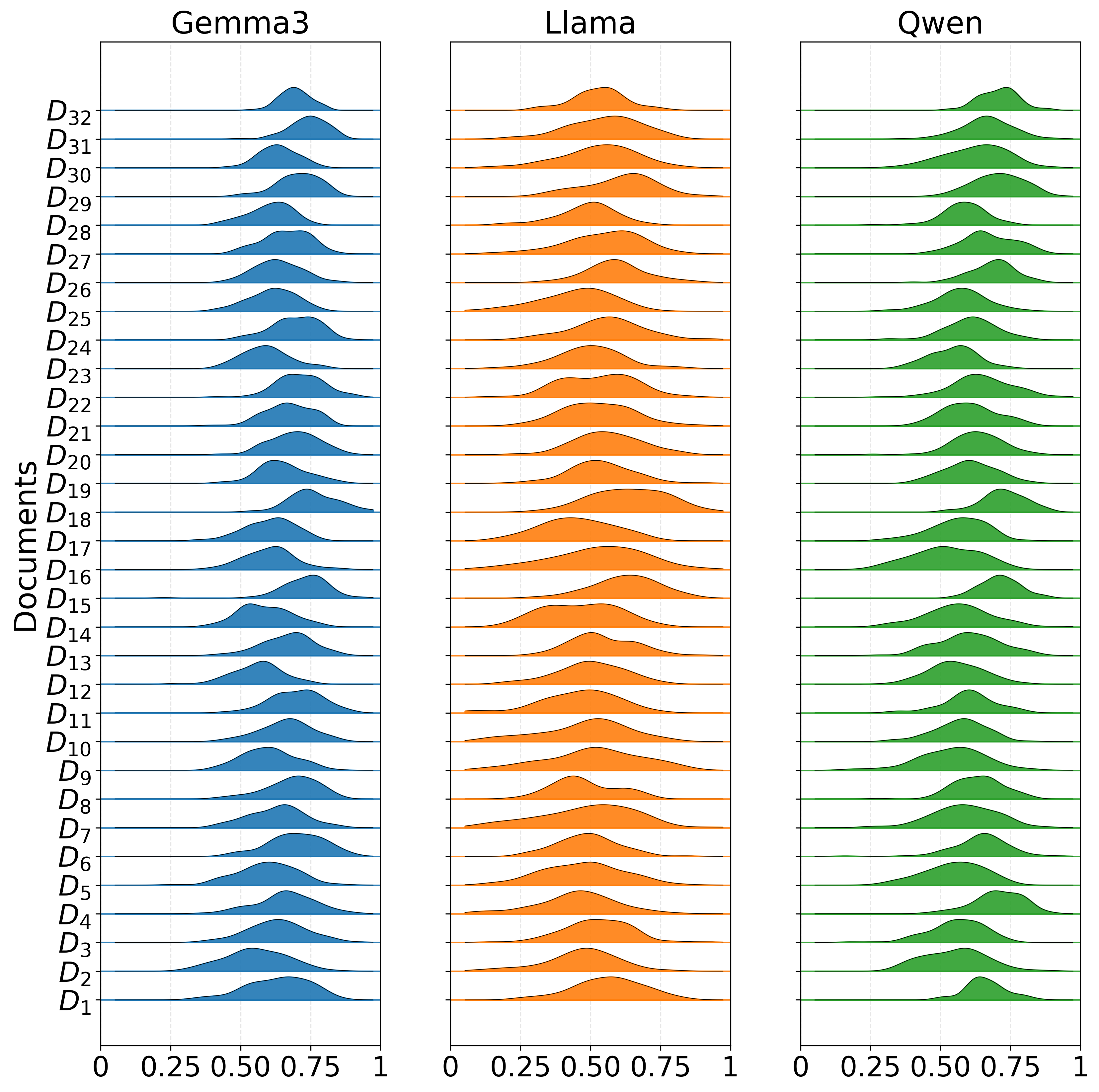}
        \caption{AlignScore}
        \label{fig:in-abs-alignscore-ridge}
    \end{subfigure}
    \hfill
    \begin{subfigure}[b]{0.32\textwidth}
        \centering
        \includegraphics[width=\textwidth]{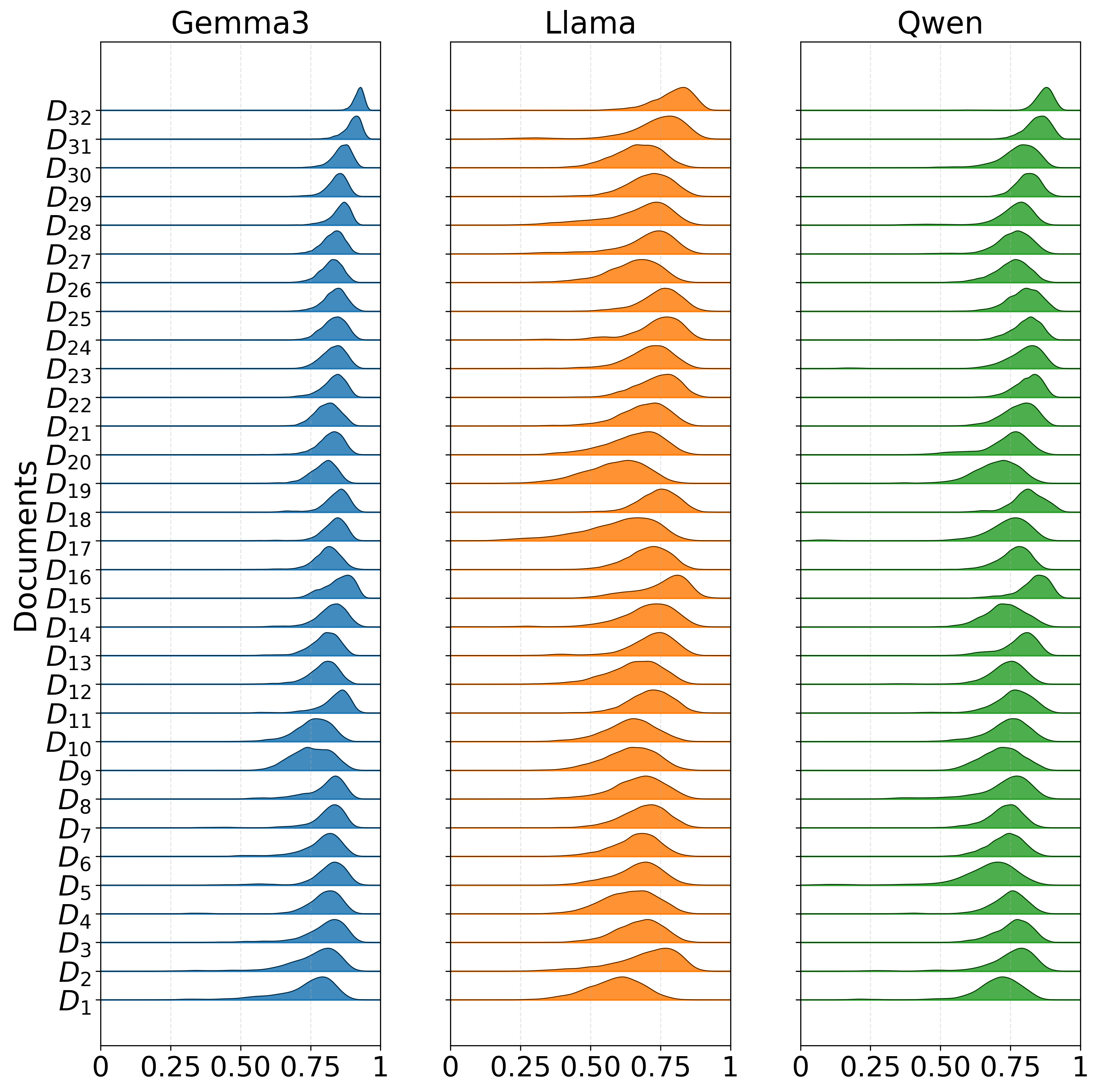}
        \caption{SC Score}
    \end{subfigure}
    
    \textbf{IN-Abs}
    \caption{Ridgeline plots of evaluation score distributions for the $32$ documents ($D_1$-$D_{32}$) from PubMed and IN-Abs datasets. Each ridgeline represents the  distribution of scores of hundred summaries for a document. Distributions of Gemma3, Llama, and Qwen are juxtaposed for comparison across each metric.}
    \label{fig:app-pubmed-in-abs-ridge-plots}
\end{figure}
We visualize the distribution of evaluation scores using Ridgeline plots for the sample of the PubMed and IN-Abs datasets (Fig.~\ref{fig:app-pubmed-in-abs-ridge-plots}). Each ridgeline represents the distribution of scores obtained for a single document. We juxtapose the score distributions produced by the Gemma3, Llama, and Qwen models to facilitate direct comparison across models for each evaluation metric. These plots provide a visual characterization of the variability in summary quality across documents and models.
\subsection{Shapiro-Wilk Test for Normality}
This statistical test confirms \textit{normality} of data distribution. Considering  100 metric scores corresponding to the generated summaries,  Table~\ref{tab:shapiro-wilk-test-5} shows the results of Shapiro-Wilk test for 32 distributions for three metrics, LLM-summarizers  for all sample sets. The null hypotheses ($H_0$: The data follows normal distribution) are tested at $5\%$ level of significance for each case. 
\begin{table}[width=.6\linewidth,cols=5,pos=h]
\caption{Results of Shapiro-Wilk Normality Test: Number of documents for which the null hypothesis is \textit{rejected} at $5\%$ level of significance.}
\label{tab:shapiro-wilk-test-5}
\begin{tabular*}{\tblwidth}{l l c cc}\toprule   
     \textbf{Datasets}&   \textbf{LLM} & \textbf{BERTScore}  & \textbf{AlignScore} & \textbf{SC Score}\\ \midrule
     & Gemma3 & 11/32&30/32 & 32/32\\
     \textbf{CNN/DM} & Llama & 15/32&25/32 & 31/32\\
     & Qwen & 14/32&26/32 & 32/32 \\ \midrule
     & Gemma3 & 16/32& 32/32&32/32\\
    \textbf{PubMed} & Llama & 23/32&30/32 & 32/32\\
     & Qwen & 26/32&30/32 & 32/32\\  \midrule
      & Gemma3 & 27/32&6/32 & 32/32\\
    \textbf{IN-Abs} & Llama & 26/32&6/32 & 32/32\\
     & Qwen & 27/32&6/32  & 32/32\\ 
     \bottomrule
    \end{tabular*}
\end{table}
\subsection{Combined coefficient of variation and Gini Coefficient} 
\begin{figure}
    \centering
    \begin{subfigure}[b]{0.48\textwidth}
        \centering
        \includegraphics[width=\textwidth]{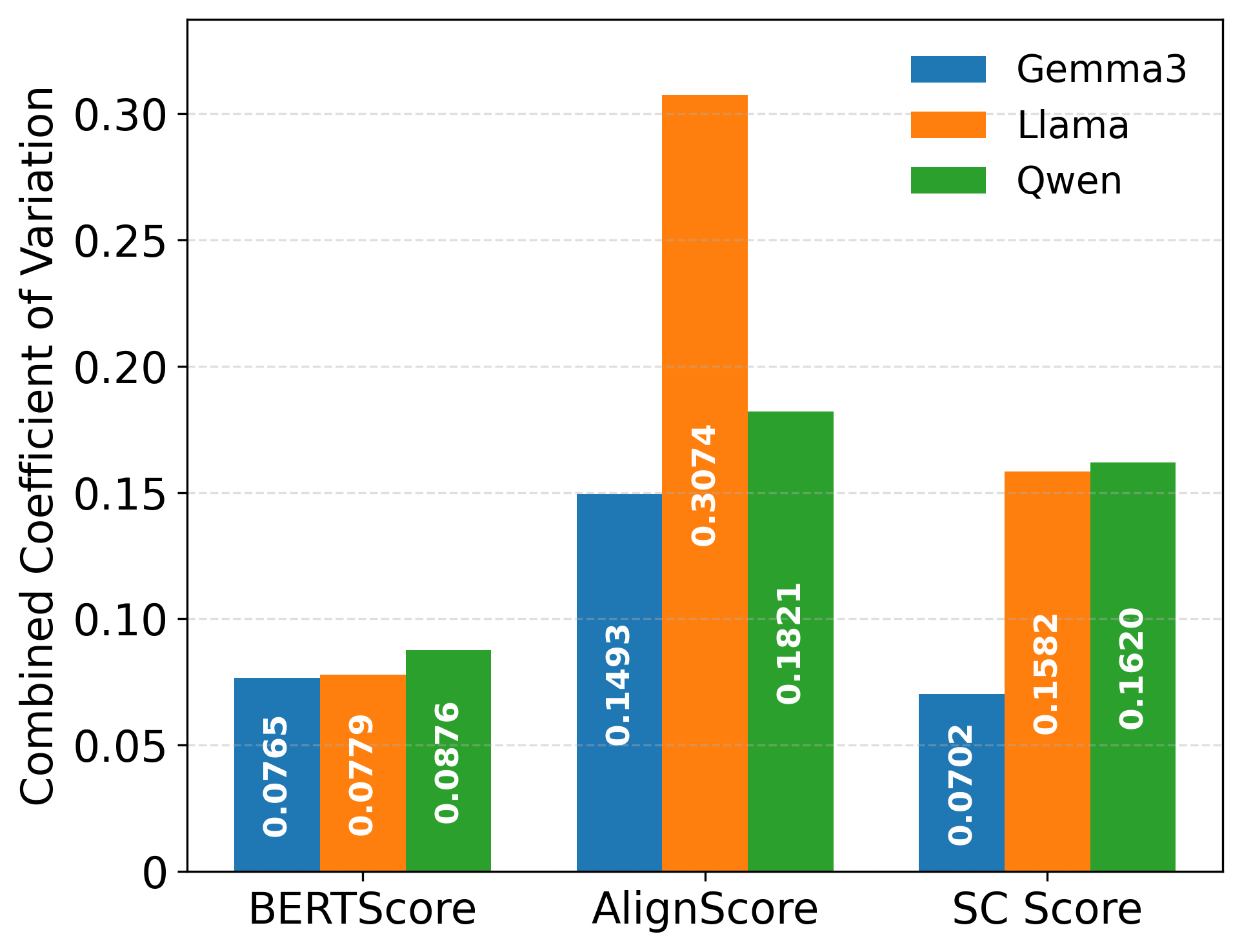}
        \caption{PubMed}
         \label{fig:app-combined-cv-pubmed}
    \end{subfigure}
    \hfill
    \begin{subfigure}[b]{0.48\textwidth}
        \centering
        \includegraphics[width=\textwidth]{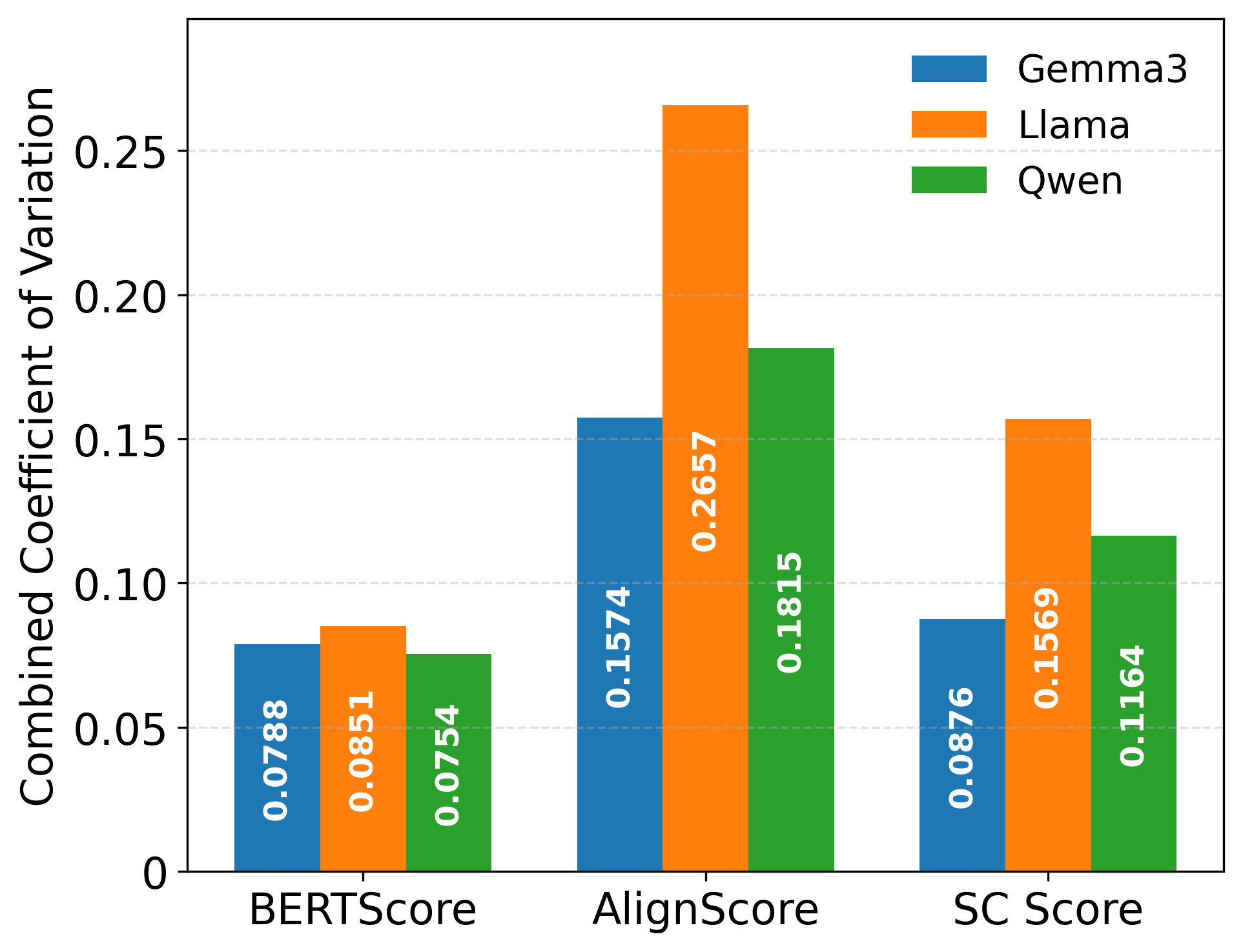}
        \caption{IN-Abs}
        \label{fig:app-combined-cv-in-abs}
    \end{subfigure}
    \caption{Combined coefficients of variation for the three metrics for  (a) PubMed and (b) IN-Abs datasets.}
    \label{fig:app-combined-cv}
\end{figure}

Figure~\ref{fig:app-combined-cv} illustrates a comparative analysis of the $CV_c$ values across the three evaluation metrics for different LLM-summarizers, using both the PubMed and IN-Abs sample sets. This comparison highlights variations in the metric scores across models and datasets.

\begin{figure}
    \centering
    \begin{subfigure}[b]{0.48\textwidth}
        \centering
        \includegraphics[width=\textwidth]{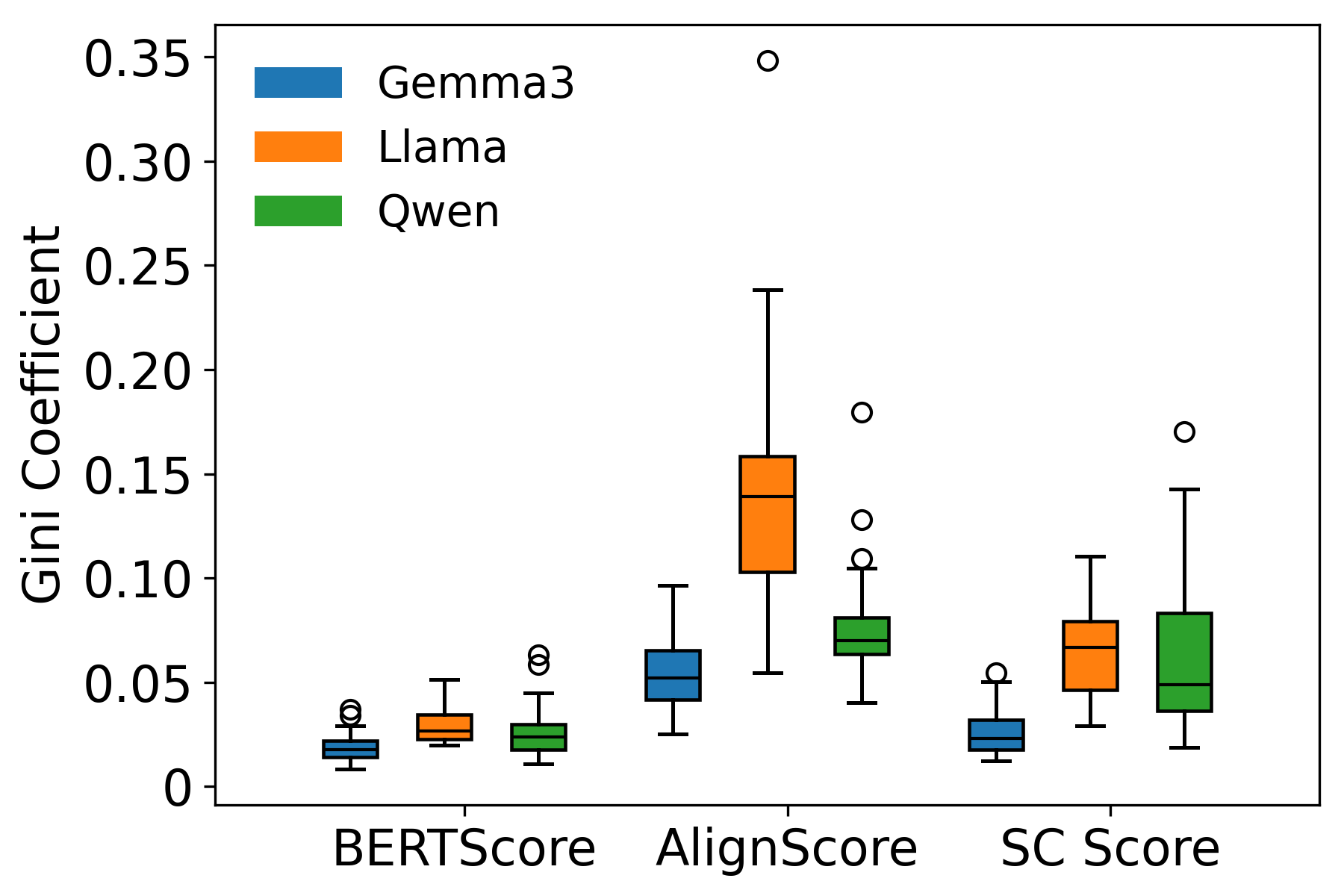}
        \caption{PubMed}
        \label{fig:app-pubmed-gini-box-plot}
    \end{subfigure}
    \hfill
    \begin{subfigure}[b]{0.48\textwidth}
        \centering
        \includegraphics[width=\textwidth]{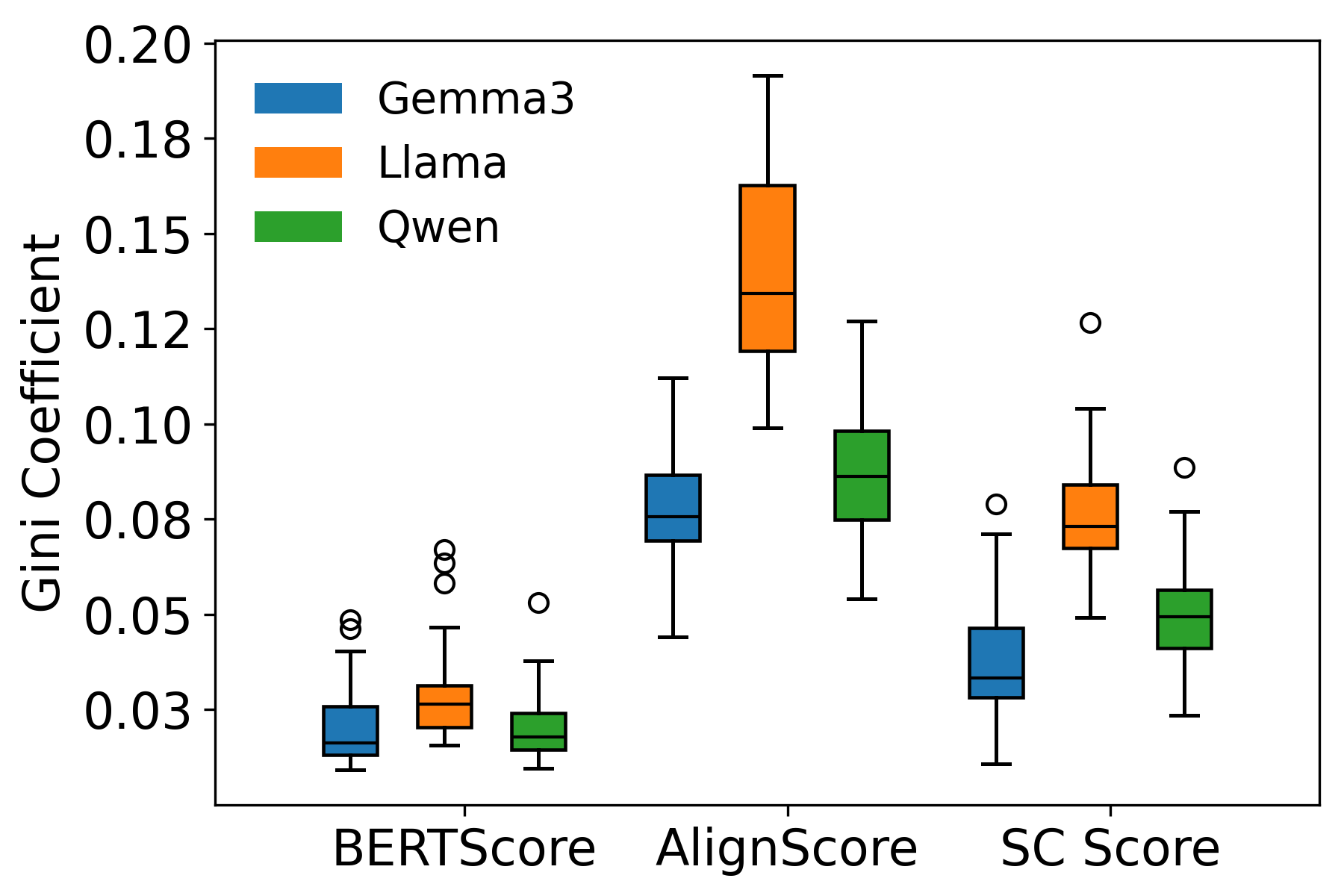}
        \caption{IN-Abs}
        \label{fig:app-in-abs-gini-box-plot}
    \end{subfigure}
    \caption{Box plots of 32 document-level Gini Coefficients of metric scores for  (a) PubMed and (b) IN-Abs datasets.}
    \label{fig:app-gini-box-plot}
\end{figure}

Figure~\ref{fig:app-gini-box-plot} presents the boxplots of the Gini coefficient values computed over 32 documents in the PubMed and IN-Abs sample sets. GC values provide a visual summary of inequality in score distributions and enabling a clearer assessment of variability and spread within generated summaries of each document.

\section{Benchmarking LLM Summarizers}
\label{appendix:llm-level-stability-analysis}
This section presents results for the PubMed and IN-Abs sample sets, comparing LLM summarizers across evaluation metrics and highlight performance differences.
\subsection{Rank Heatmap} 
\begin{figure}
    \centering
    \begin{subfigure}[b]{0.32\textwidth}
        \centering
        \includegraphics[width=\textwidth]{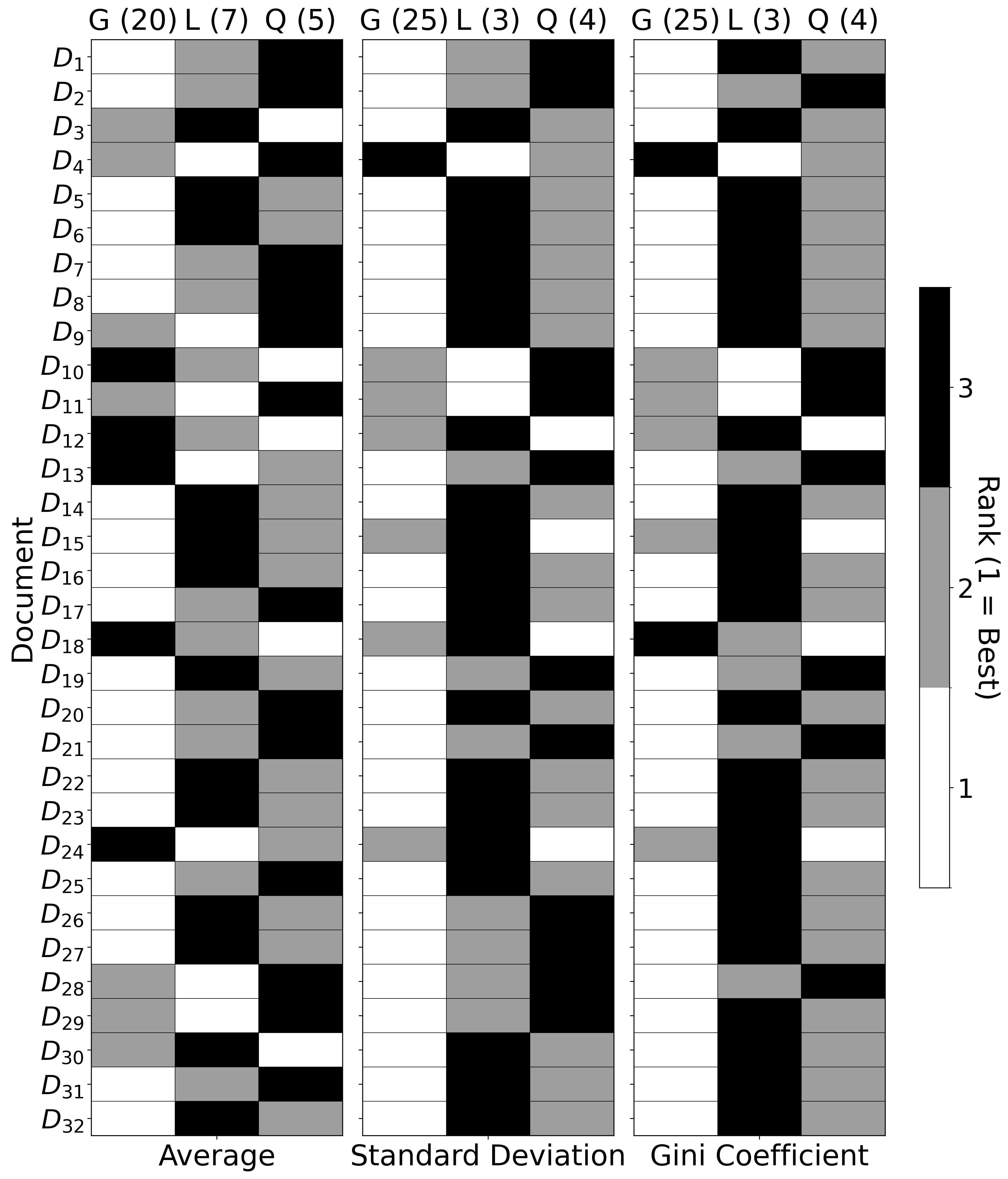}
        \caption{BERTScore}
    \end{subfigure}
    \hfill
    \begin{subfigure}[b]{0.32\textwidth}
        \centering
        \includegraphics[width=\textwidth]{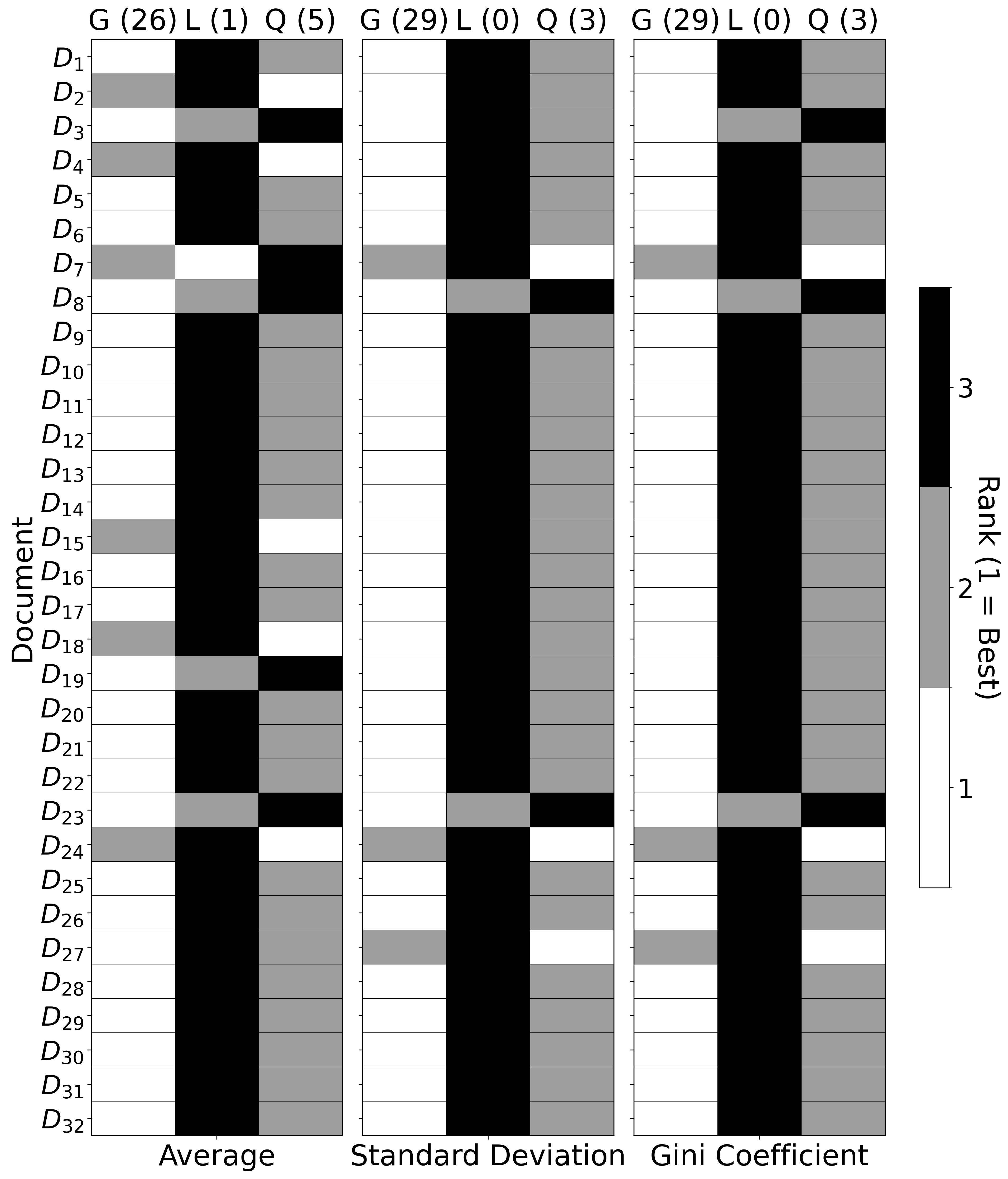}
        \caption{AlignScore}
    \end{subfigure}
    \hfill
    \begin{subfigure}[b]{0.32\textwidth}
        \centering
        \includegraphics[width=\textwidth]{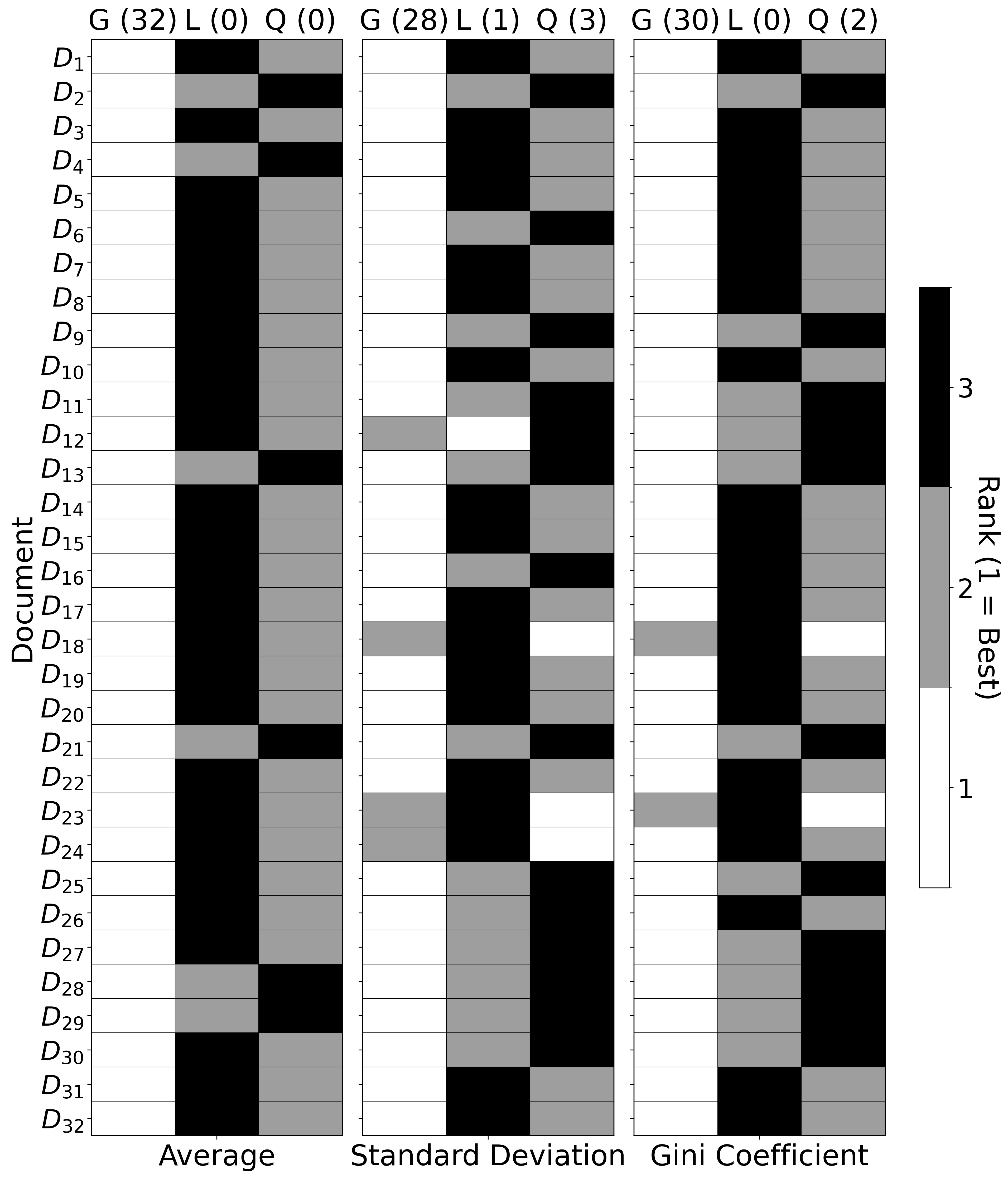}
        \caption{SC Score}
    \end{subfigure}
    \textbf{PubMed} \\[2mm]
    \vspace{4mm}
    \begin{subfigure}[b]{0.32\textwidth}
        \centering
        \includegraphics[width=\textwidth]{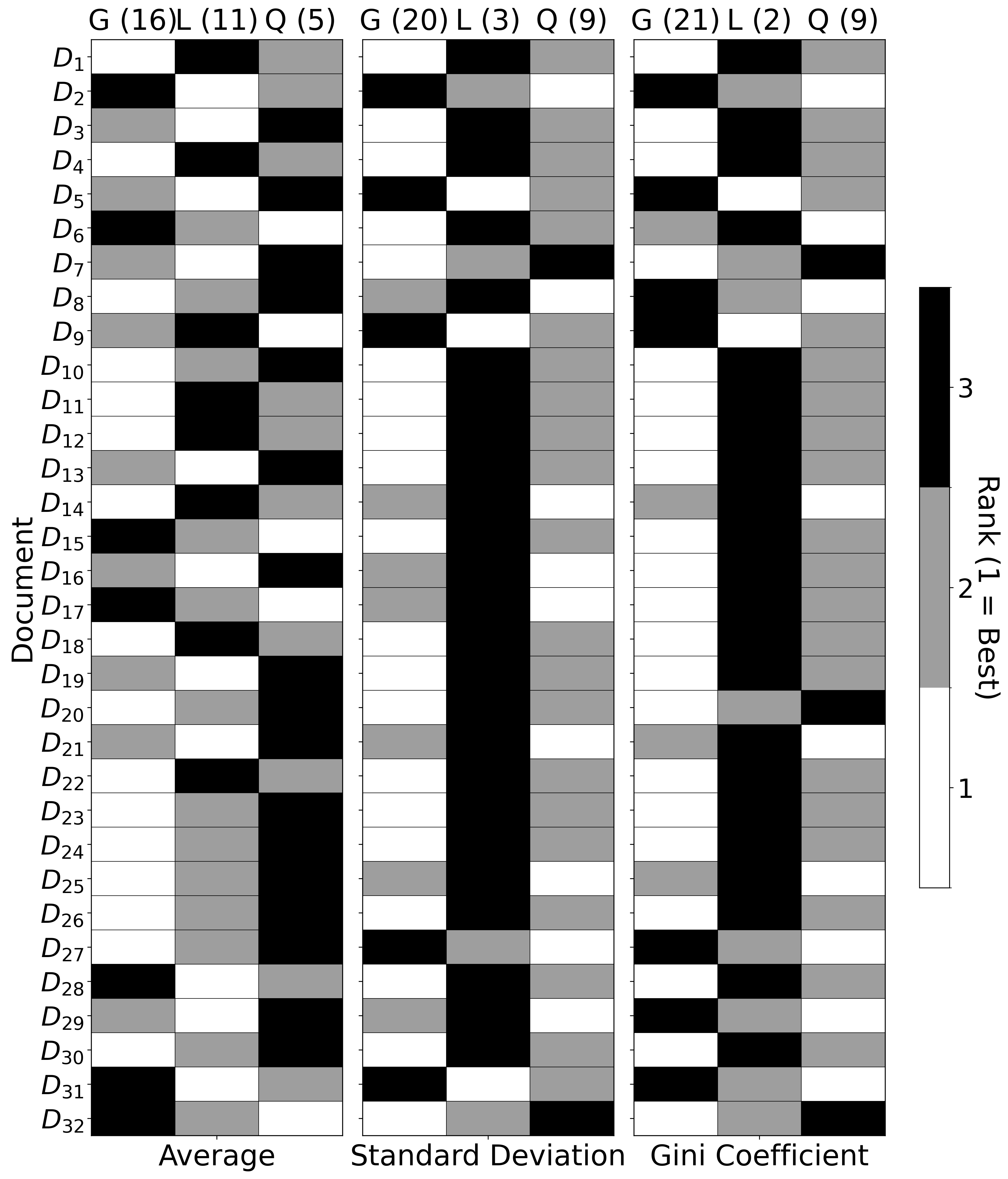}
        \caption{BERTScore}
    \end{subfigure}
    \hfill
    \begin{subfigure}[b]{0.32\textwidth}
        \centering
        \includegraphics[width=\textwidth]{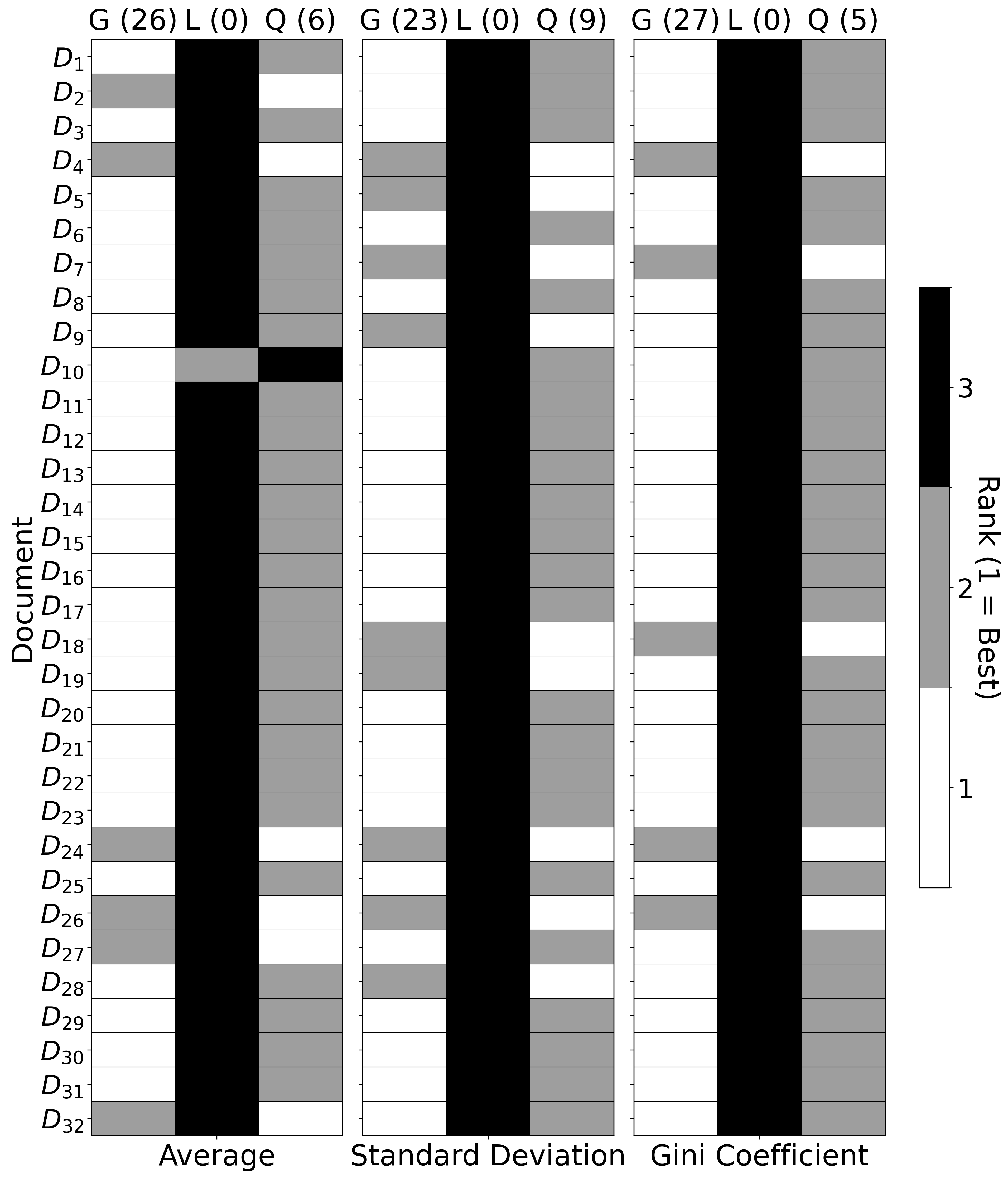}
        \caption{AlignScore}
    \end{subfigure}
    \hfill
    \begin{subfigure}[b]{0.32\textwidth}
        \centering
        \includegraphics[width=\textwidth]{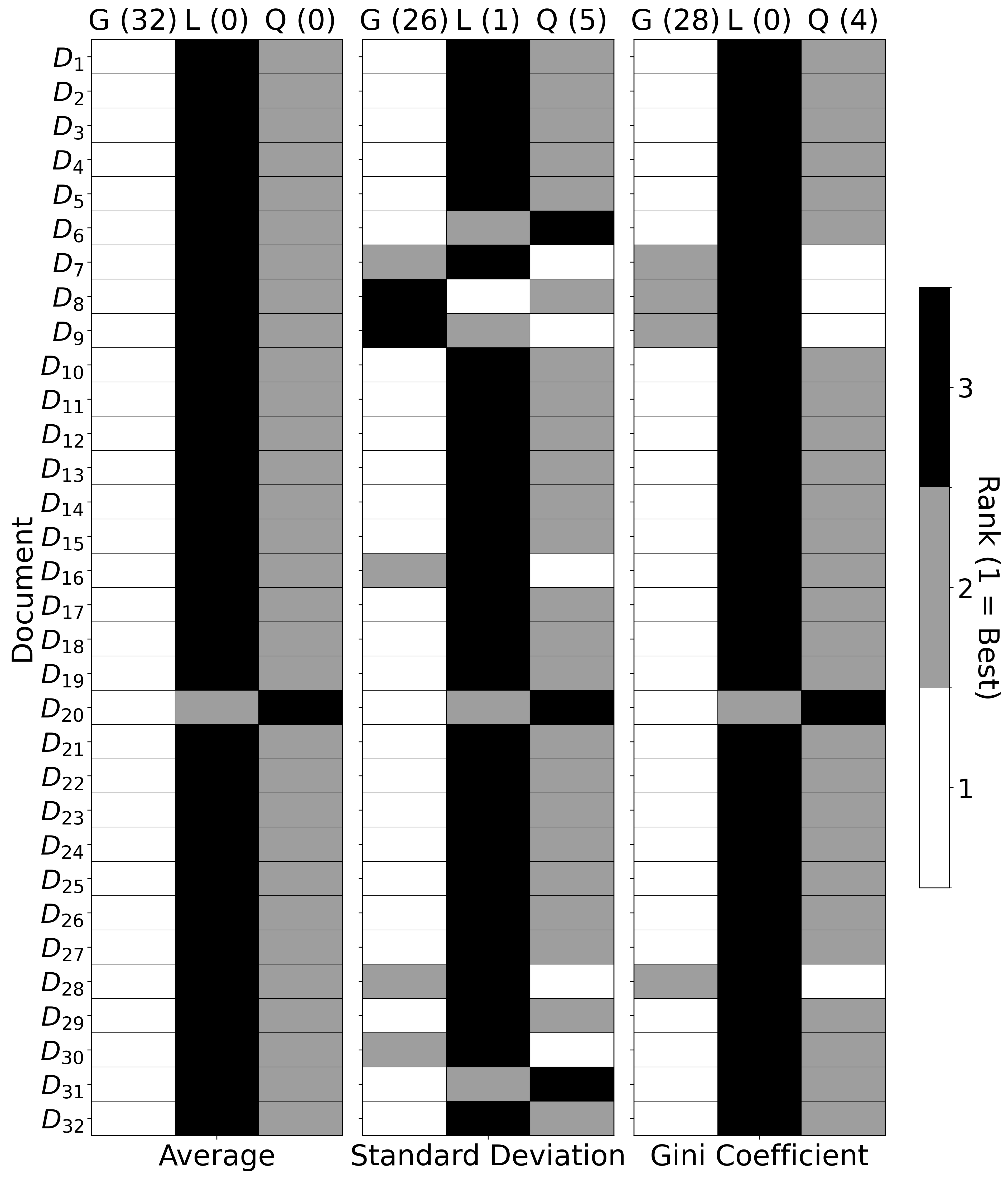}
        \caption{SC Score}
    \end{subfigure}
    \textbf{IN-Abs} \\
    \caption{Document-wise rank heatmaps for the three LLMs for  the PubMed and IN-Abs summaries. For each document ($D_1$–$D_{32}$), models are ranked  on Average score (left), Standard Deviation (middle), and Gini Coefficient (right). Rank~1 denotes the best-performing model (white), Rank~2 the intermediate (gray), and Rank~3 the worst-performing model (black). The numbers in parentheses above each column—G(•), L(•), and Q(•) — denote the number of documents for which the corresponding model achieved Rank~1 under the given metric.}
    \label{fig:app-pubmed-in-abs-ranked-heatmap-plots}
\end{figure}

Figure~\ref{fig:app-pubmed-in-abs-ranked-heatmap-plots} shows the document-wise ranked heatmap for PubMed and IN-Abs sample set to compare the three LLMs based on average, standard deviation and Gini Coefficient. The win count on top of the heatmaps show that the Gemma3 as the best summarizer for both the datasets.

\subsection{Average Ranks of LLM-summarizers}
We report the average ranks of all summarizers for the three metrics in Table~\ref{tab:pubmed-in-abs-average-rank-llm}.
\begin{table}[width=.6\linewidth,cols=5,pos=h]
 \caption{Average ranks of LLM-summarizers based on the Coefficient of variation (CV) of metric scores for PubMed and IN-Abs datasets.}
    \label{tab:pubmed-in-abs-average-rank-llm}
    \begin{tabular*}{\tblwidth}{llccc}
    \toprule
          \textbf{Dataset} & \textbf{LLM} &  \textbf{BERTScore}& \textbf{AlignScore} &  \textbf{SC Score}\\ \midrule
          &\textbf{Gemma3}&  \textbf{1.31}& \textbf{1.06} &  \textbf{1.09}  \\
          \textbf{PubMed} &\textbf{Llama}&  2.44&  2.91 &  2.56 \\ 
          &\textbf{Qwen} &  2.25& 2.03 &  2.34\\  \midrule
          &\textbf{Gemma3}&  \textbf{1.53} &  \textbf{1.16}  &  \textbf{ 1.16}\\
          \textbf{IN-Abs} &\textbf{Llama}&  2.66& 3.00 &  2.91\\
          &\textbf{Qwen}&  1.81& 1.84 &  1.94\\ 
         \bottomrule
    \end{tabular*}
\end{table}

\begin{figure}
    \centering
    \begin{subfigure}{0.32\linewidth}
        \centering
        \includegraphics[width=\linewidth]{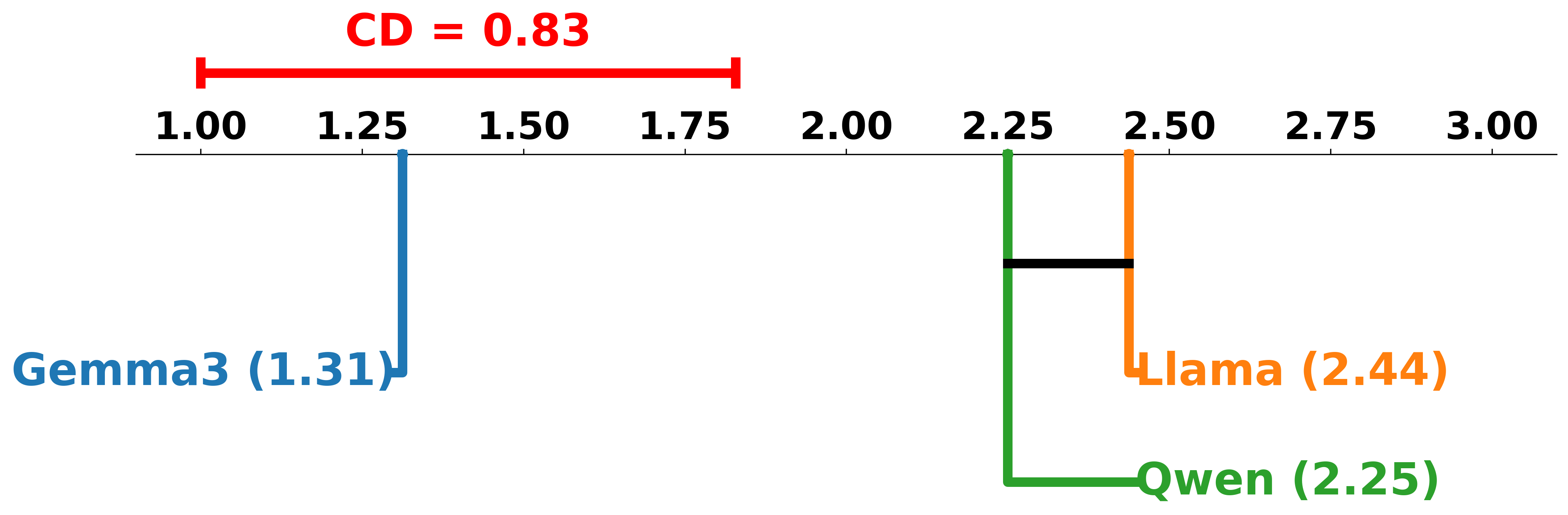}
        \caption{BERTScore}
        \label{fig:pubmed-cd-plot-bertScore}
    \end{subfigure}
    \hfill
    \begin{subfigure}{0.32\linewidth}
        \centering
        \includegraphics[width=\linewidth]{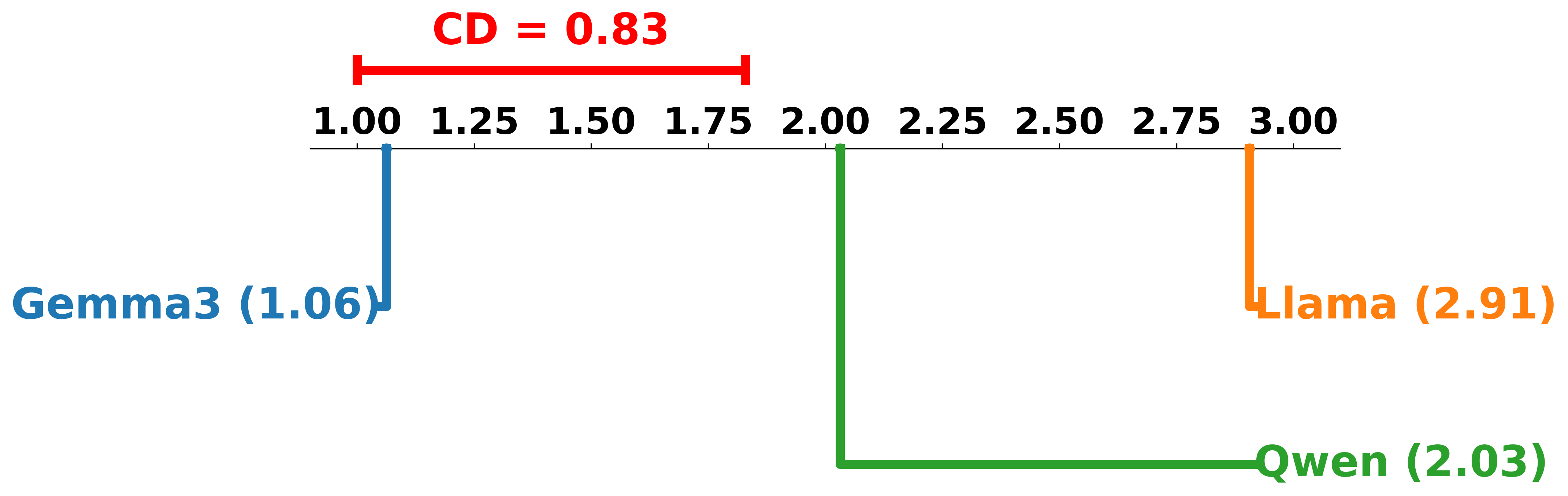}
        \caption{AlignScore}
        \label{fig:pubmed-cd-plot-alignScore}
    \end{subfigure}
    \hfill
    \begin{subfigure}{0.32\linewidth}
        \centering
        \includegraphics[width=\linewidth]{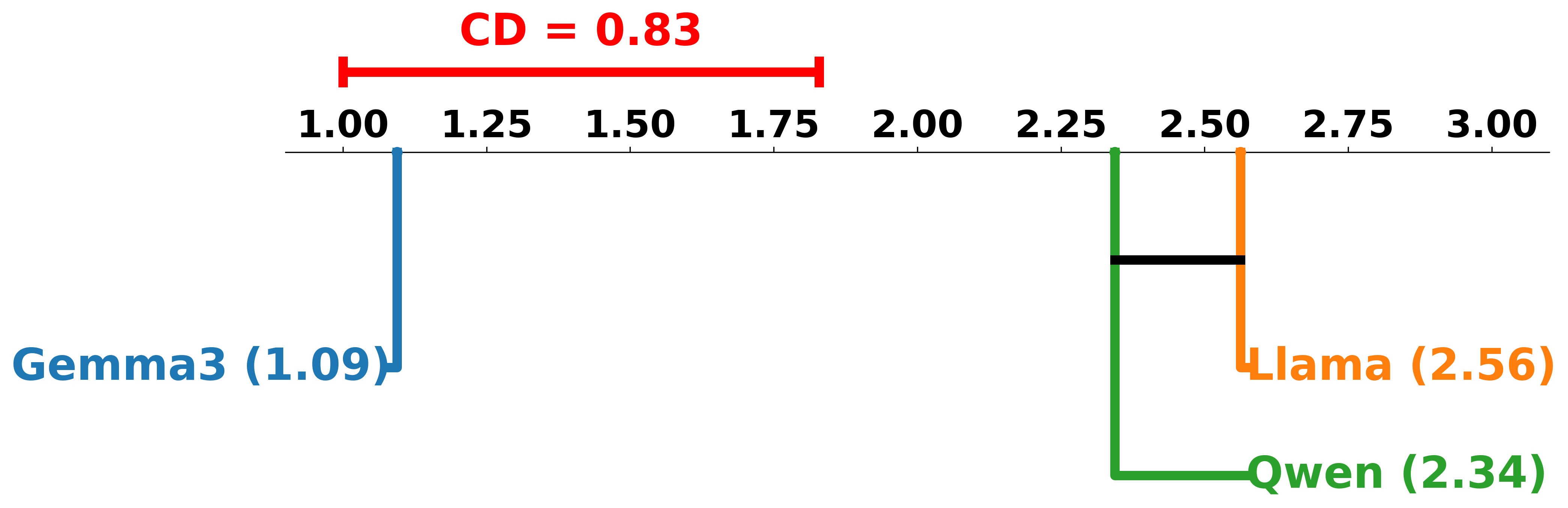}
        \caption{SC Score}
        \label{fig:pubmed-cd-plot-scScore}
    \end{subfigure}
    \textbf{PubMed} \\[2mm]
    
    \vspace{4mm}

       
    \begin{subfigure}{0.32\linewidth}
        \centering
        \includegraphics[width=\linewidth]{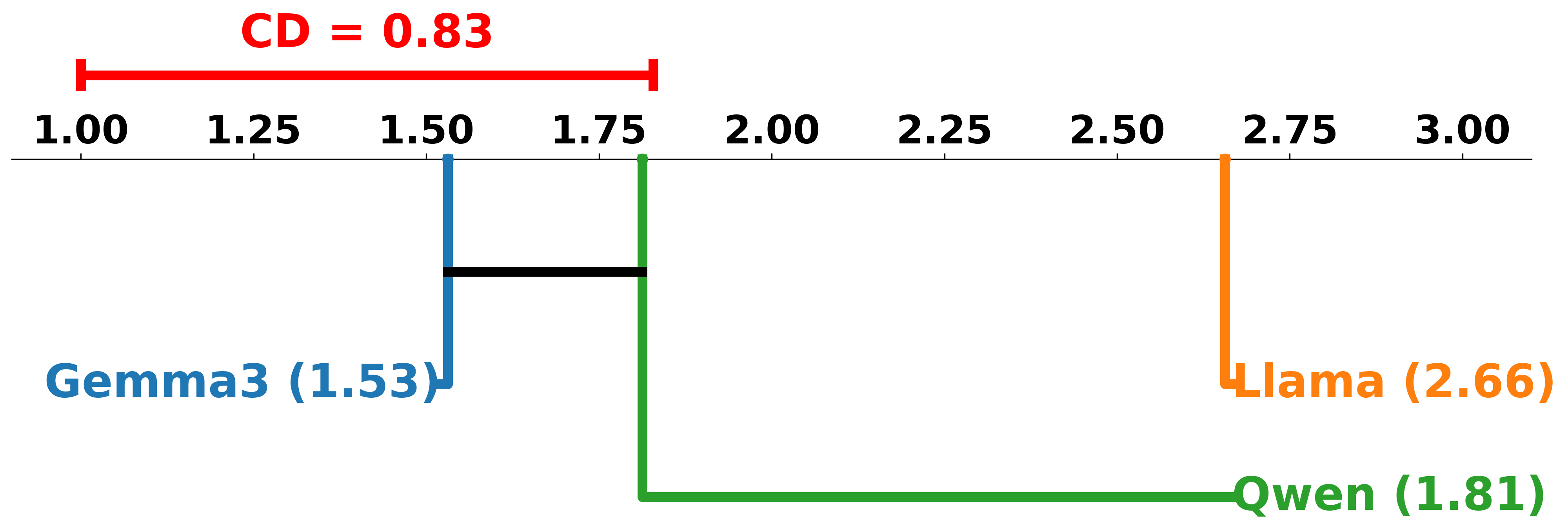}
        \caption{BERTScore}
        \label{fig:in-abs-cd-plot-bertScore}
    \end{subfigure}
     \hfill
    \begin{subfigure}{0.32\linewidth}
        \centering
        \includegraphics[width=\linewidth]{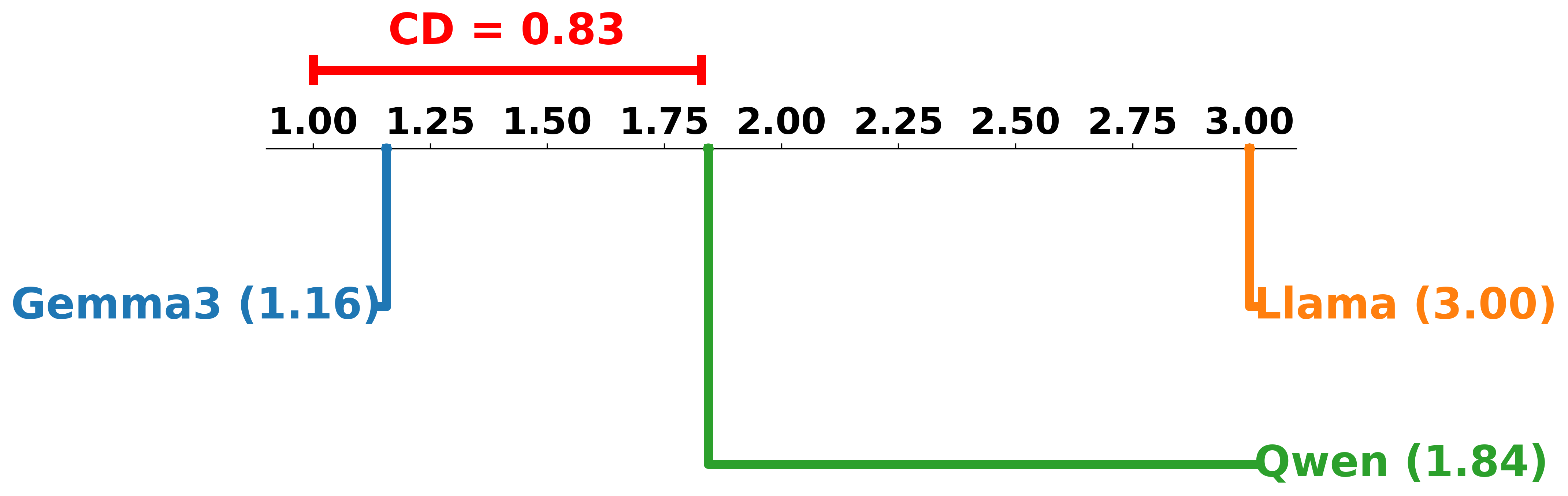}
        \caption{AlignScore}
        \label{fig:app-in-abs-cd-plot-alignScore}
    \end{subfigure}
    \hfill
    \begin{subfigure}{0.32\linewidth}
        \centering
        \includegraphics[width=\linewidth]{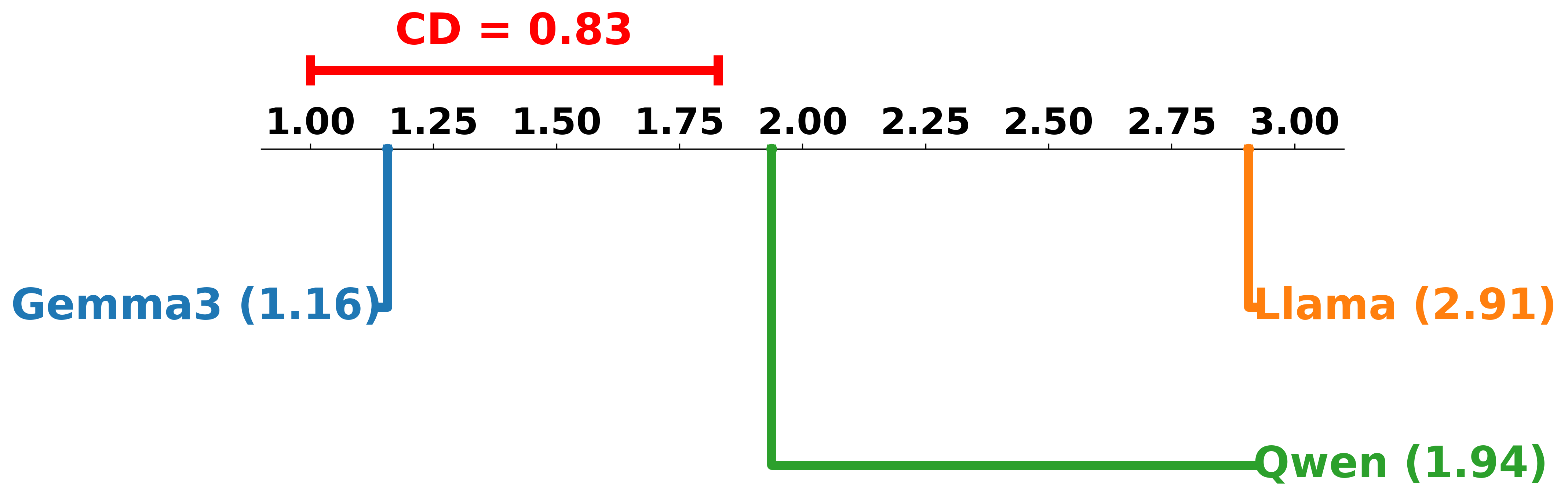}
        \caption{SC Score}
        \label{fig:in-abs-cd-plot-scScore}
    \end{subfigure}
    \textbf{IN-Abs} \\
    \caption{Critical difference (CD) diagrams showing the average ranks of LLM-summarizers based on BERTScore, Semantic Consistency (SC) Score, and AlignScore for the three datasets: (a) PubMed (medical documents), and (b) IN-Abs (legal judgments). The horizontal line denotes the critical difference (CD). Models connected by this line do not differ significantly according to the Nemenyi post-hoc test following the Friedman test.}
    \label{fig:app-cd-plots-pubmed-inabs}
\end{figure}

Figure~\ref{fig:app-cd-plots-pubmed-inabs} shows the CD diagrams for PubMed and IN-Abs sample sets. For the PubMed documents,  Gemma3 achieves the highest average rank for all metrics,which indicates (i) Gemma3 summaries better capture semantic information with reference to gold summary, (ii) better semantic consistency among the generated summaries, and (iii) best factual alignment with source documents. Pairs  <Gemma3, Llama> and <Gemma3, Qwen> display statistically significant differences in their ranks for all metrics. However, the average ranks of Llama and Qwen have no significant difference for semantic measures, though significant difference is observed for AlignScore metric. For the IN-Abs sample documents, Gemma3 maintains the highest rank for all three metrics, and Qwen is a close competitor for BERTScore. Llama and Qwen show significant difference in ranks for all three measures. 
\begin{table}[width=.46\linewidth,cols=4,pos=t]
\caption{Metric-wise stability coefficients ($\mathcal{E}$)  of Gemma3, Llama and Qwen for PubMed and IN-Abs datasets.}
\label{tab:pubmed-in-abs-epsilon-stability}
    \begin{tabular*}{\tblwidth}{llcc}
    \toprule
          \textbf{Dataset} & \textbf{LLM} &  \textbf{BERTScore}& \textbf{AlignScore}\\ \midrule
                   &\textbf{Gemma3}&  \textbf{0.8036}  & \textbf{0.4345} \\
    \textbf{PubMed} &\textbf{Llama}&  0.6834 & 0.0893 \\
                    &\textbf{Qwen} &  0.3925 & 0.1624 \\  \midrule
                    &\textbf{Gemma3}&  \textbf{0.6668} & \textbf{0.2725} \\
    \textbf{IN-Abs} &\textbf{Llama}&  0.5660 & 0.1618 \\
                    &\textbf{Qwen}&  0.6477 & 0.2352 \\ 
         \bottomrule
    \end{tabular*}
\end{table}
\subsection{Metric-wise stability coefficients}
Table~\ref{tab:pubmed-in-abs-epsilon-stability} shows the metric-wise stability coefficients for PubMed and IN-Abs sample set on all metrics and summarizers.
\clearpage
\bibliographystyle{cas-model2-names}

\bibliography{custom}

@inproceedings{23-GPT-consistency,
  title={Consistency analysis of chatgpt},
  author={Jang, Myeongjun and Lukasiewicz, Thomas},
  booktitle={Proceedings of the 2023 Conference on Empirical Methods in Natural Language Processing},
  pages={15970--15985},
  year={2023}
}

@article{2025-zhang-survey,
  title={A systematic survey of text summarization: From statistical methods to large language models},
  author={Zhang, Haopeng and Yu, Philip S and Zhang, Jiawei},
  journal={ACM Computing Surveys},
  volume={57},
  number={11},
  pages={1--41},
  year={2025},
  publisher={ACM New York, NY}
}

@inproceedings{survey-dhaini2024explainability,
  title={Explainability meets text summarization: A survey},
  author={Dhaini, Mahdi and Erdogan, Ege and Bakshi, Smarth and Kasneci, Gjergji},
  booktitle={Proceedings of the 17th International Natural Language Generation Conference},
  pages={631--645},
  year={2024}
}

@article{survey-akter2025comprehensive,
  title={A comprehensive survey on legal summarization: Challenges and future directions},
  author={Akter, Mousumi and {\c{C}}ano, Erion and Weber, Erik and Dobler, Dennis and Habernal, Ivan},
  journal={ACM Computing Surveys},
  year={2025},
  publisher={ACM New York, NY}
}

@inproceedings{survey-liu2024sum,
  title={SumSurvey: An abstractive dataset of scientific survey papers for long document summarization},
  author={Liu, Ran and Liu, Ming and Yu, Min and Zhang, He and Jiang, Jianguo and Li, Gang and Huang, Weiqing},
  booktitle={Findings of the ACL 2024},
  pages={9632--9651},
  year={2024}
}

@article{survey-2024-zhang,
  title={A comprehensive survey on process-oriented automatic text summarization with exploration of llm-based methods},
  author={Zhang, Yang and Jin, Hanlei and Meng, Dan and Wang, Jun and Tan, Jinghua},
  journal={arXiv preprint arXiv:2403.02901},
  year={2024}
}

@inproceedings{CNN-Dailymail-2016,
  title={Abstractive text summarization using sequence-to-sequence rnns and beyond},
  author={Nallapati, Ramesh and Zhou, Bowen and Dos Santos, Cicero and Gul{\c{c}}ehre, {\c{C}}a{\u{g}}lar and Xiang, Bing},
  booktitle={Proceedings of the 20th SIGNLL conference on computational natural language learning},
  pages={280--290},
  year={2016}
}

@inproceedings{pubmed-2018,
  title={A Discourse-Aware Attention Model for Abstractive Summarization of Long Documents},
  author={Cohan, Arman and Dernoncourt, Franck and Kim, Doo Soon and Bui, Trung and Kim, Seokhwan and Chang, Walter and Goharian, Nazli},
  booktitle={Proceedings of the 2018 Conference of the North American Chapter of the Association for Computational Linguistics: Human Language Technologies, Volume 2 (Short Papers)},
  pages={615--621},
  year={2018}
}

@inproceedings{IN-Abs-2022,
  title={Legal Case Document Summarization: Extractive and Abstractive Methods and their Evaluation},
  author={Shukla, Abhay and Bhattacharya, Paheli and Poddar, Soham and Mukherjee, Rajdeep and Ghosh, Kripabandhu and Goyal, Pawan and Ghosh, Saptarshi},
  booktitle={Proceedings of the 2nd Conference of the Asia-Pacific Chapter of the Association for Computational Linguistics and the 12th International Joint Conference on Natural Language Processing (Volume 1: Long Papers)},
  pages={1048--1064},
  year={2022}
}

@article{2024-legal-deroy,
  title={Applicability of large language models and generative models for legal case judgement summarization},
  author={Deroy, Aniket and Ghosh, Kripabandhu and Ghosh, Saptarshi},
  journal={Artificial Intelligence and Law},
  pages={1--44},
  year={2024},
  publisher={Springer}
}

@inproceedings{2024-legal-hakim,
  title={Case Summarization Using LLM Model for Supreme Court Judgments of an Indian Judiciary System},
  author={Hakim, Yameen and Bhardwaj, Sushil and Mulla, Rais},
  booktitle={International Conference on Information and Communication Technology for Competitive Strategies},
  pages={79--89},
  year={2024},
  organization={Springer}
}

@article{2025-legal-song,
  title={Legal text summarization via judicial syllogism with large language models},
  author={Song, Yumei and Qin, Yongbin and Huang, Ruizhang and Chen, Yanping and Lin, Chuan},
  journal={Journal of King Saud University Computer and Information Sciences},
  volume={37},
  number={5},
  pages={111},
  year={2025},
  publisher={Springer}
}

@article{2024-legal-liu,
  title={Low-resource court judgment summarization for common law systems},
  author={Liu, Shuaiqi and Cao, Jiannong and Li, Yicong and Yang, Ruosong and Wen, Zhiyuan},
  journal={Information Processing \& Management},
  volume={61},
  number={5},
  pages={103796},
  year={2024},
  publisher={Elsevier}
}

@article{2025-legal-mentzingen,
  title={Effectiveness in retrieving legal precedents: exploring text summarization and cutting-edge language models toward a cost-efficient approach},
  author={Mentzingen, Hugo and Ant{\'o}nio, Nuno and Bacao, Fernando},
  journal={Artificial Intelligence and Law},
  pages={1--21},
  year={2025},
  publisher={Springer}
}

@article{2024-news-zhang,
  title={Benchmarking large language models for news summarization},
  author={Zhang, Tianyi and Ladhak, Faisal and Durmus, Esin and Liang, Percy and McKeown, Kathleen and Hashimoto, Tatsunori B},
  journal={Transactions of the Association for Computational Linguistics},
  volume={12},
  pages={39--57},
  year={2024},
  publisher={MIT Press One Broadway, 12th Floor, Cambridge, Massachusetts 02142, USA~…}
}

@inproceedings{2023-news-tam,
  title={Evaluating the factual consistency of large language models through news summarization},
  author={Tam, Derek and Mascarenhas, Anisha and Zhang, Shiyue and Kwan, Sarah and Bansal, Mohit and Raffel, Colin},
  booktitle={Findings of the Association for Computational Linguistics: ACL 2023},
  pages={5220--5255},
  year={2023}
}

@inproceedings{2025-news-premnath,
  title={TechSSN at SemEval-2025 Task 10: A Comparative Analysis of Transformer Models for Dominant Narrative-Based News Summarization},
  author={Premnath, Pooja and Yenumulapalli, Venkatasai Ojus and Mohankumar, Parthiban and Sivanaiah, Rajalakshmi and others},
  booktitle={Proceedings of the 19th International Workshop on Semantic Evaluation (SemEval-2025)},
  pages={2205--2212},
  year={2025}
}

@inproceedings{2025-news-you,
  title={Event-based evaluation of abstractive news summarization},
  author={You, Huiling and Touileb, Samia and {\O}vrelid, Lilja and Velldal, Erik},
  booktitle={Proceedings of the Fourth Workshop on Generation, Evaluation and Metrics (GEM$^2$)},
  pages={504--510},
  year={2025}
}

@phdthesis{2024-sports-kang,
  title={Context-Aware Sports Highlight Generation Leveraging Large Language Models},
  author={Kang, Jeonghun},
  year={2024},
  school={Ulsan National Institute of Science and Technology}
}

@inproceedings{2025-education-zhong,
  title={From Information to Insight: Leveraging LLMs for Open Aspect-Based Educational Summarization},
  author={Zhong, Yang and Litman, Diane},
  booktitle={Proceedings of the 63rd Annual Meeting of the Association for Computational Linguistics (Volume 1: Long Papers)},
  pages={1914--1947},
  year={2025}
}

@inproceedings{2024-dialogue-ramprasad,
  title={Analyzing LLM Behavior in Dialogue Summarization: Unveiling Circumstantial Hallucination Trends},
  author={Ramprasad, Sanjana and Ferracane, Elisa and Lipton, Zachary C},
  booktitle={Proceedings of the 62nd Annual Meeting of the Association for Computational Linguistics (Volume 1: Long Papers)},
  pages={12549--12561},
  year={2024}
}

@inproceedings{2025-conversation-gupta,
  title={AUTOSUMM: A comprehensive framework for LLM-based conversation summarization},
  author={Gupta, Abhinav and Singh, Devendra and Cowan, Greig A and Kadhiresan, N and Srivastava, Siddharth and Sriraja, Yagneswaran and Mantri, Yoages Kumar},
  booktitle={Proceedings of the 63rd Annual Meeting of the Association for Computational Linguistics (Volume 6: Industry Track)},
  pages={500--509},
  year={2025}
}

@inproceedings{2024-scientific-keya,
  title={Leveraging LLMs for Scientific Abstract Summarization: Unearthing the Essence of Research in a Single Sentence},
  author={Keya, Farhana and Jaradeh, Mohamad Yaser and Auer, S{\'o}ren},
  booktitle={Proceedings of the 24th ACM/IEEE Joint Conference on Digital Libraries},
  pages={1--7},
  year={2024}
}

@article{2025-student-xie,
  title={Using LLM-supported lecture summarization system to improve knowledge recall and student satisfaction},
  author={Xie, Tao and Kuang, Yuanyuan and Tang, Ying and Liao, Jian and Yang, Yunong},
  journal={Expert Systems with Applications},
  volume={269},
  pages={126371},
  year={2025},
  publisher={Elsevier}
}

@article{2023-literature-antu,
  title={Using LLM (Large Language Model) to Improve Efficiency in Literature Review for Undergraduate Research.},
  author={Antu, Shouvik Ahmed and Chen, Haiyan and Richards, Cindy K},
  journal={Llm@ Aied},
  pages={8--16},
  year={2023}
}

@inproceedings{2024-scientific-procko,
  title={Leveraging Large Language Models on the Traditional Scientific Writing Workflow},
  author={Procko, Tyler Thomas and Davidoff, Alexandra and Elvira, Timothy and Ochoa, Omar},
  booktitle={2024 Conference on AI, Science, Engineering, and Technology (AIxSET)},
  pages={154--161},
  year={2024},
  organization={IEEE}
}

@article{2024-mizrahi-state,
  title={State of what art? a call for multi-prompt llm evaluation},
  author={Mizrahi, Moran and Kaplan, Guy and Malkin, Dan and Dror, Rotem and Shahaf, Dafna and Stanovsky, Gabriel},
  journal={Transactions of the Association for Computational Linguistics},
  volume={12},
  pages={933--949},
  year={2024},
  publisher={MIT Press 255 Main Street, 9th Floor, Cambridge, Massachusetts 02142, USA~…}
}

@article{2024-wang-prompt,
  title={Prompt engineering in consistency and reliability with the evidence-based guideline for LLMs},
  author={Wang, Li and Chen, Xi and Deng, XiangWen and Wen, Hao and You, MingKe and Liu, WeiZhi and Li, Qi and Li, Jian},
  journal={NPJ digital medicine},
  volume={7},
  number={1},
  pages={41},
  year={2024},
  publisher={Nature Publishing Group UK London}
}

@inproceedings{2024-related-polo-efficient,
  title={Efficient multi-prompt evaluation of LLMs},
  author={Polo, Felipe Maia and Xu, Ronald and Weber, Lucas and Silva, M{\'\i}rian and Bhardwaj, Onkar and Choshen, Leshem and de Oliveira, Allysson Flavio Melo and Sun, Yuekai and Yurochkin, Mikhail},
  booktitle={Proceedings of the 38th International Conference on Neural Information Processing Systems},
  pages={22483--22512},
  year={2024}
}

@article{2023-related-sclar-quantifying,
  title={{Quantifying Language Models' Sensitivity to Spurious Features in Prompt Design or: How I learned to start worrying about prompt formatting}},
  author={Sclar, Melanie and Choi, Yejin and Tsvetkov, Yulia and Suhr, Alane},
  journal={arXiv preprint arXiv:2310.11324},
  year={2023}
}

@article{2019-zhang-bertscore,
  title={Bertscore: Evaluating text generation with bert},
  author={Zhang, Tianyi and Kishore, Varsha and Wu, Felix and Weinberger, Kilian Q and Artzi, Yoav},
  journal={arXiv preprint arXiv:1904.09675},
  year={2019}
}

@inproceedings{2023-zha-alignscore,
  title={AlignScore: Evaluating Factual Consistency with A Unified Alignment Function},
  author={Zha, Yuheng and Yang, Yichi and Li, Ruichen and Hu, Zhiting},
  booktitle={Proceedings of the 61st Annual Meeting of the ACL (Volume 1: Long Papers)},
   pages={11328--11348},
  year={2023}
}

@article{2006-demvsar-statistical,
  title={Statistical comparisons of classifiers over multiple data sets},
  author={Dem{\v{s}}ar, Janez},
  journal={Journal of Machine learning research},
  volume={7},
  number={Jan},
  pages={1--30},
  year={2006}
}

@article{2011-yap-stats-test,
  title={Comparisons of various types of normality tests},
  author={Yap, Bee Wah and Sim, Chiaw Hock},
  journal={Journal of Statistical Computation and Simulation},
  volume={81},
  number={12},
  pages={2141--2155},
  year={2011},
  publisher={Taylor \& Francis}
}

@article{1974-levenes-test-brown,
  title={Robust tests for the equality of variances},
  author={Brown, Morton B and Forsythe, Alan B},
  journal={Journal of the American statistical association},
  volume={69},
  number={346},
  pages={364--367},
  year={1974},
  publisher={Taylor \& Francis}
}

@article{peturbation-alahmari2025large,
  title={Large language models robustness against perturbation: S. Alahmari et al.},
  author={Alahmari, Saeed S and Hall, Lawrence and Mouton, Peter R and Goldgof, Dmitry},
  journal={Scientific Reports},
  year={2025},
  publisher={Nature Publishing Group UK London}
}

@article{peturbation-singh2024robustness,
  title={Robustness of Large Language Models to Perturbations in Text},
  author={Singh, Ayush and Singh, Navpreet and Vatsal, Shubham},
  journal={arXiv preprint arXiv:2407.08989},
  year={2024}
}

@inproceedings{peturbation-romero2024resilient,
  title={How resilient are language models to text perturbations?},
  author={Romero-Alvarado, Daniel and Hern{\'a}ndez-Orallo, Jos{\'e} and Mart{\'\i}nez-Plumed, Fernando},
  booktitle={International Conference on Intelligent Data Engineering and Automated Learning},
  pages={85--96},
  year={2024},
  organization={Springer}
}

@inproceedings{peturbation-oved2021pass,
  title={Pass: Perturb-and-select summarizer for product reviews},
  author={Oved, Nadav and Levy, Ran},
  booktitle={Proceedings of the 59th Annual Meeting of the ACL and the 11th International Joint Conference on Natural Language Processing (Volume 1: Long Papers)},
  pages={351--365},
  year={2021}
}

@inproceedings{2024-llm-generated-field,
  title={A field guide to automatic evaluation of llm-generated summaries},
  author={Van Schaik, Tempest A and Pugh, Brittany},
  booktitle={Proceedings of the 47th International ACM SIGIR Conference on Research and Development in Information Retrieval},
  pages={2832--2836},
  year={2024}
}

@article{zhang2026assessing,
  title={Assessing the Accuracy of Artificial Intelligence-Generated Clinical Summaries From Ambulatory Glaucoma Subspecialty Clinical Encounters},
  author={Zhang, Yapei and Shi, Min and Chung, In Young and Liebman, Daniel L and Barna, Laura E and Pasquale, Louis R and Friedman, David S and Boland, Michael V and Shen, Lucy Q and Wang, Mengyu},
  journal={Translational Vision Science \& Technology},
  volume={15},
  number={1},
  pages={22--22},
  year={2026},
  publisher={The Association for Research in Vision and Ophthalmology}
}

@article{gilhuly2025consistency,
  title={Consistency evaluation of news article summaries generated by large (and small) language models},
  author={Gilhuly, Colleen and Shahzad, Haleh},
  journal={arXiv preprint arXiv:2502.20647},
  year={2025}
}

@article{2026-trustworthy-zhang,
  title={Trustworthy evaluation of large language models},
  author={Zhang, Xin-Yi and Ye, Han-Jia and Zhan, De-Chuan},
  journal={Frontiers of Computer Science},
  volume={20},
  number={2},
  pages={1--3},
  year={2026},
  publisher={Springer}
}

@article{2024-can-trust-schroeder,
  title={Can you trust llm judgments? reliability of llm-as-a-judge},
  author={Schroeder, Kayla and Wood-Doughty, Zach},
  journal={arXiv preprint arXiv:2412.12509},
  year={2024}
}

@inproceedings{2025-not-abstain-liu,
  title={Do not abstain! identify and solve the uncertainty},
  author={Liu, Jingyu and JingquanPeng, JingquanPeng and Wu, Xiaopeng and Li, Xubin and Ge, Tiezheng and Zheng, Bo and Liu, Yong},
  booktitle={Proceedings of the 63rd Annual Meeting of the Association for Computational Linguistics (Volume 1: Long Papers)},
  pages={17177--17197},
  year={2025}
}

@inproceedings{2025-using-huidrom,
  title={Using LLM judgements for sanity checking results and reproducibility of human evaluations in NLP},
  author={Huidrom, Rudali and Belz, Anja},
  booktitle={Proceedings of the Fourth Workshop on Generation, Evaluation and Metrics (GEM$^2$)},
  pages={354--365},
  year={2025}
}

@article{2026-efficient-inference-chen,
  title={Efficient inference for noisy LLM-as-a-judge evaluation},
  author={Chen, Yiqun T and Lu, Sizhu and Li, Sijia and Guo, Moran and Li, Shengyi},
  journal={arXiv preprint arXiv:2601.05420},
  year={2026}
}

@article{2025-llms-li,
  title={Llms cannot reliably judge (yet?): A comprehensive assessment on the robustness of llm-as-a-judge},
  author={Li, Songze and Xu, Chuokun and Wang, Jiaying and Gong, Xueluan and Chen, Chen and Zhang, Jirui and Wang, Jun and Lam, Kwok-Yan and Ji, Shouling},
  journal={arXiv preprint arXiv:2506.09443},
  year={2025}
}

@article{2024-trustllm-huang,
  title={Trustllm: Trustworthiness in large language models},
  author={Huang, Yue and Sun, Lichao and Wang, Haoran and Wu, Siyuan and Zhang, Qihui and Li, Yuan and Gao, Chujie and Huang, Yixin and Lyu, Wenhan and Zhang, Yixuan and others},
  journal={arXiv preprint arXiv:2401.05561},
  year={2024}
}

@article{2026-coin-flip-schwinn,
  title={A coin flip for safety: Llm judges fail to reliably measure adversarial robustness},
  author={Schwinn, Leo and Ladenburger, Moritz and Beyer, Tim and Mofakhami, Mehrnaz and Gidel, Gauthier and G{\"u}nnemann, Stephan},
  journal={arXiv preprint arXiv:2603.06594},
  year={2026}
}

@article{2023-AI-risk-tabassi,
  title={Artificial Intelligence Risk Management Framework (AI RMF 1.0).(2023)},
  author={Tabassi, Elham},
  journal={URL: http://nvlpubs. nist. gov/nistpubs/ai/NIST. AI},
  pages={100--1},
  year={2023}
}

@article{2025-tools-lymperopoulos,
  title={Tools in the loop: Quantifying uncertainty of llm question answering systems that use tools},
  author={Lymperopoulos, Panagiotis and Sarathy, Vasanth},
  journal={arXiv preprint arXiv:2505.16113},
  year={2025}
}

@inproceedings{2025-investigating-lupart,
  title={Investigating llm variability in personalized conversational information retrieval},
  author={Lupart, Simon and van Dijk, Dani{\"e}l and Langezaal, Eric and van Dort, Ian and Aliannejadi, Mohammad},
  booktitle={Proceedings of the 2025 Annual International ACM SIGIR Conference on Research and Development in Information Retrieval in the Asia Pacific Region},
  pages={353--363},
  year={2025}
}

@article{2025-evaluating-nikitha,
  title={Evaluating variance in visual question answering benchmarks},
  author={Nikitha, SR},
  journal={arXiv preprint arXiv:2508.02645},
  year={2025}
}


\end{document}